%% file: main.tex
\documentclass[11pt]{article}
\usepackage[preprint]{acl}
\usepackage{times}
\usepackage{latexsym}
\usepackage{graphicx}
\usepackage[utf8]{inputenc}
\usepackage{enumitem}
\usepackage{xcolor}
\usepackage{natbib}
\usepackage{hyperref}
\usepackage{booktabs}
\usepackage{pifont}

\usepackage{tabularx}
\usepackage{array}
\usepackage{tcolorbox}
\usepackage{multirow}
\tcbuselibrary{skins,breakable}
\usepackage{amsmath} 
\usepackage{pgfplots}
\pgfplotsset{compat=1.18}
\usepackage{pgfplotstable}
\usepackage{subcaption}    
\usepackage{makecell}

\definecolor{pendingred}{RGB}{200,30,30}
\newcommand{\cmark}{\ding{51}}
\newcommand{\xmark}{\ding{55}}
\definecolor{llmblue}{RGB}{31,119,180}
\definecolor{llmred}{RGB}{214,39,40}

\usepackage{tikz}
\usepackage{booktabs}
\usepackage{longtable}
\usepackage{array}
\usetikzlibrary{positioning, arrows.meta, shapes.geometric, calc}

\newcolumntype{L}[1]{>{\raggedright\arraybackslash}p{#1}}
\newcolumntype{Y}{>{\raggedright\arraybackslash}X}

\usepackage[utf8]{inputenc}
\DeclareUnicodeCharacter{2550}{-}
\DeclareUnicodeCharacter{2500}{-}

\author{
  \begin{tabular}{c}
    {\bf Muhammed Saeed} \qquad {\bf Simon Razniewski} \\
    ScaDS.AI Dresden/Leipzig \& TU Dresden, Germany \\
    \texttt{\{muhammed.saeed,simon.razniewski\}@tu-dresden.de}
  \end{tabular}
}

\title{LLMpedia: A Transparent Framework to Materialize an LLM's Encyclopedic Knowledge at Scale}

\begin{document}
\maketitle

\input{sections/all}


\bibliography{refs}

\appendix
\input{appendices/appendix}

\end{document}

%% file: sections/all.tex
\begin{abstract}
Benchmarks like MMLU suggest flagship language models approach factuality
saturation above 90\%. \emph{LLMpedia} shows this picture is incomplete.
We materialize ${\sim}$1.3M encyclopedia articles entirely from parametric
memory across three model families, then audit every claim against
Wikipedia and curated web evidence. For \texttt{gpt-5-mini}, the verifiable
true rate is 68.4\% on Wikipedia-covered subjects - more than 21\,pp below
MMLU - and the gap is driven by \emph{unverifiability} (30.5\%), not
refutation (1.2\%). Beyond Wikipedia, frontier articles audited against
curated web evidence reach 57.6\%; Wikipedia covers only 56.7\% of
model-surfaced subjects, and three model families overlap in just 7.3\% of
subject choices. In a retrieval-trap benchmark inspired by prior analysis
of Grokipedia, LLMpedia is more factual at roughly half the textual
similarity to Wikipedia. Every prompt, article, and verdict is released.
Data, code, interface: \url{https://llmpedia.net}.
\end{abstract}

\section{Introduction}
\label{sec:intro}

Hundreds of millions of users now consume LLM output as fluent
multi-paragraph prose, not multiple-choice answers
\citep{openai2025howpeopleusechatgpt,anthropic2025economic,
chatterji2025people,handa2025clio}. Yet factuality is still measured with
curated short-answer suites: MMLU \citep{hendrycks2021mmlu}, TruthfulQA
\citep{lin2022truthfulqa}, HLE \citep{hendrycks2025hle}. These suites test
only what the experimenter thought to ask-the classic
\emph{availability bias} of \citet{tversky1973availability}. A model can
saturate a fixed question set while broad regions of weak or unverifiable
parametric knowledge remains unmeasured, a gap
sharpest in the long-form modality, users actually read \citep{saeed2025surfacing}.

\begin{figure}[t]
  \centering
  \includegraphics[width=\linewidth]{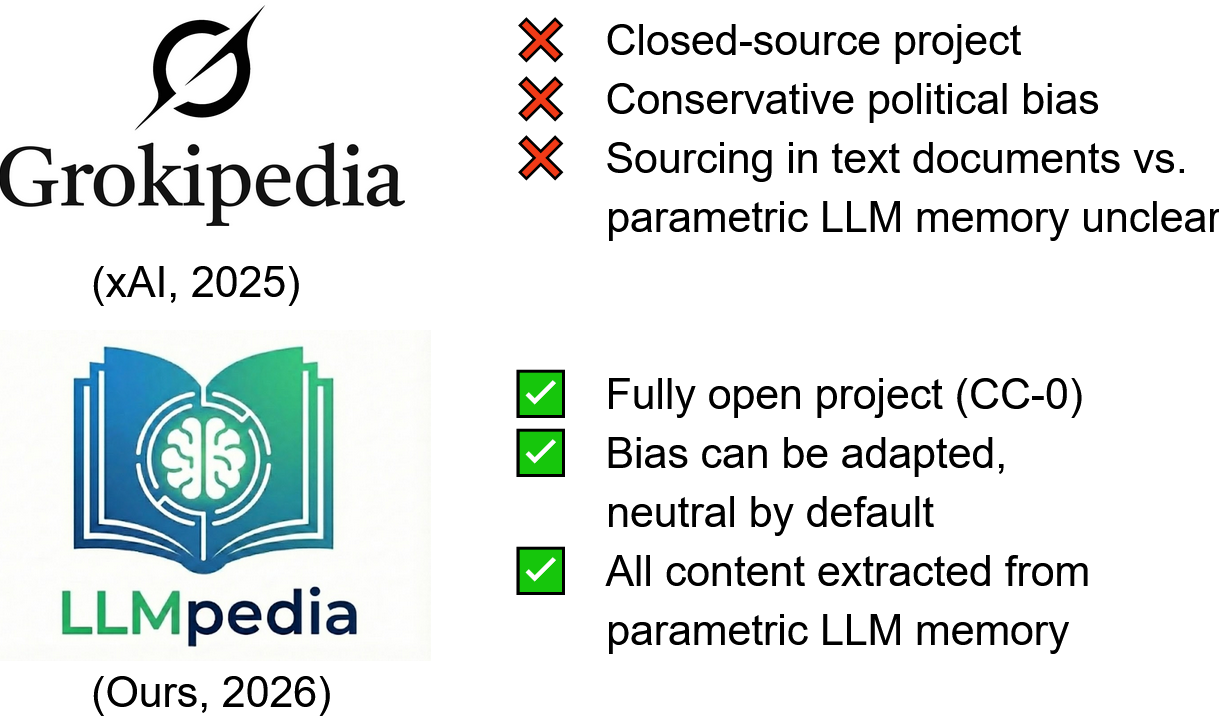}
  \caption{LLMpedia generates entirely from parametric memory with full
  auditability. Grokipedia's opaque pipeline shows evidence of
  retrieval-shaped generation \citep{yasseri2025similar}.}
  \label{fig:teaser}
\end{figure}

\emph{Materialization} offers an alternative: surface what the model
believes on its own terms rather than probing it with a fixed question
set. \citet{cohen2023crawling} introduced recursive elicitation;
\citet{HuGPTKB2025} scaled this into GPTKB (100M+ triples) and made
availability bias the explicit target. But triples are not the modality of
LLM consumption: users read discourse-level prose \cite{saeed2025surfacing}. Materialization must
move from triples to free text, and \emph{encyclopedia articles} are the
natural target-open enough to escape question-selection bias,
constrained enough for tractable claim-level evaluation, and matching the
format LLMs were trained on and that users consume
\citep{openai2025howpeopleusechatgpt,anthropic2025economic}.

Choosing encyclopedia articles brings their verification strategy with
them. Wikipedia-the largest human-curated reference of exactly this
content type-partitions claims into \emph{supported}, \emph{refuted},
or \emph{insufficient}; for subjects Wikipedia omits, a second tier of
133 curated domains (encyclopedias, government sites, academic
publishers, wire services; \S\ref{sec:eval},
Appendix~\ref{app:evidence-scoring}) reaches the rest. The two-tier
design matters: Wikipedia is documented to skew Western
\citep{WikipediaSystemicBias}, gendered \citep{WikipediaGenderBias}, and
under-represent the Global South \citep{WikipediaSizeWikipedia}; GPTKB
v1.5 finds ${\sim}$57\% of materialized entities absent from Wikipedia and found that GPTKB has more female representations compared to Wikipedia as debiasing work of LLMs
\citep{HuGPTKB2025,saeed-etal-2025-beyond}. The \emph{insufficient}
class is therefore informative-it isolates claims that even the
world's largest encyclopedia cannot adjudicate. At the article level,
43\% of LLMpedia subjects fall outside Wikipedia, with a 57.6\% true
rate on the verifiable portion under curated-web verification.

Auditability is non-negotiable because the most visible LLM-generated
encyclopedia, Grokipedia \citep{grokipedia}, is methodologically opaque.
Prior work reports TF-IDF cosine of 0.46--0.49 against Wikipedia
\citep{yasseri2025similar}; we document further retrieval-shaped
indicators in Appendix~\ref{app:grokipedia}. We call this the
\emph{retrieval trap}: lexical proximity to Wikipedia does not imply
factual superiority and may reflect rewriting that obscures both
parametric knowledge and its failure modes.

We present \emph{LLMpedia}, an open framework for parametric
encyclopedia generation across \texttt{gpt-5-mini},
\texttt{Llama-3.3-70B}, and \texttt{DeepSeek-V3-0324}: ${\sim}$1M
articles from GPT-5-mini and 120K each from the open-weight models in
general-domain BFS, 27K in topic-focused runs over a controversy
gradient (\emph{Ancient Babylon}, \emph{US Civil Rights}, \emph{Dutch
Colonization in Southeast Asia}) plus two low-contestedness controls
(\emph{One Piece}, \emph{Quantum Physics}), and 1K for the
retrieval-trap benchmark.

\noindent\textbf{RQ1}: What do topic-focused parametric encyclopedias
reveal about cross-model factuality, coverage, and persona effects?\\
\textbf{RQ2}: At $10^5$--$10^6$ scale, how do three model families
differ in knowledge breadth, subject overlap, and factuality across BFS
depth?\\
\textbf{RQ3}: Can parametric generation escape the Wikipedia retrieval
trap-producing factually competitive but structurally independent
content?

\input{sections/tables/comparisons}

\begin{enumerate}[nosep]
  \item A fully open pipeline for parametric encyclopedia
    construction-prompts, artifacts, and verdicts.
  \item Evidence that \emph{availability bias is a first-order
    distortion}: true rate is 68.4\% on Wikipedia-covered subjects
    ($>$21\,pp below MMLU), driven by unverifiability (30.5\%), not
    falsehood (1.2\%); validated against human verdicts.
  \item Evidence that the beyond-Wikipedia frontier is real but
    difficult: 43\% of subjects fall outside Wikipedia; among those
    with usable curated-web evidence, true rate is 57.6\%.
  \item A retrieval-trap benchmark where LLMpedia outperforms
    Grokipedia at roughly half the Wikipedia similarity.
\end{enumerate}

\section{Related Work}
\label{sec:related}

\paragraph{Knowledge probing and its limits.}
LAMA \citep{petroni2019lama} probes recall via cloze prompts; MMLU
\citep{hendrycks2021mmlu} spans knowledge-heavy domains; TruthfulQA
\citep{lin2022truthfulqa} targets misconceptions; HLE
\citep{hendrycks2025hle} pushes frontier difficulty. All measure only
what the experimenter thought to ask. LLMpedia complements benchmarks
by materializing knowledge at scale and decomposing outputs into
supported, refuted, and insufficient.

\paragraph{From triples to articles.}
\citet{cohen2023crawling} introduced recursive elicitation;
\citet{HuGPTKB2025} scaled this into GPTKB (101M triples for 2.9M
entities) and made availability bias the explicit target.
\citet{ghosh2025mining} mine the resulting KB for novel knowledge;
\citet{giordano-razniewski-2026-foundations} find within-model
variance from seed choice comparable to repeated-run variance, while
cross-model variance is much larger. LLMpedia materializes
discourse-level articles, enabling evaluation of structure, subject
choice, and unverifiability in free text.

\paragraph{Atomic claim factuality.}
The decompose--retrieve--verify paradigm is standard for long-form
factuality. \textsc{FActScore} \citep{min-etal-2023-factscore} introduced it
with binary \emph{Supported}/\emph{Not-supported} verification against
Wikipedia; \textsc{SAFE}
\citep{wei2024safe} extended retrieval to Google Search; \textsc{VeriScore}
\citep{song-etal-2024-veriscore} added a three-label scheme restricted to
verifiable claims; \textsc{VeriFastScore}
\citep{rajendhran-etal-2025-verifastscore} collapsed the pipeline but
reverted to binary. We follow VeriScore's
three labels, but spend additional effort on curating websites used for evidence retrieval, compared with generic web search. As evaluation costs are small in our case, we disregard local models like VeriScore and directly prompt commercial APIs.

\paragraph{Encyclopedia generation and the retrieval trap.}
STORM \citep{shao-etal-2024-assisting} models pre-writing through
web-grounded interactions but remains retrieval-dependent.
\citet{Gao2024SurveyGen} find outlines improve generation; LLMpedia
extends this with dynamic, entity-tailored outlines. Grokipedia
\citep{grokipedia} operates at the largest scale but discloses no
methodology; \citet{yasseri2025similar} report TF-IDF cosine of
0.46--0.49 against Wikipedia (LLMpedia: 0.256), suggesting
retrieval-shaped repackaging.

\paragraph{Hallucination and provenance.}
Hallucination remains central to factual generation
\citep{ji2023hallucination}. RAG can improve factuality
\citep{lewis2020rag,mallen2023nottrust} but complicates provenance:
once external passages are introduced, separating parametric memory
from retrieved text becomes hard
\citep{AlJazeera2025Grokipedia,yasseri2025similar}. LLMpedia avoids
retrieval at generation time precisely to keep the two separable.

\section{LLMpedia Framework}
\label{sec:method}

\begin{figure*}[t]
  \centering
  \includegraphics[width=\textwidth]{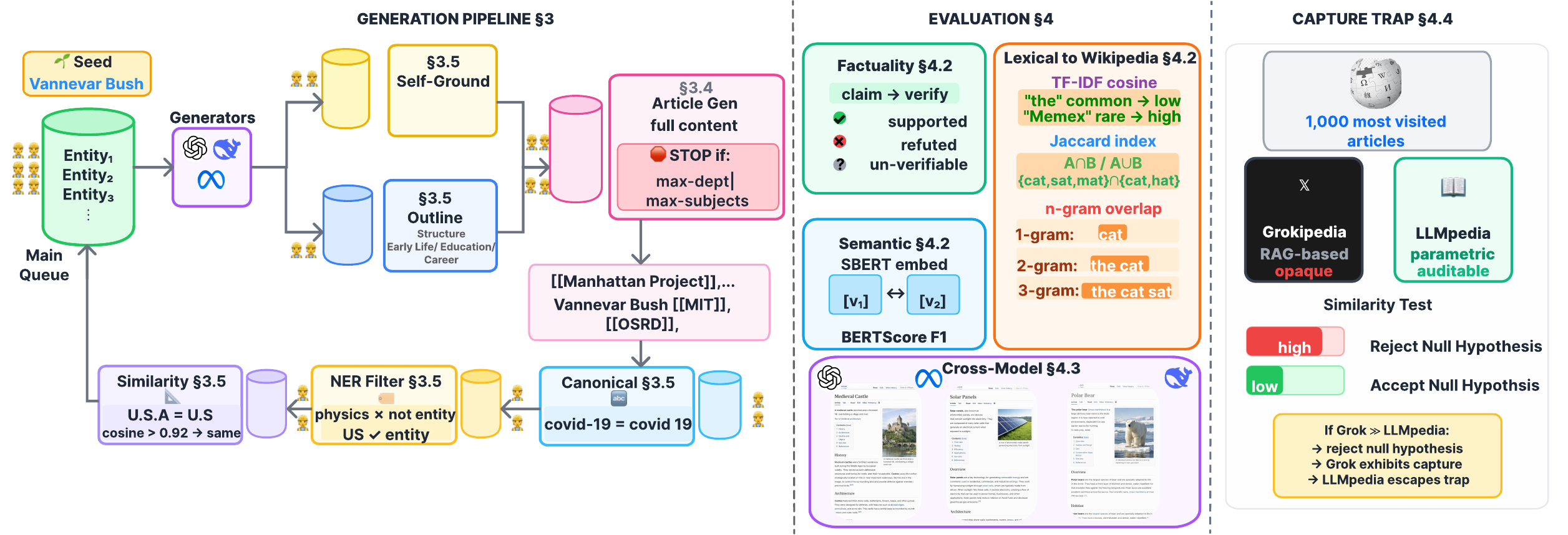}
  \caption{LLMpedia pipeline. Each subject flows through optional
  self-grounding, dynamic outline generation, and article
  elicitation. Extracted \texttt{[[wikilinks]]} undergo canonical
  normalization, LLM-based encyclopedic filtering, and
  embedding-based deduplication before surviving children enter the
  BFS queue. Details in Appendix~\ref{app:implementation}.}
  \label{fig:pipeline}
\end{figure*}

Figure~\ref{fig:pipeline} shows the full pipeline. From a seed
entity, the system expands breadth-first through optional
self-grounding, outline generation, article elicitation, and
three-stage entity sanitization before enqueuing surviving children.
Prompts and artifacts are logged; full templates in
Appendix~\ref{app:prompts}.

\subsection{Materialization Modes}
\label{sec:materialization}

\begin{figure}[h]
\centering
\resizebox{0.97\columnwidth}{!}{%
\begin{tikzpicture}[
  >=Stealth,
  every node/.style={font=\normalsize},
  seed/.style={draw, rounded corners=3pt, inner sep=4pt,
               minimum height=20pt, fill=orange!12,
               line width=0.9pt, font=\normalsize\bfseries},
  accepted/.style={draw, rounded corners=3pt, inner sep=4pt,
                   minimum height=20pt, fill=white,
                   text=blue, font=\normalsize},
  rejected/.style={draw, rounded corners=3pt, inner sep=4pt,
                   minimum height=20pt, fill=white,
                   text=red, dashed, draw=red!70,
                   font=\normalsize\itshape},
  hop2/.style={draw, rounded corners=3pt, inner sep=3pt,
               minimum height=18pt, fill=white,
               text=blue, font=\small},
  myarrow/.style={-{Stealth[scale=0.6]}, thick, black!60},
  rejarrow/.style={-{Stealth[scale=0.6]}, thick, red!45, dashed},
]
\node[seed]      (vb)   at (0,0)      {Vannevar Bush};
\node[accepted]  (mit)  at (4.0,1.0)
  {\textcolor{blue}{MIT}};
\node[accepted]  (osrd) at (4.0,-0.2)
  {\textcolor{blue}{Office of Scientific R\&D}};
\node[rejected]  (eng)  at (4.0,-1.4)
  {\textcolor{red}{engineer}};
\draw[myarrow]  (vb) -- (mit);
\draw[myarrow]  (vb) -- (osrd);
\draw[rejarrow] (vb) -- (eng);
\node[hop2] (rf)   at (8.0,1.5)
  {\textcolor{blue}{Richard Feynman}};
\node[hop2] (mp)   at (8.0,0.5)
  {\textcolor{blue}{Manhattan Project}};
\node[hop2] (wwii) at (8.0,-0.5)
  {\textcolor{blue}{World War II}};
\draw[myarrow, thin, black!45] (mit)  -- (rf);
\draw[myarrow, thin, black!45] (mit)  -- (mp);
\draw[myarrow, thin, black!45] (osrd) -- (wwii);
\node[font=\sffamily\footnotesize, text=black!50,
      anchor=south] at (0,1.8)   {hop\,0};
\node[font=\sffamily\footnotesize, text=black!50,
      anchor=south] at (4.0,1.8) {hop\,1};
\node[font=\sffamily\footnotesize, text=black!50,
      anchor=south] at (8.0,1.8) {hop\,2};
\end{tikzpicture}%
}
\caption{BFS from \emph{Vannevar Bush}. \textcolor{blue}{Blue}
nodes survive sanitization; \textcolor{red}{red dashed} nodes are
filtered as generic.}
\label{fig:graph-exploration}
\end{figure}

\paragraph{General-domain expansion.}
Starting from a single seed entity (\emph{Vannevar Bush},
Figure~\ref{fig:graph-exploration}), BFS expands without topical
restrictions. The choice of seed is empirically inconsequential:
\citet{HuGPTKB2025} construct a 100M-triple KB from the same seed,
and \citet{giordano-razniewski-2026-foundations} systematically
perturb seeds and find within-model variance comparable to
repeated-run variance. Using BFS also matches prior materialization
work \citep{cohen2023crawling,HuGPTKB2025}, enabling direct
comparison. Each generated article is written in Wikitext with
\texttt{[[wikilinks]]}; links that survive sanitization are enqueued
as future article subjects. This mode scales to $10^5$--$10^6$
articles.

\paragraph{Topic-focused generation.}
Topic-focused BFS is constrained around a root topic
(\emph{Ancient Babylon}, \emph{US Civil Rights Movement},
\emph{Dutch Colonization in Southeast Asia}). Each run is seeded
with a topic-specific subject list; generated
\texttt{[[wikilinks]]} must remain topically associated
(Appendix~\ref{app:prompts}); only relevant survivors are
re-enqueued. This yields a controlled regime for cross-model,
persona, and domain-difficulty analysis.

\begin{figure}[h]
\centering
\small
\colorbox{blue!8}{\parbox{0.95\linewidth}{\texttt{%
\textbf{(a) Outline} \textit{(three sections, adapted to entity)}\\[1pt]
\textbf{Vannevar Bush}\\
~~1.~Early life and education\\
~~2.~Engineering career and Raytheon\\
~~3.~OSRD and wartime science leadership
}}}\\[3pt]
\colorbox{orange!10}{\parbox{0.95\linewidth}{\texttt{%
\textbf{(b) Generated Wikitext}\\[1pt]
'''Vannevar Bush''' was an American
\textcolor{red}{[[engineer]]} who headed the
\textcolor{blue}{[[Office of Scientific R\&D]]}
during \textcolor{blue}{[[World War II]]}\ldots\\[2pt]
== Early life and education ==\\
Bush was born in \textcolor{blue}{[[Everett, Massachusetts]]} and
studied at \textcolor{blue}{[[Tufts University]]}\ldots\\[2pt]
== OSRD and wartime science leadership ==\\
In 1940, Bush proposed the formation of the
\textcolor{blue}{[[NDRC]]}\ldots
}}}
\caption{The model proposes an entity-tailored outline (top, blue)
then writes the article (bottom, orange) under verbatim section
headers. \textcolor{blue}{Blue} wikilinks survive sanitization;
\textcolor{red}{red} \texttt{[[engineer]]} is filtered as generic.}
\label{fig:outline-and-wikitext}
\end{figure}

\subsection{Outline, Generation, and Prompting}
\label{sec:outline-and-generation}

Outlines improve LLM-based encyclopedia generation
\citep{Gao2024SurveyGen}, typically with fixed templates. LLMpedia
proposes 6--7 section headings tailored to the entity, then writes
Wikitext to match (Figure~\ref{fig:outline-and-wikitext}).

\noindent Optional self-grounding \citep{tian2023justask} produces a
per-entity fact sheet (summary, aliases, predicate--object pairs
with confidence $\geq 0.75$); we treat it as an ablation variable
(\S\ref{sec:ablation}). The \emph{baseline} prompt requests plain
wikilinks; the \emph{calibrated} prompt requires per-entity
confidence (\texttt{[[Entity (0.85)]]}, range 0.75--1.00); entities
below 0.75 are discarded before NER.

\subsection{Entity Sanitization Pipeline}
\label{sec:sanitization}

GPTKB v1 and v1.5 document persistent entity disambiguation failures
\citep{HuGPTKB2025}. We address these with three stages, each with
a distinct role.

\paragraph{Stage~1: Canonical Normalization (re-encounter dedup).}
Unicode NFKC, case-folding, and punctuation removal map surface
variants of the \emph{same} entity to identical keys
(``U.S.''/``US''$\to$\texttt{us};
``Covid-19''/``Covid 19''$\to$\texttt{covid19}). This collapses
re-encounters of already-committed entities; \emph{it is not entity
loss}. The large raw-to-canon reduction (70.2M$\to$2.29M for
GPT-5-mini) reflects natural re-linking - every article re-mentions
common entities the pipeline has already canonicalized. Stage~1
cannot disambiguate homonyms (e.g., \emph{Apple} the company vs.\
the fruit); that is Stage~3's job.

\paragraph{Stage~2: LLM-Based Encyclopedic Filtering.}
Not every wikilink warrants its own article. Generic forms like
\texttt{[[engineer]]} (Figure~\ref{fig:graph-exploration}) produce
thin, looping sub-articles; in contrast,
\texttt{[[Office of Scientific Research and Development]]} expands
into a self-contained article. Following \citet{HuGPTKB2025} we use
the LLM itself to classify candidate batches. The filter is
intentionally conservative: a candidate must be positively
classified non-encyclopedic to drop, and filtered candidates can be
re-nominated at later BFS waves.

\paragraph{Stage~3: Semantic Dedup and Sense Disambiguation.}
This stage handles two distinct cases: (i) semantic duplicates with
different surface forms (\emph{Deutschland}/\emph{Germany},
\emph{Oxford University}/\emph{University of Oxford}), and
(ii) true homonyms with identical surface forms (\emph{Apple} the
company vs.\ the fruit). We maintain an embedding index
(\texttt{text-embedding-3-small}) over committed entities;
candidates above cosine 0.90 trigger LLM arbitration over the first
${\sim}$30 words of each candidate's lead paragraph plus parent
context. For case (i), the leads converge and the candidates merge;
for case (ii), the leads diverge semantically (Apple Inc.\ discusses
Cupertino, consumer electronics; Apple the fruit discusses
\emph{Malus domestica}, cultivars) and arbitration keeps them
separate. Within-wave deduplication merges concurrent proposals
before commit (Appendix~\ref{app:dedup-guarantee}). Homonyms with
identical surface forms \emph{and} closely related leads (e.g.,
\emph{Dresden} the German city vs.\ U.S.\ locality) remain a known
limitation (\S\ref{sec:limitations}).

\paragraph{Architecture and persona injection.}
Each stage has an independent work queue with retries and
exponential backoff, supporting batch and online modes
(Appendix~\ref{app:settings}). To study framing as an explicit
variable, we inject three personas (scientific-neutral,
left-leaning, conservative) at the system level of \emph{all}
topic-focused stages, affecting prose, outlines, link proposals,
and entity filtering.

\section{Experimental Setup}
\label{sec:setup}

\subsection{Models and Conditions}

We evaluate \texttt{gpt-5-mini} \citep{openai2025gpt5mini},
\texttt{Llama-3.3-70B-Instruct} \citep{grattafiori2024llama}
(4$\times$A100 on-prem), and \texttt{DeepSeek-V3-0324} (671B/37B
MoE, 16$\times$H100 on-prem) \citep{liu2024deepseek}. All runs use
temperature~0 with fixed seeds. GPT-5-mini is primary for
instruction-following compliance and cost efficiency; the
open-weight models test generalization beyond a single API.

\paragraph{Topic-focused (RQ1).}
Motivated by bias concerns around Grokipedia
\citep{grokipedia,yasseri2025similar}, we test whether persona
injection produces measurable framing shifts on the same subject.
We span a controversy gradient: \emph{US Civil Rights} and
\emph{Dutch Colonization in Southeast Asia} as contested
narratives, \emph{Ancient Babylon} as a politically neutral
control. Each topic seeds ${\sim}$1K subjects under three personas
across all three models - 27 cells, ${\sim}$27K articles.

\paragraph{Large-scale materialization (RQ2).}
All three models start from \emph{Vannevar Bush}. GPT-5-mini
produces ${\sim}$1M articles; open-weight models produce
${\sim}$120K each. We compare on the first 120K subjects.

\paragraph{Retrieval-trap (RQ3).}
Following \citet{yasseri2025similar}, we collect the 1{,}000
most-edited English Wikipedia articles confirmed present on
Grokipedia and evaluate both systems against Wikipedia (Tier~1) and
curated web sources (Tier~2).

\subsection{Evaluation Protocol}
\label{sec:eval}

We adopt the decompose--retrieve--verify paradigm established by
\textsc{FActScore} \citep{min-etal-2023-factscore} and extended by
\textsc{SAFE} \citep{wei2024safe}, \textsc{VeriScore}
\citep{song-etal-2024-veriscore}, and \textsc{VeriFastScore}
\citep{rajendhran-etal-2025-verifastscore}: decompose each article
into atomic claims, retrieve evidence per subject, and verify each
claim independently against the retrieved evidence. We follow
\textsc{VeriScore} in using a three-label verdict scheme
(\emph{supported} / \emph{refuted} / \emph{insufficient}) and pair
it with a two-tier evidence stack (Wikipedia + 133 curated domains,
the latter reaching subjects Wikipedia does not cover). The judge
is \texttt{gpt-4.1-nano}; we validate it directly against human
verdicts in \S\ref{sec:human}.

\paragraph{Choice of judge.}
Three design choices justify our verification pipeline.
\emph{First, the label scheme.} VeriScore's three labels (vs.\ binary)
prevent collapsing contradicted and evidence-silent claims into a
single bucket - the distinction that makes the 1.2\% false vs.\
30.5\% unverifiable gap, the signature of availability bias,
measurable. \emph{Second, the judge is human-validated.} Following FActScore's
protocol, we audit \texttt{gpt-4.1-nano} on balanced 33/33/33 samples
from both tiers (\S\ref{sec:human}): per-class agreement on
\emph{supported} and \emph{insufficient} reaches
\textbf{87.9--93.9\%}, with disagreements cancelling in aggregate.
FactScore cross-check on  some sample returned
89.8\% supported, broadly consistent though only partially comparable
across label schemes. \emph{Third, the evidence.} Rather than VeriScore's open Google
Search or FActScore's Wikipedia-only retrieval, we use a curated
133-domain stack with per-domain quality scores
(Appendix~\ref{app:evidence-scoring}) - reaching the 43\% of subjects
Wikipedia omits while excluding the long tail of low-quality pages
an unconstrained Google search would surface.

\paragraph{Stage~1 - Claim extraction.}
The article is sent to the judge with an extraction prompt asking
for up to 10 atomic, self-contained claims; each must mention the
subject explicitly, state one predicate, and avoid opinion or
hedging. Verifiability constraints mirror \textsc{VeriScore}: events
and states with necessary modifiers, excluding opinions,
hypotheticals, and instructions.

\paragraph{Stage~2 - Evidence retrieval.}
Evidence is retrieved \emph{independently} of the generated article,
using only the subject name as query, so the article cannot shape
its own evidence. \textbf{Tier~1 (Wikipedia)} fetches the full
article via MediaWiki with redirect resolution; subjects Wikipedia
does not cover contribute to coverage but not Tier~1 factuality.
\textbf{Tier~2 (curated web)} searches a vetted 133-domain list
(Appendix~\ref{app:evidence-scoring}); Wikipedia and mirrors are
blocked. Up to three top-quality URLs per subject are concatenated.

\paragraph{Stage~3: Semantic Dedup and Sense Disambiguation.}
This stage handles two cases: semantic duplicates with different
surface forms (\emph{Deutschland}/\emph{Germany}, \emph{Oxford
University}/\emph{University of Oxford}) and qualified homonyms
(\emph{Apple Inc.}\ vs.\ \emph{Apple}; \emph{Dresden, Germany} vs.\
\emph{Dresden, Maine}). An embedding index
(\texttt{text-embedding-3-small}) over committed entities triggers
LLM arbitration above cosine 0.90 on the first ${\sim}$30 words of
each candidate's lead. Duplicates merge when leads converge;
homonyms stay separate when leads diverge (Apple Inc.\ discusses
Cupertino; the fruit discusses \emph{Malus domestica}). Bare
homonyms with identical, unqualified surface forms (e.g., two
entities both rendered \emph{Dresden}) collapse at commit - a known
limitation (\S\ref{sec:limitations}).

\paragraph{On long-form vs.\ targeted elicitation.}
Long-form generation entangles parametric knowledge retrieval with
composition, and we do not claim to fully separate them.
Decomposition with independent per-claim verification factors out
composition at the verification step, with the prompt restricted to
propositional support. Targeted elicitation
\citep{petroni2019lama,sun2024headtail} measures knowledge in
isolation but sacrifices the modality users consume - our long-form
regime is complementary.
\paragraph{Metrics.}
With $N$ total claims, $n_s$ supported, $n_r$ refuted, $n_u$
insufficient:
\[
\begin{aligned}
\mathrm{Prec} &= \frac{n_s}{n_s+n_r}, &
\mathrm{True}  &= \frac{n_s}{N},\\[3pt]
\mathrm{False} &= \frac{n_r}{N}, &
\mathrm{Unv}   &= \frac{n_u}{N}.
\end{aligned}
\]
Precision excludes unverifiable claims; \emph{True}, \emph{False},
and \emph{Unv} sum to~1. All metrics are macro-averaged at article
level.

\paragraph{Similarity.}
TF-IDF cosine, token Jaccard, $n$-gram overlap ($n{=}1{-}3$), and
semantic cosine (\texttt{text-embedding-3-small}) between generated
articles and Wikipedia.

\subsection{Human Validation}
\label{sec:human}

To audit the LLM judge, we drew balanced stratified samples from
its verdict distribution and asked a human annotator to
agree/disagree and supply the correct class otherwise.
\textsc{Tier~1 (Wikipedia)}: 99 (claim, evidence, verdict) triples
balanced 33/33/33; \textsc{Tier~2 (curated web)}: 30 triples
balanced 10/10/10 from the frontier-Web subset. Per-class agreement
on the contribution-bearing classes (\emph{supported}, driving true
rate; \emph{insufficient}, driving unverifiable rate) reaches
\textbf{87.9--93.9\%} across both tiers. Per-claim disagreements
largely cancel in aggregate: human totals are 37/24/38 (Tier~1) and
13/7/10 (Tier~2), so \emph{supported} drifts by only $+4$ of 33 and
$+3$ of 10, keeping the macro estimate robust. Full details are in Appendix~\ref{app:human}.

\subsection{Ablation}
\label{sec:ablation}

A $2{\times}2{\times}2$ factorial ablation over
\textbf{self-grounding} (SG; on/off), \textbf{prompting strategy}
(baseline vs.\ calibrated), and \textbf{reasoning budget}
(minimal vs.\ low) on \texttt{gpt-5-mini} (500 articles, seed~42);
full design in Appendix~\ref{app:ablation-table}. Self-grounding
with calibration costs ${\sim}$33\% more for only $+$1.6\,pp
precision over the operating point. We adopt
\textbf{no SG\,+\,baseline\,+\,minimal}$^\dagger$ for all
large-scale runs, prioritizing reproducibility and cost.

\begin{table}[h]
\centering
\small
\setlength{\tabcolsep}{2.5pt}
\renewcommand{\arraystretch}{1.08}
\begin{tabular}{ccc rrrr}
\toprule
\textbf{SG} & \textbf{Prom.} & \textbf{Reas.} &
\textbf{Prec} & \textbf{True} & \textbf{False} & \textbf{Unv} \\
\midrule
\cmark & calib & min & \textbf{96.3} & \textbf{93.7} & 3.4 & 2.9 \\
\cmark & base  & min & 96.2 & 93.0 & 3.6 & 3.4 \\
\xmark & calib & min & 94.5 & 89.7 & 4.5 & 5.8 \\
\xmark & base  & min & $94.7^\dagger$ & 89.5 & 4.5 & 6.0 \\
\bottomrule
\end{tabular}
\caption{Ablation on \texttt{gpt-5-mini} (minimal reasoning, 500
articles, \%). SG\,=\,self-grounding. $\dagger$\,=\,operating
point.}
\label{tab:ablation_main}
\end{table}

\section{Results}
\label{sec:results}

\subsection{Entity Sanitization at Scale}
\label{sec:funnel}

GPT-5-mini produces 1{,}008{,}947 articles and 70{,}165{,}356 raw
wikilink mentions, consolidated to 12{,}479{,}792 pre-NER
candidates. Canonical deduplication keeps 2{,}293{,}089 (18.4\%);
NER accepts 1{,}654{,}284; Stage~3 accepts 1{,}120{,}843;
1{,}063{,}929 children enter the BFS queue - 1.52\% of raw mentions
(Table~\ref{tab:funnel_main}). DeepSeek and Llama show the same
funnel shape at 120K-article scale with higher raw-to-queue
survival (6.10\%, 5.07\%) because their committed indexes are
smaller (Appendix~\ref{app:funnel_full}).

\begin{table}[t]
\centering
\scriptsize
\setlength{\tabcolsep}{2pt}
\renewcommand{\arraystretch}{0.95}
\resizebox{\columnwidth}{!}{%
\begin{tabular}{lrrr}
\toprule
& \textbf{GPT-5-mini} & \textbf{DeepSeek} & \textbf{Llama} \\
\midrule
Generated articles        & 1{,}008{,}947 & 120{,}139 & 120{,}100 \\
Raw \texttt{[[links]]}    & 70{,}165{,}356 & 6{,}490{,}752 & 9{,}820{,}122 \\
Pre-NER input             & 12{,}479{,}792 & 1{,}999{,}933 & 4{,}326{,}286 \\
After canonical dedup     & 2{,}293{,}089 & 858{,}516 & 1{,}247{,}222 \\
After NER                 & 1{,}654{,}284 & 443{,}736 & 683{,}681 \\
After similarity          & 1{,}120{,}843 & 396{,}254 & 499{,}245 \\
Queue-inserted children   & 1{,}063{,}929 & 396{,}090 & 498{,}225 \\
\midrule
Raw $\rightarrow$ queue   & 1.52\% & 6.10\% & 5.07\% \\
\bottomrule
\end{tabular}%
}
\caption{Cross-model sanitization funnel. Full per-stage survival
in Appendix~\ref{app:funnel_full}.}
\label{tab:funnel_main}
\end{table}

\subsection{RQ1: Topic-Focused Analysis}
\label{sec:results-topic}

GPT-5-mini leads precision across all three topics (98.3--99.4\%),
followed by DeepSeek (95.9--97.3\%) and Llama-70B (93.7--96.5\%),
with false-claim rates compressed below 3\% everywhere
(Table~\ref{tab:topic_results}). The dominant cross-model signal is
\emph{unverifiability}: Llama exceeds GPT by 11.3\,pp on US Civil
Rights, 9.6\,pp on Ancient Babylon, and 12.0\,pp on Dutch
Colonization - the same availability-bias pattern that emerges at
scale (\S\ref{sec:results-rq2-hop}). Dutch Colonization is hardest
(Llama 35.1\% unverifiable), reflecting weaker Wikipedia coverage of
the non-Anglophone domain (W\% 60.7). Cross-model canonical entity
overlap is low (CaJ 0.17--0.22): even on the same topic, models
foreground substantially different entities, consistent with
\citet{giordano-razniewski-2026-foundations}.

\begin{table}[h]
\centering
\small
\setlength{\tabcolsep}{2.5pt}
\renewcommand{\arraystretch}{1.08}
\begin{tabular}{llccccc}
\toprule
\textbf{Topic} & \textbf{Model}
  & \textbf{Prec} & \textbf{Hall} & \textbf{Unv}
  & \textbf{W\%} & \textbf{CaJ} \\
\midrule
\multirow{3}{*}{\shortstack[l]{Ancient\\Babylon}}
  & GPT-5-mini & \textbf{99.4} & 0.5 & 18.8 & 71.0
  & \multirow{3}{*}{.19} \\
  & DeepSeek   & 97.3 & 1.8 & 24.6 & 75.7 &  \\
  & Llama-70B  & 93.7 & 2.9 & 28.4 & 79.7 &  \\
\midrule
\multirow{3}{*}{\shortstack[l]{US Civil\\Rights}}
  & GPT-5-mini & \textbf{98.3} & 1.5 & 10.8 & 78.3
  & \multirow{3}{*}{.22} \\
  & DeepSeek   & 96.9 & 2.3 & 17.4 & 83.0 &  \\
  & Llama-70B  & 96.5 & 2.4 & 22.1 & 75.3 &  \\
\midrule
\multirow{3}{*}{\shortstack[l]{Dutch\\Colon.}}
  & GPT-5-mini & \textbf{98.8} & 0.8 & 23.1 & 66.3
  & \multirow{3}{*}{.17} \\
  & DeepSeek   & 95.9 & 2.3 & 24.8 & 70.3 &  \\
  & Llama-70B  & 95.2 & 2.5 & 35.1 & 60.7 &  \\
\bottomrule
\end{tabular}
\caption{Topic-focused results (persona-averaged). \textbf{Prec}:
precision; \textbf{Hall}: false rate; \textbf{Unv}: unverifiable
rate; \textbf{W\%}: Wikipedia coverage; \textbf{CaJ}: avg.\
canonical entity Jaccard across model pairs. Factuality on
Wikipedia-covered subjects.}
\label{tab:topic_results}
\end{table}

\paragraph{Persona effects.}
Each article is scored on a 24-dimensional topic-relevant lexicon
(Appendix~\ref{app:persona}) per 1{,}000 content tokens using a
deterministic word-list classifier. On intersection-sampled subjects (30 per
model$\times$persona$\times$topic cell), paired Wilcoxon
signed-rank tests with Bonferroni correction over 648 comparisons
yield 37 significant effects in topically expected directions. On
Dutch Colonization, left-leaning uses colonized-side vocabulary
(\emph{exploitation}, \emph{plunder}, \emph{dispossession}) 5.6
hits per 1{,}000 tokens more than conservative; conservative flips
toward development framing ($+$2.3). On US Civil Rights,
left-leaning skews progressive ($+$4.6 vs.\ conservative) and
foregrounds grassroots organizers ($+$1.8); conservative centres
canonical institutions. On Ancient Babylon - the neutral
control - contested political axes collapse to non-significance.
Factual precision is unchanged ($\leq$3.6\,pp between any two
personas in the same cell): persona changes \emph{what} the article
says, not \emph{how often it is right}. Extending the analysis to
two non-contested domains (\emph{Quantum Physics},
\emph{One Piece}; Appendix~\ref{app:persona-extended}) yields only
\textbf{6} Bonferroni-significant shifts versus 37 on the primary
topics, confirming that persona framing is triggered by
domain-specific ideological affordances, not applied
indiscriminately.

\subsection{RQ2: Cross-Model at 120K}
\label{sec:results-rq2}

All three models start from the same seed under general-domain BFS.
Despite the shared seed, only 7.3\% of subjects appear in all three
corpora (Table~\ref{tab:crossmodel_main}); early wikilink choices
cascade into substantially different graphs. The robustness
analysis of \citet{giordano-razniewski-2026-foundations} shows
within-model variance matches repeated-run variance, ruling out
BFS noise - the 7.3\% overlap reflects genuine cross-model
knowledge divergence.

A consistent precision-vs-true-rate gap emerges. Precision is
preserved from shared to independent (max drop 1.1\,pp); true rate
falls sharply (GPT 88.6$\to$78.6; DeepSeek 85.8$\to$77.6; Llama
79.1$\to$64.5; max drop 14.6\,pp), tracking Wikipedia coverage
(94.0\%$\to$70.6--72.4\%). The shared intersection is the canonical
core all three models agree is encyclopedic; once we leave it for
each model's long tail, models remain right about what they assert
but external evidence increasingly cannot verify it. The ranking
GPT-5-mini\,$>$\,DeepSeek\,$>$\,Llama is stable across conditions.

\begin{table}[h]
\centering
\small
\setlength{\tabcolsep}{2pt}
\renewcommand{\arraystretch}{1.06}
\resizebox{0.9\columnwidth}{!}{%
\begin{tabular}{l rrr}
\toprule
& \textbf{GPT-5-mini} & \textbf{DeepSeek} & \textbf{Llama-70B} \\
\midrule
\multicolumn{4}{l}{\textit{Corpus (120K each)}} \\
Mean \texttt{[[links]]}/art & 70.6 & 51.5 & 81.0 \\
Mean sections      & 6.0 & 5.2 & 2.1 \\
Mean words         & 885 & 739 & 845 \\
\midrule
\multicolumn{4}{l}{\textit{Subject overlap (120K$\times$3)}} \\
Union / Intersection & \multicolumn{3}{c}{280{,}494 / 20{,}417 (7.3\%)} \\
\midrule
\multicolumn{4}{l}{\textit{Shared intersection (1K, Tier~1)}} \\
Wiki coverage  & 94.0\% & 94.0\% & 94.0\% \\
Precision      & \textbf{98.5} & 96.1 & 95.2 \\
True rate      & \textbf{88.6} & 85.8 & 79.1 \\
\midrule
\multicolumn{4}{l}{\textit{Independent (1K, Tier~1)}} \\
Wiki coverage  & 72.4\% & 70.6\% & 72.4\% \\
Precision      & \textbf{97.8} & 95.5 & 94.1 \\
True rate      & \textbf{78.6} & 77.6 & 64.5 \\
\bottomrule
\end{tabular}%
}
\caption{Cross-model comparison at 120K. Factuality on
Wikipedia-covered subjects only. Supplementary figures in
Appendix~\ref{app:crossmodel-figures}.}
\label{tab:crossmodel_main}
\end{table}

\subsubsection*{Hop-Stratified and Frontier Results}
\label{sec:results-rq2-hop}

Table~\ref{tab:hop_factuality} is the empirical backbone of the
paper. On a uniform-random 1{,}000-article sample from the
${\sim}$1M GPT-5-mini corpus, Wikipedia covers only 567 subjects
(56.7\%). On those, true rate is 68.4\%, false 1.2\%, unverifiable
30.5\%. The picture sharpens with BFS depth: true rate falls from
94.0\% at hop~1 to 56.0\% at hop~6 - a 38-point degradation. This
collapse is not driven by hallucination: false rate stays $<$2\% at
every depth and precision moves marginally (97.9$\to$96.5\%),
similar to \citet{HuGPTKB2025}.

\begin{table}[h]
\centering
\small
\setlength{\tabcolsep}{1.6pt}
\renewcommand{\arraystretch}{1.06}
\resizebox{0.95\columnwidth}{!}{%
\begin{tabular}{llrrccccc}
\toprule
\textbf{Bucket} & \textbf{Ref} & \textbf{Sampled} & \textbf{Found}
  & \textbf{Cov \%} & \textbf{Prec} & \textbf{True}
  & \textbf{False} & \textbf{Unv} \\
\midrule
hop~0   & Wiki &   1 &   1 & 100  & 100.0 & \textbf{70.0} & 0.0 & 30.0 \\
hop~1   & Wiki &  10 &  10 & 100  &  97.9 & \textbf{94.0} & 2.0 &  4.0 \\
hop~2   & Wiki & 200 & 184 &  92  &  98.7 & \textbf{84.6} & 1.1 & 14.4 \\
hop~3   & Wiki & 200 & 168 &  84  &  98.7 & \textbf{80.3} & 0.6 & 19.1 \\
hop~4   & Wiki & 200 & 134 &  67  &  97.7 & \textbf{72.6} & 1.4 & 26.0 \\
hop~5   & Wiki & 200 & 111 & 55.5 &  97.6 & \textbf{69.0} & 1.2 & 29.8 \\
hop~6   & Wiki & 200 & 102 &  51  &  96.5 & \textbf{56.0} & 1.0 & 43.0 \\
\midrule
random  & Wiki & 1000 & 567 & 56.7 & 97.1 & 68.4 & 1.2 & 30.5 \\
random  & Web  & 1000 & 779 & 77.9 & 97.7 & 56.5 & 0.7 & 42.8 \\
frontier& Web  &  433 & 311 & 71.8 & 98.3 & 57.6 & 0.6 & 41.8 \\
\bottomrule
\end{tabular}%
}
\caption{GPT-5-mini factuality by BFS depth and evidence tier.
\textbf{Sampled}: subjects attempted; \textbf{Found}: with usable
evidence; \textbf{Cov}: Found/Sampled. Frontier row restricts to
subjects absent from Wikipedia. All rates \%.}
\label{tab:hop_factuality}
\end{table}

What grows in lockstep with the true-rate drop is the
\emph{unverifiable} rate (4\%$\to$43\%), tracking the parallel
collapse in Wikipedia coverage (100\%$\to$51\%). The model is not
asserting more falsehoods at depth; evaluation cannot reach the
knowledge it does encode. This is the signature availability bias
predicts: fixed-question benchmarks miss the long tail because the
long tail is, by construction, not in the benchmark. On the web
tier, 77.9\% of random-sample subjects yield usable evidence;
among 433 frontier subjects absent from Wikipedia, 311 (71.8\%)
yield usable evidence with true rate 57.6\% and precision
98.3\% - the benchmark-vs-open-ended gap is a coverage problem,
not a hallucination problem.

\subsection{RQ3: Retrieval Trap}
\label{sec:results-rq3}

Table~\ref{tab:retrieval-trap} shows the central pattern. LLMpedia
leads Grokipedia by \textbf{+6.9\,pp} true rate against Wikipedia
and \textbf{+2.9\,pp} on the web tier. Grokipedia's false rate
(1.8\%) is more than $2\times$ LLMpedia's (0.8\%) under Wikipedia
verification. LLMpedia's unverifiable rate is also lower (13.3\%
vs.\ 19.1\% Wiki). LLMpedia achieves this at roughly half the
TF-IDF cosine (0.256 vs.\ 0.493) and much lower $n$-gram overlap,
while semantic cosine remains comparable because both systems
discuss the same subjects. This is the retrieval trap: Grokipedia
stays close to Wikipedia's surface form yet is less factually
reliable - consistent with retrieval-shaped rewriting without
robust entity disambiguation (Appendix~\ref{app:grokipedia}).
Lexical proximity to Wikipedia is not a proxy for truth.

\begin{table}[h]
\centering
\small
\setlength{\tabcolsep}{3pt}
\renewcommand{\arraystretch}{1.10}
\begin{tabular}{l cc}
\toprule
\textbf{Metric} & \textbf{LLMpedia} & \textbf{Grokipedia} \\
\midrule
\multicolumn{3}{l}{\textit{Factuality (Wiki\,/\,Web)}} \\
Precision (\%)  & \textbf{98.6\,/\,98.8} & 97.3\,/\,97.0 \\
True rate (\%)  & \textbf{86.0\,/\,76.3} & 79.1\,/\,73.4 \\
False rate (\%) & \textbf{0.8\,/\,0.7}   & 1.8\,/\,1.7 \\
Unverifiable (\%) & 13.3\,/\,23.0 & 19.1\,/\,24.9 \\
\midrule
\multicolumn{3}{l}{\textit{Similarity to Wikipedia}} \\
TF-IDF cosine   & 0.256 & 0.493 \\
Bigram overlap  & 0.090 & 0.200 \\
Trigram overlap & 0.026 & 0.079 \\
Semantic cosine & 0.773 & 0.811 \\
\midrule
Words (mean / med.) & 2{,}016\,/\,1{,}958 & 7{,}376\,/\,6{,}193 \\
\bottomrule
\end{tabular}
\caption{Retrieval-trap on 1{,}000 most-edited Wikipedia articles.
Factuality: \textbf{Tier~1\,(Wiki)\,/\,Tier~2\,(Web)}. Full
breakdown with standard deviations in
Appendix~\ref{app:retrieval-trap}.}
\label{tab:retrieval-trap}
\end{table}

\section{Conclusion}
\label{sec:conclusion}
LLMpedia materializes ${\sim}$1.3M articles from parametric memory
across three model families, audited against Wikipedia and curated web
evidence. Three findings: (i) benchmarks overstate long-form
reliability - \texttt{gpt-5-mini} reaches 68.4\% true rate (21\,pp
below MMLU), driven by unverifiability (30.5\%), not falsehood
(1.2\%); (ii) the beyond-Wikipedia frontier is hard - 43\% of subjects
lie outside Wikipedia, with 57.6\% verified under curated-web evidence;
(iii) lexical proximity to Wikipedia is not a proxy for truth - LLMpedia
beats Grokipedia at half the TF-IDF similarity. Artifacts:
\url{https://llmpedia.net}.

\section{Limitations}
\label{sec:limitations}

\paragraph{Single-pass sampling and capability conflation.}
Every article is one generation under fixed prompt and
temperature-0 decoding, surfacing a sample of latent knowledge
rather than the entirety. Self-consistency \citep{wang2022self}
could recover more but multiplies the dominant generation cost
(already ${\sim}$\$3{,}500; Appendix~\ref{app:cost}). Long-form
prose also entangles latent knowledge retrieval with composition
under context; decomposition mitigates this at the verification
step but not the generation step. Targeted single-claim elicitation
\citep{petroni2019lama,sun2024headtail} measures knowledge in
isolation but loses the modality users consume; pairing both
protocols on the same entity is a natural extension.

\paragraph{Verifier choice.}
We follow the decompose--retrieve--verify pipeline of FActScore and
VeriScore but use a general-purpose judge (\texttt{gpt-4.1-nano})
rather than a fine-tuned verifier. Their verifiers were developed
and validated at ${\sim}$6.5K-generation scale; we validate the
judge directly against human verdicts on balanced 33/33/33 samples
(\S\ref{sec:human}) and invest the additional rigor on the evidence
side (Appendix~\ref{app:evidence-scoring}). A FActScore cross-check
on a sampled article returned 89.8\% factuality, broadly consistent
with our judge's scores, though direct comparison is partial
because of the binary-vs-three-label difference. 

\paragraph{Temporal snapshot and selection bias.}
All runs were conducted between January and March 2026; Wikipedia
and web references evolve continuously, so a claim judged
unverifiable today may become verifiable or refutable later.
Separately, 28.2\% of frontier subjects yield no usable web
evidence under strict source-quality filtering, potentially
biasing Tier~2 estimates toward better-documented subjects.

\paragraph{Scale asymmetry.}
Only \texttt{gpt-5-mini} reaches ${\sim}$1M articles; the
open-weight models are constrained to ${\sim}$120K each, so the
deepest hop-stratified analysis is GPT-5-mini specific.

\paragraph{Scope of measurement.}
Our framework captures propositional factual support, not
coherence, neutrality, salience, completeness, or omission. An
article can achieve high precision while being editorially
unbalanced. Our goal is to measure \emph{how much knowledge models
surface} in the long-form modality users consume, not to certify
articles as well-rounded encyclopedic reference; the latter
requires an orthogonal evaluation we leave to future work.

\paragraph{Homonymy and privacy.}
Stage~3 LLM arbitration handles two regimes well: it merges semantic
duplicates with different surface forms
(\emph{Deutschland}/\emph{Germany}, \emph{U.S.}/\emph{USA}) by
detecting that their lead paragraphs converge, and it keeps
\emph{qualified} homonyms (\emph{Apple Inc.}\ vs.\ \emph{Apple};
\emph{Dresden, Germany} vs.\ \emph{Dresden, Maine}) separate because
their distinct surface forms yield distinct canonical keys and
distinct embeddings before arbitration even runs. The failure mode is
\emph{bare} homonymy: two genuinely distinct entities surfaced under
identical, unqualified strings (e.g., both written simply as
\emph{Dresden} - the German city and a U.S.\ locality). In that case
the surface form is the only key the canonical-dedup stage sees, and
the second arrival is treated as a re-encounter of the first; the
two senses are merged before Stage~3's lead-based arbitration is ever
consulted, even though their leads would in principle separate them.
Resolving this requires context-aware entity linking with
sense-specific identifiers, which we leave to future work.
Separately, generated articles may mention real public figures
because encyclopedic subjects include them; the pipeline is not
designed to expose private information, and any inadvertent
occurrence is filtered from releases.

%% file: sections/tables/comparisons.tex






\begin{table}[t!]
\centering
\footnotesize
\setlength{\tabcolsep}{1.6pt}
\renewcommand{\arraystretch}{0.88}

\begin{tabularx}{\columnwidth}{@{}Xcccc@{}}
\toprule
\textbf{System} &
\textbf{Param.} &
\begin{tabular}{@{}c@{}}\textbf{Fact.}\\\textbf{Full Art.}\end{tabular} &
\textbf{Serendip.} &
\textbf{Scale} \\
\midrule

\multicolumn{5}{@{}c@{}}{\textit{Benchmarks}} \\
MMLU \cite{hendrycks2021mmlu} & \cmark & \xmark & \xmark & $10^4$ \\
HLE \cite{hendrycks2025hle} & \cmark & \xmark & \xmark & $10^3$ \\

\addlinespace[1pt]
\multicolumn{5}{@{}c@{}}{\textit{Knowledge materialization}} \\
GPTKB \cite{HuGPTKB2025} & \cmark & \xmark & \cmark & $10^6$ \\
Cosmopedia \cite{benallal2024cosmopedia_blog} & \cmark & \xmark & (\cmark) & $10^7$ \\

\addlinespace[1pt]
\multicolumn{5}{@{}c@{}}{\textit{Encyclopedia generation}} \\
WikiSum \cite{Liu2018WikiSum} & \xmark & \cmark & \xmark & $10^4$ \\
STORM \cite{shao-etal-2024-assisting} & \xmark & \cmark & \xmark & $10^2$ \\
Grokipedia \cite{grokipedia} & \xmark\,(?) & \cmark & \textbf{?} & $10^6$ \\

\midrule
\textbf{LLMpedia (ours)} & \cmark & \cmark & \cmark & $10^6$ \\
\bottomrule
\end{tabularx}

\caption{Comparison across paradigms. Param.=parametric-only generation; Fact. Full Art.=factual full-article generation; Serendip.=not-preconceived knowledge discovery.}
\label{tab:comparisonrelated}
\end{table}

%% file: appendices/appendix.tex

\appendix

\section{Ablation: Full Results}
\label{app:ablation-table}

We conduct a $2{\times}2{\times}2$ factorial ablation over
\textbf{self-grounding} (on/off), \textbf{prompting strategy} (baseline
vs.\ calibrated), and \textbf{reasoning budget} (minimal vs.\ low) on
\texttt{gpt-5-mini}.
For each of the eight configurations, we generate 5K articles and evaluate a
fixed random sample of 500 articles (seed~42).
Claims are verified against full Wikipedia pages by \texttt{gpt-4.1-nano}.

\paragraph{Prompting strategy.}
Switching from baseline to calibrated yields only small and inconsistent gains.
Without self-grounding, precision changes from 94.7\% to 94.5\%; with
self-grounding, from 96.2\% to 96.3\%.
The main effect is a modest reshaping of the true--unverifiable balance, not a
large precision jump.

\paragraph{Self-grounding.}
Self-grounding improves results consistently.
Under calibrated+minimal, precision rises from 94.5\% to 96.3\% and the
unverifiable rate falls from 5.8\% to 2.9\%.
Its main benefit is reducing unverifiable claims.
The cost is substantial: approximately 33\% higher generation cost
(${\sim}$\$850 more per million articles).

\paragraph{Reasoning budget.}
Increasing the reasoning budget from minimal to low \emph{reduces} precision
in all four SG$\times$prompt settings.
Longer reasoning traces introduce extra speculative detail rather than
improving factual reliability.

\paragraph{Configuration choice.}
The best configuration is \textbf{SG\,+\,calibrated\,+\,minimal} at 96.3\%
precision and 93.7\% true rate.
We choose \textbf{no SG\,+\,baseline\,+\,minimal}$^\dagger$ for all
large-scale runs: it saves ${\sim}$\$850 per million articles while
sacrificing only 1.6\,pp precision.

\begin{table}[h]
\centering
\small
\setlength{\tabcolsep}{2.0pt}
\renewcommand{\arraystretch}{1.06}
\begin{tabular}{ccc rrrr}
\toprule
\textbf{SG} & \textbf{Prompt} & \textbf{Reas.} &
\textbf{Prec.\%} & \textbf{True\%} & \textbf{False\%} & \textbf{Unv.\%} \\
\midrule
\xmark & base  & low & 90.7 & 84.6 & 8.2 & 7.3 \\
\xmark & base  & min & $94.7^\dagger$ & 89.5 & 4.5 & 6.0 \\
\xmark & calib & low & 92.1 & 85.4 & 7.0 & 7.7 \\
\xmark & calib & min & 94.5 & 89.7 & 4.5 & 5.8 \\
\cmark & base  & low & 93.9 & 89.3 & 5.5 & 5.2 \\
\cmark & base  & min & 96.2 & 93.0 & 3.6 & 3.4 \\
\cmark & calib & low & 95.3 & 92.2 & 4.3 & 3.6 \\
\cmark & calib & min & \textbf{96.3} & \textbf{93.7} & \textbf{3.4} & \textbf{2.9} \\
\bottomrule
\end{tabular}
\caption{Full $2^3$ ablation on \texttt{gpt-5-mini} (500 articles, seed~42).
SG = self-grounding. $\dagger$: selected operating point.}
\label{tab:ablation_full}
\end{table}

\section{Models and Costs}
\label{app:cost}

\subsection{Model Configuration}

All \texttt{gpt-5-mini} runs use the OpenAI Batch API with temperature~0 and
minimal reasoning effort, receiving a 50\% cost reduction.
A fraction of NER and similarity calls were executed via the online API due
to batch queue overflow during peak expansion waves.
\texttt{Llama-3.3-70B-Instruct} and \texttt{DeepSeek-V3-0324} are served
on-premise (4$\times$A100 and 16$\times$H100 respectively);
their marginal cost is GPU-hours not monetized here.

\begin{table}[h]
\centering
\scriptsize
\setlength{\tabcolsep}{3pt}
\renewcommand{\arraystretch}{1.08}
\begin{tabular}{l l l}
\toprule
\textbf{Model} & \textbf{Execution} & \textbf{Role} \\
\midrule
\texttt{gpt-5-mini}             & OpenAI Batch API        & generation, NER, sim \\
\texttt{Llama-3.3-70B}          & 4$\times$A100 (on-prem) & generation, NER, sim \\
\texttt{DeepSeek-V3-0324}       & 16$\times$H100 (on-prem) & generation, NER, sim \\
\texttt{text-embedding-3-small} & OpenAI Batch API        & dedup embeddings \\
\texttt{gpt-4.1-nano}           & OpenAI API              & evaluation judge \\
\bottomrule
\end{tabular}
\caption{Models and execution modes across all experimental conditions.}
\label{tab:model_configs}
\end{table}

\subsection{Aggregate Cost}

Total project expenditure across all API-billed components-including
large-scale generation (${\sim}$1M articles for \texttt{gpt-5-mini}), the
full $2^3$ ablation, 27 topic-focused configurations, the retrieval-trap
benchmark, automatic evaluation, embedding costs, and preliminary development
runs-is approximately \$3{,}500.
Open-weight models contribute GPU-hours but no API cost.

\section{Retrieval-Trap Benchmark: Full Details}
\label{app:retrieval-trap}

\subsection{Experimental Design}

The retrieval-trap benchmark asks whether parametric generation can remain
structurally independent on subjects where a retrieval-shaped system has
maximal opportunity to track Wikipedia closely.

\paragraph{Title selection.}
Following~\citet{yasseri2025similar}, we collected 1{,}000 of the most-edited
English Wikipedia titles that were also present on Grokipedia.
This deliberately targets high-visibility, heavily maintained subjects where
Wikipedia coverage is rich and a retrieval-centered system should have every
incentive to converge toward Wikipedia's wording and structure.

\paragraph{Generation.}
LLMpedia generated all 1{,}000 articles using \texttt{gpt-5-mini} in the
baseline large-scale configuration (no self-grounding, minimal reasoning).
Grokipedia articles were fetched from the live site (February 24, 2026).
Wikipedia references were retrieved through the MediaWiki API with redirect
resolution.

\paragraph{Evaluation.}
For each title, we extracted up to 10 atomic claims from both systems
and verified them against two evidence tiers using the verification
prompt described in Appendix~\ref{app:eval-prompt}, which enforces the
absence-of-evidence asymmetry (silence $\to$ insufficient,
contradiction $\to$ refuted).

\subsection{Detailed Results}

\begin{table}[t]
\centering
\scriptsize
\setlength{\tabcolsep}{3pt}
\renewcommand{\arraystretch}{1.10}
\begin{tabular}{l cc}
\toprule
\textbf{Metric} & \textbf{LLMpedia} & \textbf{Grokipedia} \\
\midrule
\multicolumn{3}{l}{\textit{Factuality (Wiki\,/\,Web)}} \\
Precision (\%)    & \textbf{98.6$\pm$8.3\,/\,98.8$\pm$6.7}
                  & 97.3$\pm$10.3\,/\,97.0$\pm$9.6 \\
True rate (\%)    & \textbf{86.0$\pm$28.9\,/\,76.3$\pm$33.4}
                  & 79.1$\pm$31.8\,/\,73.4$\pm$31.4 \\
False rate (\%)   & \textbf{0.8$\pm$3.7\,/\,0.7$\pm$4.3}
                  & 1.8$\pm$6.4\,/\,1.7$\pm$4.7 \\
Unverifiable (\%) & 13.3$\pm$28.6\,/\,23.0$\pm$33.2
                  & 19.1$\pm$31.3\,/\,24.9$\pm$31.2 \\
\midrule
\multicolumn{3}{l}{\textit{Similarity to Wikipedia}} \\
TF-IDF cosine      & 0.256$\pm$0.120 & 0.493$\pm$0.169 \\
Bigram overlap     & 0.090$\pm$0.041 & 0.200$\pm$0.082 \\
Trigram overlap    & 0.026$\pm$0.017 & 0.079$\pm$0.053 \\
Semantic cosine    & 0.773$\pm$0.079 & 0.811$\pm$0.059 \\
\midrule
Mean / Median words & 2{,}016\,/\,1{,}958 & 7{,}376\,/\,6{,}193 \\
Wikipedia ref. mean & \multicolumn{2}{c}{4{,}105\,/\,2{,}434} \\
\bottomrule
\end{tabular}
\caption{Full retrieval-trap results on 1{,}000 shared titles.
Mean $\pm$ std macro-averaged per article.
All verdicts under the evaluation prompt described in
Appendix~\ref{app:eval-prompt}.}
\label{tab:retrieval_trap_detail}
\end{table}

\paragraph{Interpretation.}
Grokipedia stays much closer to Wikipedia at the lexical level (TF-IDF 0.493
vs.\ 0.256) yet achieves lower true rate ($-$6.9\,pp Wiki, $-$2.9\,pp Web)
and a false rate more than $2\times$ higher.
LLMpedia's unverifiable rate is also substantially lower (13.3\% vs.\ 19.1\%
Wiki), meaning it makes fewer claims that evidence cannot resolve.
This is the retrieval trap: lexical proximity does not imply factual
superiority, and may reflect retrieval-shaped rewriting without robust
disambiguation.

\section{Evaluation Prompt}
\label{app:eval-prompt}

\subsection{Claim Extraction Prompt}

\vspace{4pt}
\noindent\colorbox{orange!8}{\parbox{0.95\linewidth}{\small
\textbf{System:}
You are a factuality evaluator.
Extract up to \{$K_\mathrm{max}$\} distinct, atomic, verifiable factual
claims from an article about ``\{subject\}''.
Each claim must be ONE atomic fact (one predicate, one relationship),
self-contained (include the subject name), and verifiable against a
reference encyclopedia.
Skip opinions, hedged language, and vague statements.
Output ONLY valid JSON: \texttt{\{"claims": ["claim1", "claim2", ...]\}}
\\[3pt]
\textbf{User:}
Subject: \{subject\} \quad
Article text: \{article\_text\}

Return JSON with up to \{$K_\mathrm{max}$\} claims.
}}

\subsection{Factuality Verification Prompt}

This is the verification prompt used for all results reported in this
paper.
The absence-of-evidence asymmetry is encoded as a hard rule with three
worked examples; without this rule a naive judge conflates
``evidence does not mention X'' with ``evidence refutes X'',
mechanically inflating the false-claim rate and obscuring the
distinction between knowledge gaps and factual errors that is central
to our analysis.

\vspace{4pt}
\noindent\colorbox{red!6}{\parbox{0.95\linewidth}{\small
\textbf{System:}
You are a careful fact-checker.
Verify \{$n$\} claims about ``\{subject\}'' against the evidence below.

Use exactly one of these three verdicts per claim:

\textbf{``supported''}: the evidence explicitly states the claim, or
the claim follows directly from what the evidence says.

\textbf{``refuted''}: the evidence explicitly states something that
CONTRADICTS the claim. There must be a positive statement in the
evidence that is logically incompatible with the claim
(e.g.\ claim says ``founded in 1850'' but evidence says ``founded in
1920'').

\textbf{``insufficient''}: the evidence does not address the claim,
addresses it only partially, or is silent on the topic.
THIS INCLUDES every case where the evidence simply does not mention the
subject of the claim.

\textbf{CRITICAL RULE:} Absence of evidence is NOT evidence of absence.
If the evidence does not mention X, the verdict for any claim about X
is ``insufficient'', NEVER ``refuted''.
Mark ``refuted'' ONLY when the evidence contains an explicit positive
statement that contradicts the claim.

\textbf{EXAMPLES:}

Claim: ``The school was founded in 1920.''\\
Evidence: ``The school was founded in 1850.''\\
Verdict: \emph{refuted} (evidence states a different founding year)

Claim: ``The school was founded in 1920.''\\
Evidence: ``The school is located in Paris and offers art programs.''\\
Verdict: \emph{insufficient} (evidence does not address the founding
year)

Claim: ``The curriculum was influenced by the Bauhaus.''\\
Evidence: ``There is no mention of Bauhaus or curriculum influences.''\\
Verdict: \emph{insufficient} (absence of mention is not contradiction)

When verdict is \textbf{``refuted''}, additionally output an
\texttt{error\_type} field with one of these values:
\texttt{temporal} (wrong date/year/era),
\texttt{numerical} (wrong quantity),
\texttt{spatial\_geographic} (wrong location),
\texttt{person\_attribution} (wrong person credited),
\texttt{entity\_relation} (wrong relationship),
\texttt{causal\_motivational} (wrong cause/consequence),
\texttt{categorical\_taxonomic} (wrong classification),
\texttt{definitional\_property} (wrong intrinsic property),
\texttt{fabricated\_entity} (entity does not exist),
\texttt{partial\_truth\_overgeneralization} (partly correct but overstated),
\texttt{terminological\_naming} (wrong name/label),
\texttt{other}.
Set \texttt{error\_type} to null for supported or insufficient.

Output ONLY valid JSON:\\
\texttt{\{"verdicts":[}\\
\texttt{\quad\{"idx":1,"verdict":"...",
"error\_type":"...",}\\
\texttt{\quad "confidence":0.0--1.0,
"explanation":"..."\}}\\
\texttt{]\}}
\\[3pt]
\textbf{User:}
Claims:\\
\{claims\_block\}

Evidence source: \{evidence\_source\}\\
\{evidence\_snippets\}

Return JSON with verdicts.
}}

\paragraph{Parser safety net.}
The parser additionally coerces any verdict labeled \emph{refuted} whose
explanation contains silence-indicating phrases (\emph{does not mention},
\emph{not specified}, \emph{there is no indication}) into
\emph{insufficient}, logging each coercion.
This affected 2.3\% of initially refuted verdicts across our runs.

\section{Human Validation Protocol}
\label{app:human}

\subsection{Design}

We ran a two-stage human validation to audit the LLM judge against
expert human verdicts.

\paragraph{Sampling.}
For each evaluation tier, we drew a balanced stratified sample from
the judge's verdict distribution.
\textbf{Tier~1 (Wikipedia evidence)}: 99 (claim, evidence, judge
verdict) triples, exactly balanced as
\textbf{33 supported / 33 refuted / 33 insufficient}, drawn from the
random subsample of the 1M GPT-5-mini corpus.
\textbf{Tier~2 (curated web evidence)}: 30 triples, balanced as
\textbf{10 / 10 / 10}, drawn from the frontier-Web subset.

\paragraph{Annotation.}
For each (claim, evidence, verdict) triple, a human expert was shown
the full evidence snippets used by the judge and annotated:
(a) do you agree with the judge verdict? and
(b) if not, which of the three classes do you believe is correct?

\subsection{Results}

\begin{table}[h]
\centering
\small
\setlength{\tabcolsep}{6pt}
\renewcommand{\arraystretch}{1.08}
\begin{tabular}{l cc}
\toprule
\textbf{Class} & \textbf{Tier~1 (Wiki)} & \textbf{Tier~2 (Web)} \\
\midrule
Supported    & 87.9\% & 93.3\% \\
Refuted      & 81.8\% & 70.0\% \\
Insufficient & 93.9\% & 90.0\% \\
\midrule
Overall      & 87.9\% & 84.4\% \\
\bottomrule
\end{tabular}
\caption{Human--judge per-class agreement.
Agreement on the contribution-bearing classes
(\emph{supported} drives true rate; \emph{insufficient} drives
unverifiable rate) ranges \textbf{87.9--93.9\%} across both tiers, with a per-class
     floor of 87.9\%.}
\label{tab:human_agreement}
\end{table}

\paragraph{Aggregate conservation.}
Per-claim agreement is imperfect, but disagreements largely cancel in
aggregate.
From a balanced judge sample of 33/33/33 (Tier~1), human class totals
are 37/24/38; from a balanced 10/10/10 (Tier~2), totals are 13/7/10.
The \emph{supported} total-which determines the headline true
rate-drifts by only $+4$ of 33 (Tier~1) and $+3$ of 10 (Tier~2),
so the macro estimate is robust to per-claim variation.
This conservation is what allows the 68.4\% headline true rate and the 30.5\% unverifiable
     rate to be reported with confidence despite
imperfect per-claim agreement.

\paragraph{Discussion.}
Refuted carries the lowest agreement, which is expected: the refuted
class is the one where human and judge most often disagree about
whether the evidence positively contradicts the claim or merely fails
to mention it.
The verification prompt (Appendix~\ref{app:eval-prompt}) addresses
this by stating the absence-of-evidence asymmetry as a hard rule and
including three worked examples; the parser additionally coerces any
verdict labelled refuted whose explanation contains silence-indicating
phrases (\emph{does not mention}, \emph{not specified},
\emph{there is no indication}) into insufficient, which affected 2.3\%
of initially refuted verdicts.
\section{Web Evidence Pipeline}
\label{app:evidence-scoring}

The pipeline auto-detects available search backends (Valyu $\to$ Serper $\to$
Brave $\to$ DuckDuckGo); Valyu was the primary backend in all large-scale
experiments.
Each URL receives a quality score on a 0--100 scale; only sources with score
$\geq 60$ are eligible for full-page fetching.
Tables~\ref{tab:explicit_domain_scores_a}--\ref{tab:explicit_domain_scores_b}
list the \textbf{133 explicitly scored domains}.
Unknown \texttt{.edu}/\texttt{.gov}/\texttt{.org} domains default to 70;
unknown HTTPS domains default to 35.

\begin{table*}[t]
\centering
\scriptsize
\setlength{\tabcolsep}{4pt}
\renewcommand{\arraystretch}{1.08}
\begin{tabular}{p{0.20\linewidth}p{0.74\linewidth}}
\toprule
\textbf{Category} & \textbf{Blocked root domains} \\
\midrule
Circular / Wikimedia
& \texttt{wikipedia.org}, \texttt{en.wikipedia.org}, \texttt{en.m.wikipedia.org},
  \texttt{wikidata.org}, \texttt{www.wikidata.org}, \texttt{wikimedia.org} \\
Search engines
& \texttt{google.com}, \texttt{bing.com}, \texttt{duckduckgo.com}, \texttt{yahoo.com} \\
Social / video
& \texttt{facebook.com}, \texttt{instagram.com}, \texttt{twitter.com}, \texttt{x.com},
  \texttt{tiktok.com}, \texttt{reddit.com}, \texttt{quora.com}, \texttt{youtube.com},
  \texttt{pinterest.com} \\
E-commerce
& \texttt{amazon.com}, \texttt{ebay.com} \\
AI answer engines
& \texttt{perplexity.ai} \\
\bottomrule
\end{tabular}
\caption{Root-domain exclusions. Exclusion uses root-domain matching, so
subdomains of blocked roots are also excluded.}
\label{tab:blocked_domains}
\end{table*}

\begin{table*}[t]
\centering
\scriptsize
\setlength{\tabcolsep}{3pt}
\renewcommand{\arraystretch}{1.06}
\begin{tabular}{@{}p{0.16\linewidth}p{0.18\linewidth}r p{0.18\linewidth}r p{0.18\linewidth}r@{}}
\toprule
\textbf{Category} & \textbf{Domain} & \textbf{Score} & \textbf{Domain} & \textbf{Score} & \textbf{Domain} & \textbf{Score} \\
\midrule
\textit{Encyclopedias}
& \texttt{britannica.com} & 100 & \texttt{worldhistory.org} & 90 & \texttt{encyclopedia.com} & 85 \\
& \texttt{scholarpedia.org} & 88 & \texttt{plato.stanford.edu} & 95 & \texttt{iep.utm.edu} & 88 \\
& \texttt{newworldencyclopedia.org} & 75 & & & & \\
\addlinespace[2pt]
\textit{Government}
& \texttt{loc.gov} & 97 & \texttt{archives.gov} & 97 & \texttt{congress.gov} & 95 \\
& \texttt{usa.gov} & 93 & \texttt{cia.gov} & 90 & \texttt{state.gov} & 90 \\
& \texttt{whitehouse.gov} & 90 & \texttt{nasa.gov} & 95 & \texttt{nih.gov} & 94 \\
& \texttt{cdc.gov} & 93 & \texttt{fda.gov} & 90 & \texttt{epa.gov} & 89 \\
& \texttt{noaa.gov} & 90 & \texttt{usgs.gov} & 90 & \texttt{nps.gov} & 85 \\
& \texttt{si.edu} & 92 & \texttt{parliament.uk} & 88 & \texttt{gov.uk} & 87 \\
& \texttt{europarl.europa.eu} & 85 & \texttt{un.org} & 88 & \texttt{who.int} & 90 \\
& \texttt{worldbank.org} & 88 & \texttt{imf.org} & 87 & & \\
\addlinespace[2pt]
\textit{Wire services}
& \texttt{reuters.com} & 94 & \texttt{apnews.com} & 94 & \texttt{bbc.com} & 93 \\
& \texttt{bbc.co.uk} & 93 & \texttt{nytimes.com} & 92 & \texttt{washingtonpost.com} & 90 \\
& \texttt{theguardian.com} & 91 & \texttt{economist.com} & 90 & \texttt{ft.com} & 89 \\
& \texttt{wsj.com} & 89 & \texttt{theatlantic.com} & 87 & \texttt{newyorker.com} & 87 \\
& \texttt{npr.org} & 88 & \texttt{pbs.org} & 88 & \texttt{aljazeera.com} & 85 \\
& \texttt{dw.com} & 84 & \texttt{france24.com} & 83 & & \\
\addlinespace[2pt]
\textit{Academic journals}
& \texttt{nature.com} & 95 & \texttt{science.org} & 95 & \texttt{sciencedirect.com} & 90 \\
& \texttt{springer.com} & 89 & \texttt{link.springer.com} & 89 & \texttt{wiley.com} & 88 \\
& \texttt{onlinelibrary.wiley.com} & 88 & \texttt{tandfonline.com} & 87 & \texttt{cell.com} & 93 \\
& \texttt{thelancet.com} & 93 & \texttt{bmj.com} & 92 & \texttt{nejm.org} & 94 \\
& \texttt{pnas.org} & 92 & \texttt{ncbi.nlm.nih.gov} & 92 & \texttt{pubmed.ncbi.nlm.nih.gov} & 92 \\
& \texttt{jstor.org} & 90 & \texttt{arxiv.org} & 85 & \texttt{ssrn.com} & 82 \\
& \texttt{researchgate.net} & 75 & \texttt{scholar.google.com} & 80$^*$ & \texttt{doaj.org} & 80 \\
& \texttt{semanticscholar.org} & 82 & \texttt{ieee.org} & 89 & \texttt{ieeexplore.ieee.org} & 89 \\
& \texttt{acm.org} & 88 & \texttt{dl.acm.org} & 88 & & \\
\bottomrule
\end{tabular}
\caption{Explicitly scored domains (Part~I).
$^*$\texttt{scholar.google.com} is scored but excluded at runtime because
\texttt{google.com} is blocked by root-domain matching.}
\label{tab:explicit_domain_scores_a}
\end{table*}

\begin{table*}[t]
\centering
\scriptsize
\setlength{\tabcolsep}{3pt}
\renewcommand{\arraystretch}{1.06}
\begin{tabular}{@{}p{0.16\linewidth}p{0.18\linewidth}r p{0.18\linewidth}r p{0.18\linewidth}r@{}}
\toprule
\textbf{Category} & \textbf{Domain} & \textbf{Score} & \textbf{Domain} & \textbf{Score} & \textbf{Domain} & \textbf{Score} \\
\midrule
\textit{Museums / Libraries}
& \texttt{smithsonianmag.com} & 87 & \texttt{nationalgeographic.com} & 86 & \texttt{metmuseum.org} & 90 \\
& \texttt{moma.org} & 87 & \texttt{nga.gov} & 88 & \texttt{bl.uk} & 90 \\
& \texttt{bnf.fr} & 88 & \texttt{dpla.org} & 83 & \texttt{europeana.eu} & 83 \\
\addlinespace[2pt]
\textit{History / Biography}
& \texttt{history.com} & 80 & \texttt{biography.com} & 75 & \texttt{oxforddnb.com} & 90 \\
& \texttt{anb.org} & 88 & \texttt{historytoday.com} & 78 & \texttt{historyextra.com} & 77 \\
\addlinespace[2pt]
\textit{Science / Technology}
& \texttt{scientificamerican.com} & 86 & \texttt{newscientist.com} & 83 & \texttt{livescience.com} & 75 \\
& \texttt{space.com} & 76 & \texttt{phys.org} & 78 & \texttt{sciencenews.org} & 82 \\
& \texttt{quantamagazine.org} & 88 & \texttt{arstechnica.com} & 80 & \texttt{spectrum.ieee.org} & 84 \\
& \texttt{technologyreview.com} & 84 & & & & \\
\addlinespace[2pt]
\textit{Universities}
& \texttt{mit.edu} & 93 & \texttt{stanford.edu} & 93 & \texttt{harvard.edu} & 93 \\
& \texttt{ox.ac.uk} & 92 & \texttt{cam.ac.uk} & 92 & \texttt{berkeley.edu} & 91 \\
& \texttt{caltech.edu} & 91 & \texttt{yale.edu} & 91 & \texttt{princeton.edu} & 91 \\
& \texttt{columbia.edu} & 90 & \texttt{uchicago.edu} & 90 & \texttt{cornell.edu} & 89 \\
& \texttt{cmu.edu} & 89 & \texttt{ethz.ch} & 89 & \texttt{mpg.de} & 89 \\
& \texttt{khanacademy.org} & 80 & & & & \\
\addlinespace[2pt]
\textit{Fact-checking / Data}
& \texttt{snopes.com} & 82 & \texttt{factcheck.org} & 84 & \texttt{politifact.com} & 80 \\
& \texttt{ourworldindata.org} & 88 & \texttt{statista.com} & 78 & \texttt{data.gov} & 88 \\
& \texttt{census.gov} & 90 & \texttt{bls.gov} & 89 & \texttt{bea.gov} & 88 \\
\addlinespace[2pt]
\textit{Legal}
& \texttt{law.cornell.edu} & 90 & \texttt{supremecourt.gov} & 92 & \texttt{courtlistener.com} & 80 \\
& \texttt{oyez.org} & 82 & & & & \\
\addlinespace[2pt]
\textit{Medical / Health}
& \texttt{mayoclinic.org} & 88 & \texttt{clevelandclinic.org} & 85 & \texttt{webmd.com} & 72 \\
& \texttt{medlineplus.gov} & 90 & \texttt{hopkinsmedicine.org} & 87 & \texttt{uptodate.com} & 88 \\
\bottomrule
\end{tabular}
\caption{Explicitly scored domains (Part~II). Together with
Table~\ref{tab:explicit_domain_scores_a}, this yields \textbf{133} explicitly
scored domains.}
\label{tab:explicit_domain_scores_b}
\end{table*}

Pages shorter than 200 characters are flagged as unusable.
CAPTCHA challenges, paywalls, and JavaScript-only responses are excluded.
Up to three candidate URLs are attempted in descending quality-score order,
falling back to the search snippet when a full-page fetch is blocked.

\subsection{Frontier Results Summary}

Of 433 Wikipedia-absent subjects in the 1{,}000-article random sample,
311 (71.8\%) yielded usable web evidence.
Precision 98.3\%, true rate 57.6\%, false rate 0.6\%, unverifiable 41.8\%.
The remaining 28.2\% are excluded from Tier~2 frontier statistics-a
source-selection bias discussed in \S\ref{sec:limitations}.

\section{Pipeline Implementation}
\label{app:implementation}

\subsection{System Configuration and Design Decisions}
\label{app:settings}

\paragraph{Execution modes.}
LLMPedia supports two execution modes that trade latency against cost.
In \emph{online mode}, API calls are issued concurrently across a multi-threaded worker pool, with all five pipeline stages running in parallel.
A global concurrency cap is enforced across all stages simultaneously to prevent rate-limit exhaustion.
In \emph{batch mode}, requests are submitted to the OpenAI Batch API, which processes them asynchronously at a 50\% cost reduction with a completion window of up to 24 hours.
The two modes are architecturally equivalent and produce identical outputs; batch mode was used in our main experiments for cost efficiency, while online mode is reserved for small-scale ablations where turnaround time matters.

\paragraph{Model configuration and stage independence.}
Each pipeline stage-self-grounding, outline generation, article elicitation, NER filtering, and similarity arbitration-can be assigned an independent language model, temperature, and token budget.
In practice, all stages within a given run share a single model unless explicitly differentiated.
The one principled exception is the embedding model used for similarity deduplication, which is always configured independently: embedding and completion models serve fundamentally different functions and conflating their configuration would silently degrade deduplication quality.

\paragraph{BFS depth and article budget.}
Expansion depth and total article count are both configurable hard caps.
In \textbf{topic-focused runs} (\S\ref{sec:results-topic}), depth is capped at two hops from the thematic seed to maintain topical coherence.
In \textbf{general-domain expansion} (\S\ref{sec:results-rq2}), no depth cap is imposed: GPT-5-mini reaches hop~6 over ${\sim}$1M articles
     (1{,}008{,}947 generated), with 95.9\% of filtered-in subjects at
     hop~4 or beyond (Table~\ref{tab:funnel_coverage_hop}).
In both modes, the seed article at hop~0 is generated but excluded from evaluation metrics to avoid confounding the model's generation behaviour with the trivially known anchor entity.

\paragraph{Similarity threshold and arbitration.}
The cosine similarity threshold above which two entity embeddings trigger a deduplication check is set to 0.90.
This value is conservative: surface-form variants of the same entity (e.g.\ \textit{United States} and \textit{USA}) consistently fall above it, while genuinely distinct but thematically related entities consistently fall below it.
When the threshold is exceeded, a secondary LLM arbitration step is invoked rather than rejecting automatically.
This two-stage design prevents both false positives and false negatives.

\paragraph{NER confidence and elicitation filtering.}
Under the calibrated prompt variants, both the NER stage and the elicitation stage assign explicit confidence scores.
Candidates below the NER confidence threshold are rejected before reaching the similarity stage.
Rejection at either stage is non-permanent: if a future parent article re-proposes the same entity with sufficient confidence, it re-enters the pipeline from the beginning.

\paragraph{Persona.}
Each pipeline stage accepts a persona specification injected at the system level.
Three personas are provided: scientifically neutral, left-leaning, and conservative.
All three share identical structural prompts; only the framing and evaluative language differ.

\paragraph{Fault tolerance and strict-gate semantics.}
Failed API calls are retried with exponential back-off up to a configurable maximum.
After all retries are exhausted, a strict-gate policy applies: a stage that cannot produce a valid output causes the corresponding entity to be \emph{dropped} rather than passed through with a degraded result.
For NER, a parse failure produces no candidates rather than passing all candidates unchecked.
For similarity arbitration, an API failure causes the candidate to be treated as a duplicate and rejected.
This conservative policy introduces a small downward bias on recall but eliminates the risk of contaminating the corpus with candidates that bypassed quality controls.

\subsection{Deduplication Correctness Under Parallel Execution}
\label{app:dedup-guarantee}

LLMPedia maintains four independent parallel queues: an \emph{elicitation queue}, a \emph{NER queue}, a \emph{similarity queue}, and the \emph{canon queue}.
Workers on each queue operate concurrently and independently.
This parallelism creates a structural race: multiple NER workers can extract and accept the same entity from different parent articles within the same BFS wave, before any of them reaches the similarity worker.
Figure~\ref{fig:dedup-race} shows a concrete instance and its resolution.

\begin{figure}[h]
\centering
\scriptsize
\colorbox{gray!10}{\parbox{0.95\linewidth}{
\textbf{BFS Wave $t$ - three concurrent NER workers}\\[4pt]
\colorbox{blue!10}{\parbox{0.29\linewidth}{
\texttt{NER Worker 1}\\
Parent: \textit{Copenhagen}\\[2pt]
Extracts:\\
\texttt{[[Niels Bohr]]}\\
\texttt{[[Quantum Mechanics]]}\\[2pt]
Checks \textsc{CanonKeys}:\\
\textit{Niels Bohr} $\notin$ keys \cmark\\
$\Rightarrow$ sends to SIM
}}\hfill
\colorbox{blue!10}{\parbox{0.29\linewidth}{
\texttt{NER Worker 2}\\
Parent: \textit{Atomic Model}\\[2pt]
Extracts:\\
\texttt{[[Niels Bohr]]}\\
\texttt{[[Ernest Rutherford]]}\\[2pt]
Checks \textsc{CanonKeys}:\\
\textit{Niels Bohr} $\notin$ keys \cmark\\
$\Rightarrow$ sends to SIM
}}\hfill
\colorbox{blue!10}{\parbox{0.29\linewidth}{
\texttt{NER Worker 3}\\
Parent: \textit{Nobel Prize}\\[2pt]
Extracts:\\
\texttt{[[Niels Bohr]]}\\
\texttt{[[Werner Heisenberg]]}\\[2pt]
Checks \textsc{CanonKeys}:\\
\textit{Niels Bohr} $\notin$ keys \cmark\\
$\Rightarrow$ sends to SIM
}}\\[6pt]
\textbf{Similarity Worker receives batch for wave $t$}\\[3pt]
\colorbox{orange!15}{\parbox{0.95\linewidth}{
Batch: \{\textit{Niels Bohr} $\times$3, \textit{Quantum Mechanics},
\textit{Ernest Rutherford}, \textit{Werner Heisenberg}, \ldots\}\\[3pt]
\textbf{Step 1 - within-wave dedup:} collapse identical canonical keys
$\Rightarrow$ \textit{Niels Bohr} reduced to \emph{one} proposal\\[2pt]
\textbf{Step 2 - embedding-index check:} cosine against all previously
committed entities; \textit{Niels Bohr} not present $\Rightarrow$ passes\\[2pt]
\textbf{Step 3 - atomic commit:} \textit{Niels Bohr} added to
\textsc{CanonQueue} \textbf{and} \textsc{CanonKeys} updated in one
operation
}}\\[6pt]
\colorbox{green!12}{\parbox{0.44\linewidth}{
\textbf{CanonQueue (main article queue)}\\
\textit{Quantum Mechanics} \cmark\\
\textit{Ernest Rutherford} \cmark\\
\textit{Niels Bohr} \cmark\ \ \textit{(exactly once)}\\
\textit{Werner Heisenberg} \cmark
}}\hfill
\colorbox{red!10}{\parbox{0.44\linewidth}{
\textbf{Silently dropped}\\
\textit{Niels Bohr} (Worker 2 copy) \xmark\\
\textit{Niels Bohr} (Worker 3 copy) \xmark\\[2pt]
\small\textit{Not data loss-design property.}
}}
}}
\caption{Three concurrent NER workers independently accept \texttt{[[Niels Bohr]]} from different parent articles. Within-wave deduplication collapses the three proposals to one before any commit.}
\label{fig:dedup-race}
\end{figure}

\paragraph{The main article queue is structurally duplicate-free.}
\textsc{CanonQueue} is the \emph{only} channel through which an entity can reach article generation.
An entity enters \textsc{CanonQueue} if and only if it simultaneously passes Stage~3 similarity \emph{and} is absent from \textsc{CanonKeys}; both operations execute atomically.

\paragraph{Why commitment happens after similarity, not after NER.}
A candidate is registered in \textsc{CanonKeys} only after it passes Stage~3.
If commitment happened at NER acceptance, an entity later rejected by similarity would be permanently suppressed even though it was never materialised.
Deferring commitment keeps future parent articles free to re-propose the same entity.

\paragraph{Within-batch collisions.}
A single parent article may produce the same wikilink in two surface forms that canonicalize identically.
Within-wave cross-candidate deduplication handles this before any index update.

\paragraph{Correctness guarantee.}
By induction on BFS waves, the embedding index at the start of wave $t$ contains exactly the entities committed in waves $0,\ldots,t{-}1$.
The main article queue therefore contains no duplicate entities at any point during or after execution.


\section{Entity Sanitization Funnel: Full Breakdown}
\label{app:funnel_full}
 
This appendix presents the full sanitization funnel for all three
models: cross-model comparison figures
(Appendix~\ref{app:funnel_combined}), per-model deep dives
(Appendices~\ref{app:funnel_gpt}--\ref{app:funnel_llama}), pipeline
coverage statistics distinguishing the filtered-in corpus from the
expanded-out subset (Appendix~\ref{app:funnel_coverage}), the alias
absorption introduced by Stage~3 disambiguation
(Appendix~\ref{app:funnel_aliases}), and the small residual
queue-insert race rate that the parallel pipeline incurs
(Appendix~\ref{app:funnel_race}).
 
\begin{table}[h]
\centering
\small
\setlength{\tabcolsep}{3pt}
\renewcommand{\arraystretch}{1.08}
\begin{tabular}{l rrr}
\toprule
& \textbf{GPT-5-mini} & \textbf{DeepSeek} & \textbf{Llama} \\
\midrule
Generated articles     & 1{,}009K & 120K & 120K \\
Raw \texttt{[[links]]} & 70.2M & 6.5M & 9.8M \\
Pre-NER candidates     & 12.48M & 2.00M & 4.33M \\
After canonical dedup  & 2.29M & 859K & 1.25M \\
After NER              & 1.65M & 444K & 684K \\
After similarity       & 1.12M & 396K & 499K \\
New queued subjects    & 1.06M & 396K & 498K \\
\midrule
Raw $\to$ pre-NER & 17.8\% & 30.8\% & 44.1\% \\
Canonical dedup survival & 18.4\% & 42.9\% & 28.8\% \\
NER survival     & 72.1\% & 51.7\% & 54.8\% \\
Sim.\ survival   & 64.7\% & 87.5\% & 66.5\% \\
Raw $\to$ queue  & 1.52\% & 6.10\% & 5.07\% \\
\bottomrule
\end{tabular}
\caption{Full funnel with per-stage survival rates across all three
models. Raw $\to$ pre-NER reflects article-level candidate
construction and surface-form consolidation before any LLM stage runs.
Canonical deduplication is dominated by previously committed entities:
as the committed index grows, canonical survival falls and canonical
deduplication loss increases.}
\label{tab:funnel_summary_table_app}
\end{table}
 
\subsection{Cross-Model Comparison}
\label{app:funnel_combined}
 
\begin{figure}[h]
  \centering
  \includegraphics[width=0.95\linewidth]{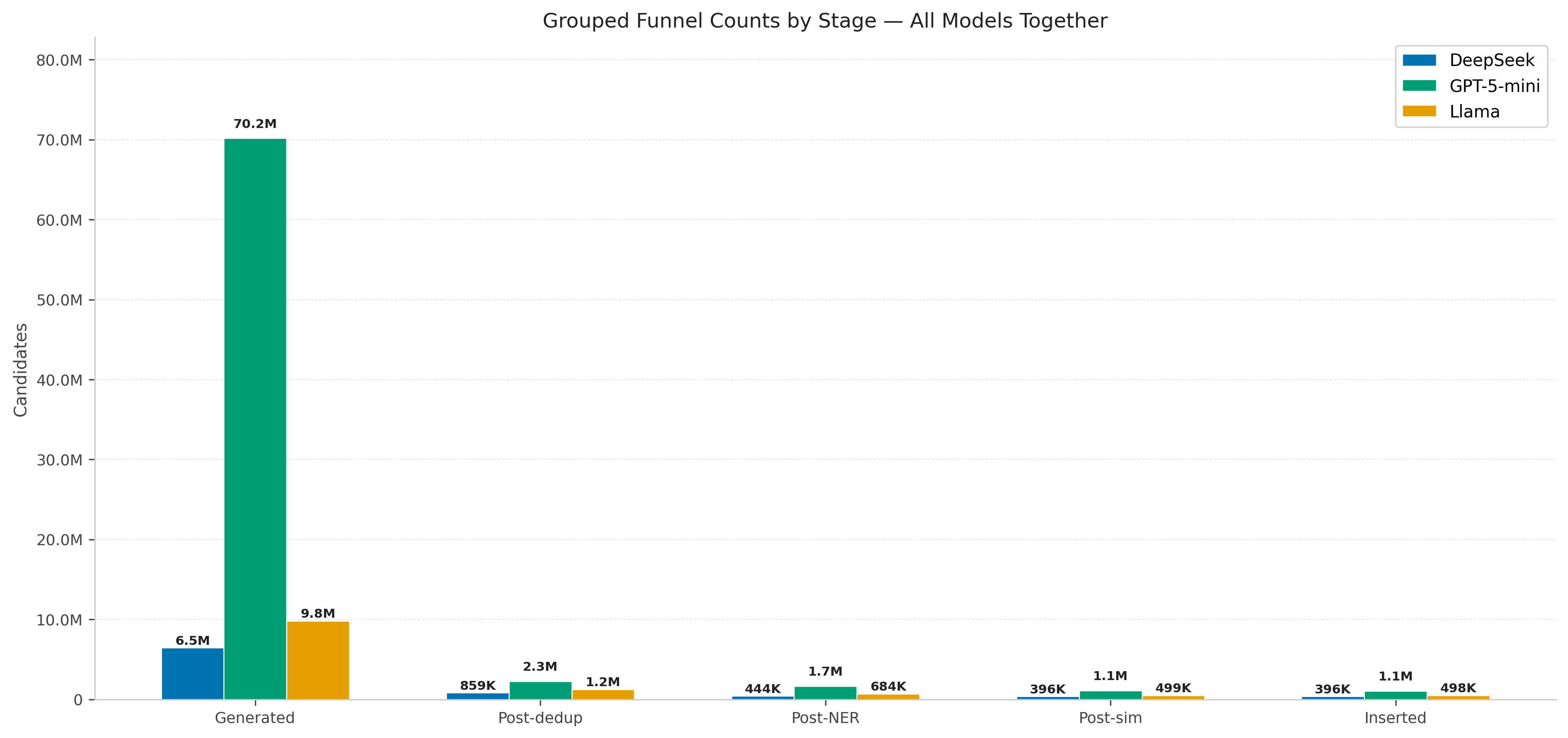}
  \caption{Funnel counts per stage, grouped by model. Each cluster on
  the x-axis is a pipeline stage; bars within a cluster are the three
  models. The decay from \textit{Generated} to \textit{Inserted} is
  visible as a near-monotone right-to-left shrinking pattern for every
  model. GPT-5-mini operates at roughly an order of magnitude larger
  scale at every stage, but the \emph{shape} of the funnel is
  remarkably similar across models.}
  \label{fig:funnel_grouped_counts}
\end{figure}

\begin{figure}[h]
  \centering
  \includegraphics[width=0.95\linewidth]{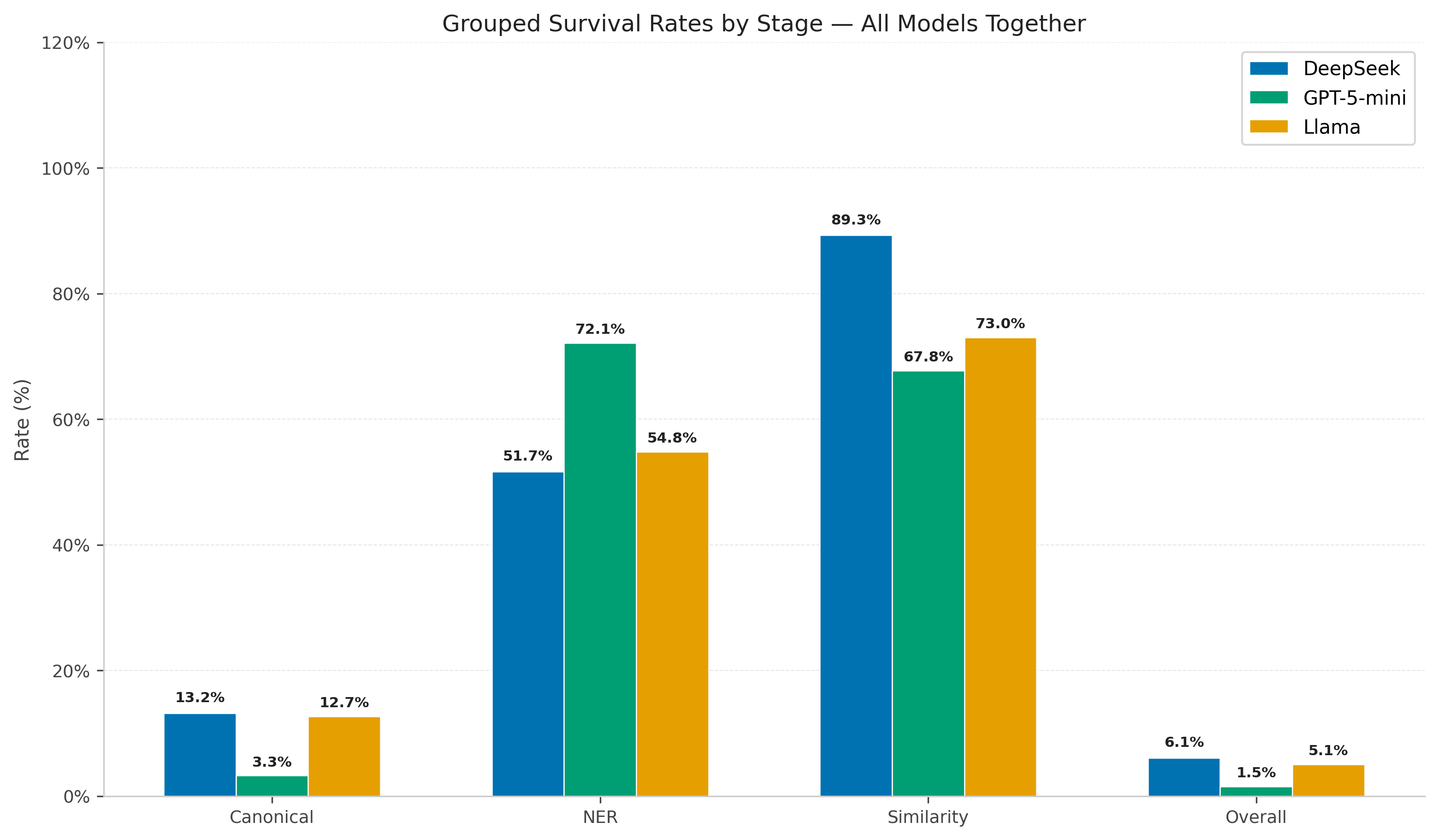}
  \caption{Per-stage survival rates, grouped by model. Three patterns
  stand out: (i)~GPT-5-mini has the lowest canonical survival (18.4\%)
  because its much larger committed index makes more candidates
  redundant; (ii)~Llama's NER survival is the lowest at 54.8\% and is
  dragged down by parse failures rather than legitimate ``not a named
  entity'' rejections (Appendix~\ref{app:funnel_llama}); (iii)~DeepSeek's
  similarity survival is the highest at 87.5\%, indicating its
  candidate set carries less semantic redundancy at the post-NER stage.}
  \label{fig:funnel_grouped_survival}
\end{figure}

\begin{figure}[h]
  \centering
  \includegraphics[width=0.95\linewidth]{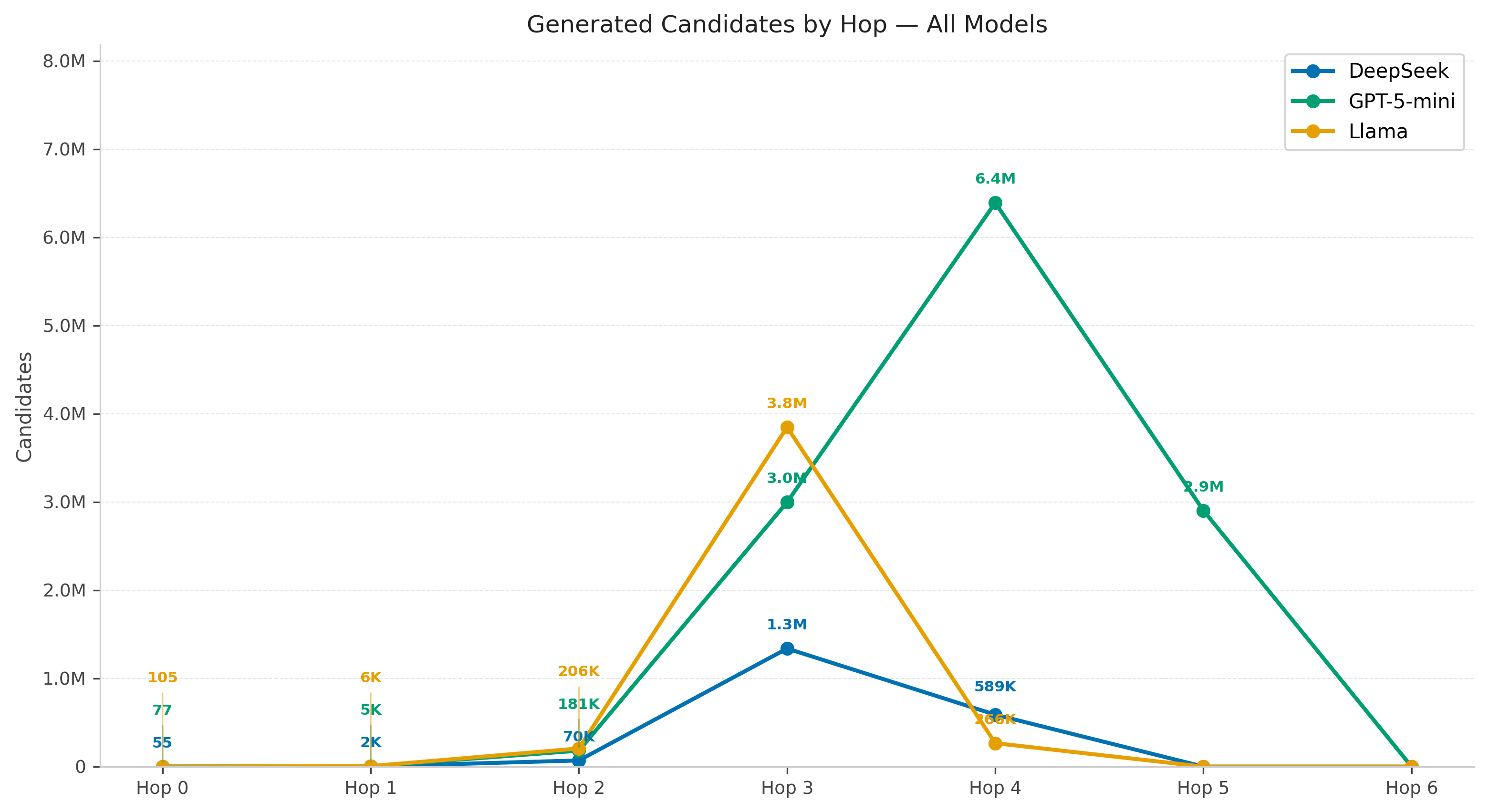}
  \caption{Pre-NER candidate count by BFS hop, one line per model. The
  characteristic peak-and-decay shape reflects two regimes: rapid
  growth as BFS reaches the wide middle hops (peaking at hop~3 for the
  open-weight models and hop~4 for GPT-5-mini), followed by collapse
  once most surfaced entities are already in the committed index and
  most parents are pruned at canonical dedup.}
  \label{fig:funnel_generated_hop}
\end{figure}

\begin{figure}[h]
  \centering
  \includegraphics[width=0.95\linewidth]{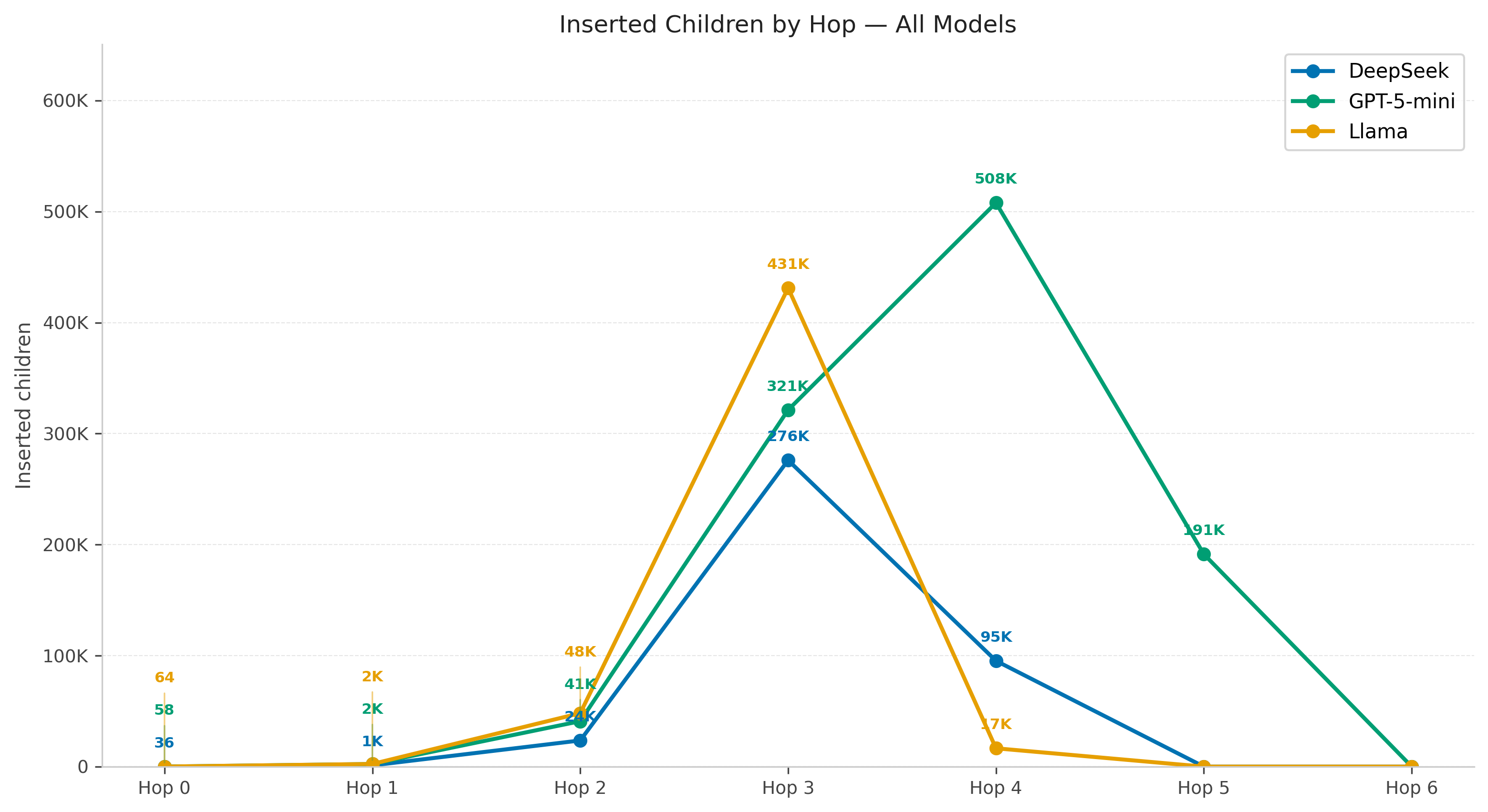}
  \caption{Inserted children per hop, one line per model. This is the
  shape of the resulting corpus: the bulk of GPT-5-mini's ${\sim}$1M
  articles are at hops 4--5, whereas the open-weight runs terminate at
  hops 4--5 because 120K max-subject budget was reached. The shared
  asymmetry - light hops 0--2, dense hops 3--5, depleted hops
  beyond - is characteristic of BFS over a power-law-degree
  encyclopedic graph.}
  \label{fig:funnel_inserted_hop}
\end{figure}

\begin{figure}[h]
  \centering
  \includegraphics[width=0.95\linewidth]{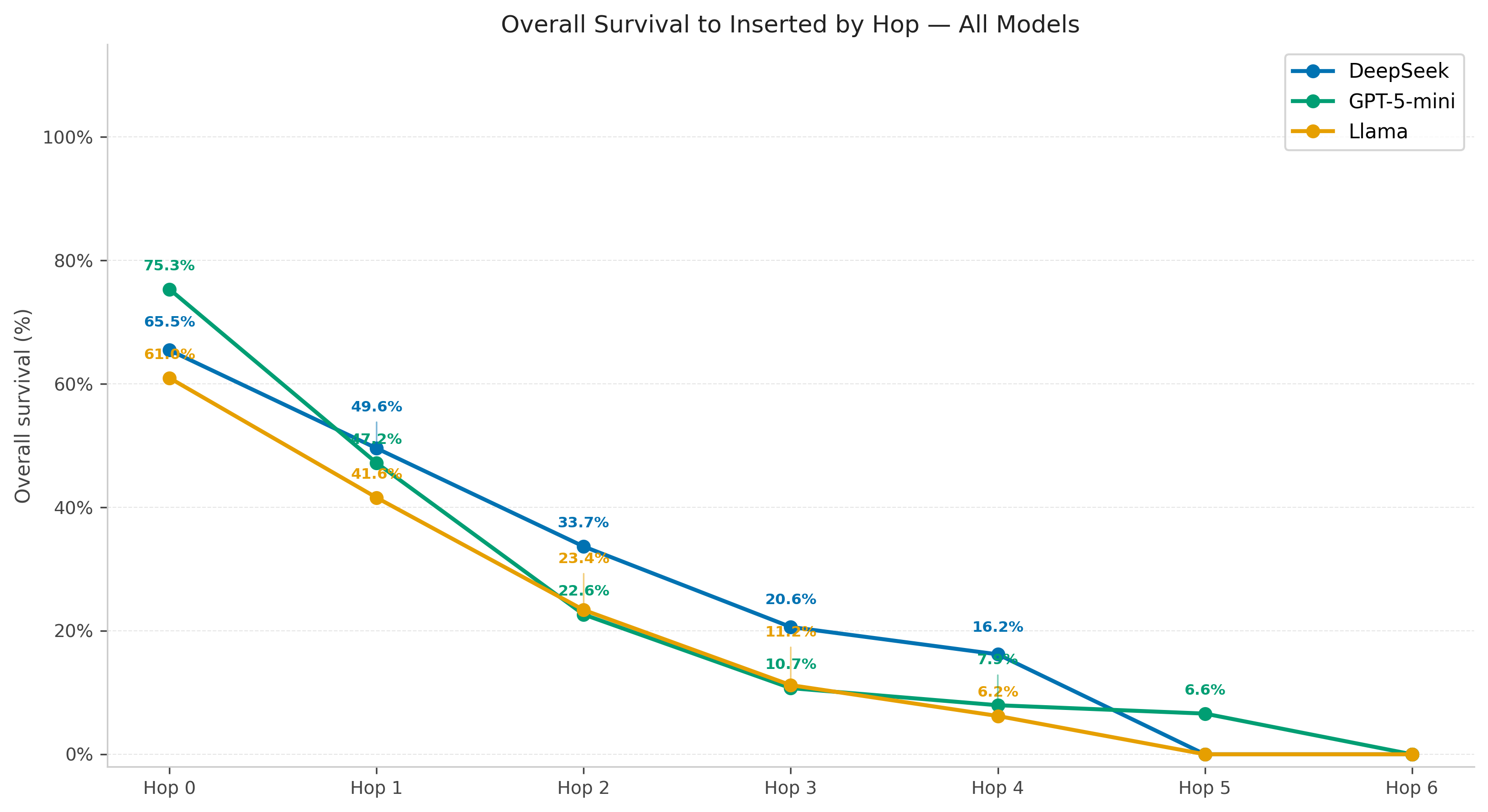}
  \caption{Overall survival from raw candidate to inserted queue
  subject, by BFS hop and model. Survival falls sharply with depth:
  for GPT-5-mini the rate drops from 75.3\% at hop~0 to 6.6\% at
  hop~5, and the open-weight models trace similar trajectories. The
  cause is not the LLM stages (NER and similarity survival are
  relatively flat with depth) but canonical dedup, whose survival
  collapses as the committed index saturates the local neighborhood
  of the BFS frontier.}
  \label{fig:funnel_overall_surv_hop}
\end{figure}

Figures~\ref{fig:funnel_grouped_counts}--\ref{fig:funnel_overall_surv_hop}
present the five cross-model figures generated by the combined funnel
analysis.
 
\paragraph{Reading the curves together.}
Figures~\ref{fig:funnel_generated_hop}--\ref{fig:funnel_overall_surv_hop}
tell a consistent story across all three models: the funnel becomes
strictly more selective at each successive hop, not because the
filters become harsher (their per-call behavior is nearly stationary)
but because the population entering the filters becomes
proportionally more redundant. This is the structural reason the
hop-stratified factuality results in
Table~\ref{tab:hop_factuality} attach to a thinning external-evidence
tail rather than to deteriorating model knowledge.
 
\subsection{GPT-5-mini Deep Dive}
\label{app:funnel_gpt}
 
Total: 1{,}063{,}949 entities; 1{,}008{,}947 with articles. Only 27
entities (0.003\%) had articles generated but were never processed
through the full pipeline - the cleanest of the three runs.
 
\begin{figure}[h]
  \centering
  \includegraphics[width=0.95\linewidth]{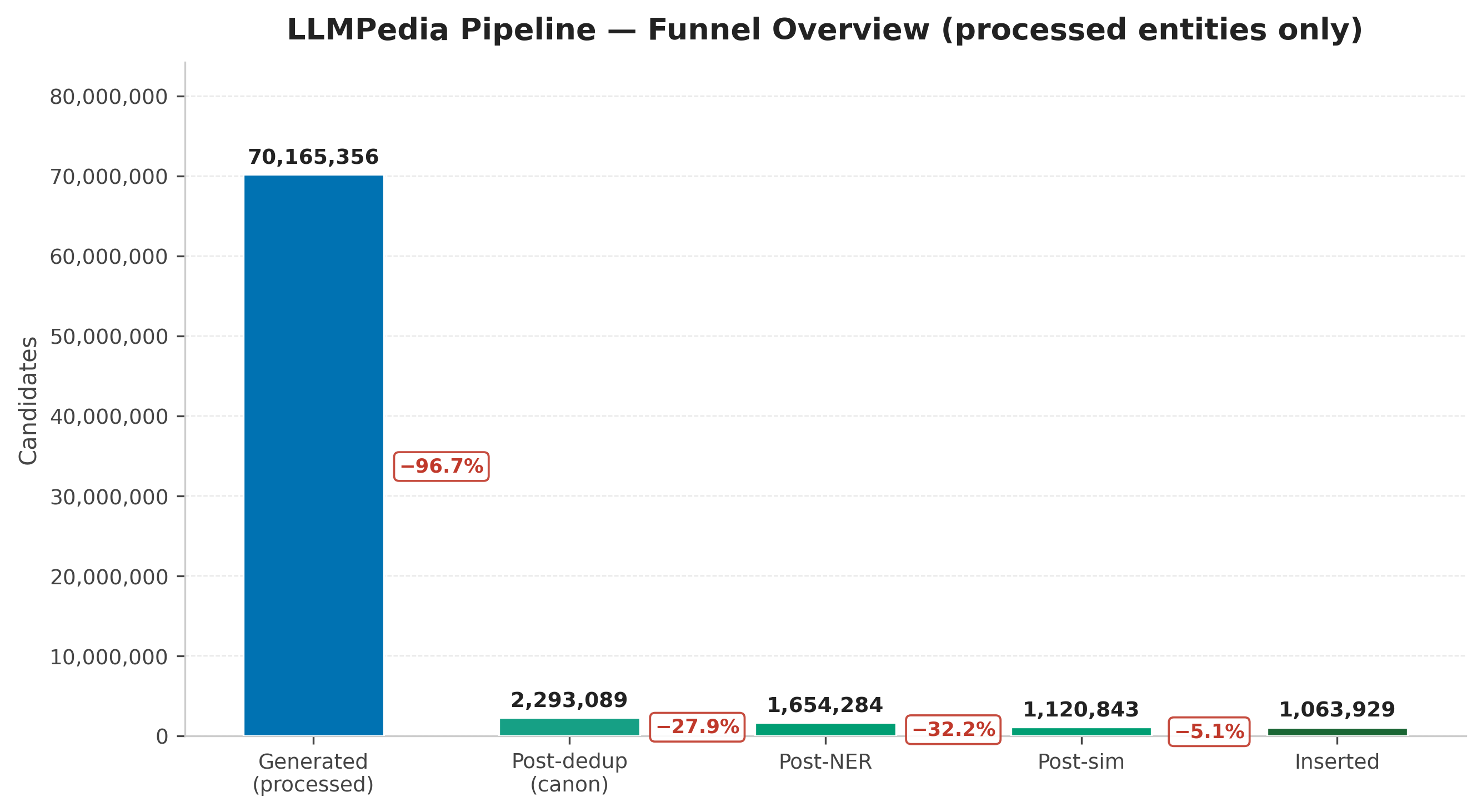}
  \caption{GPT-5-mini five-stage funnel from 70.2M raw candidates to
  1.06M queued subjects. The largest absolute loss is at pre-NER
  canonical deduplication ($-$82.3\%), reflecting that most wikilinks
  in any given article are surface variants of entities already
  committed elsewhere in the corpus.}
  \label{fig:gpt_funnel_stages}
\end{figure}
 
\begin{figure}[h]
  \centering
  \includegraphics[width=0.95\linewidth]{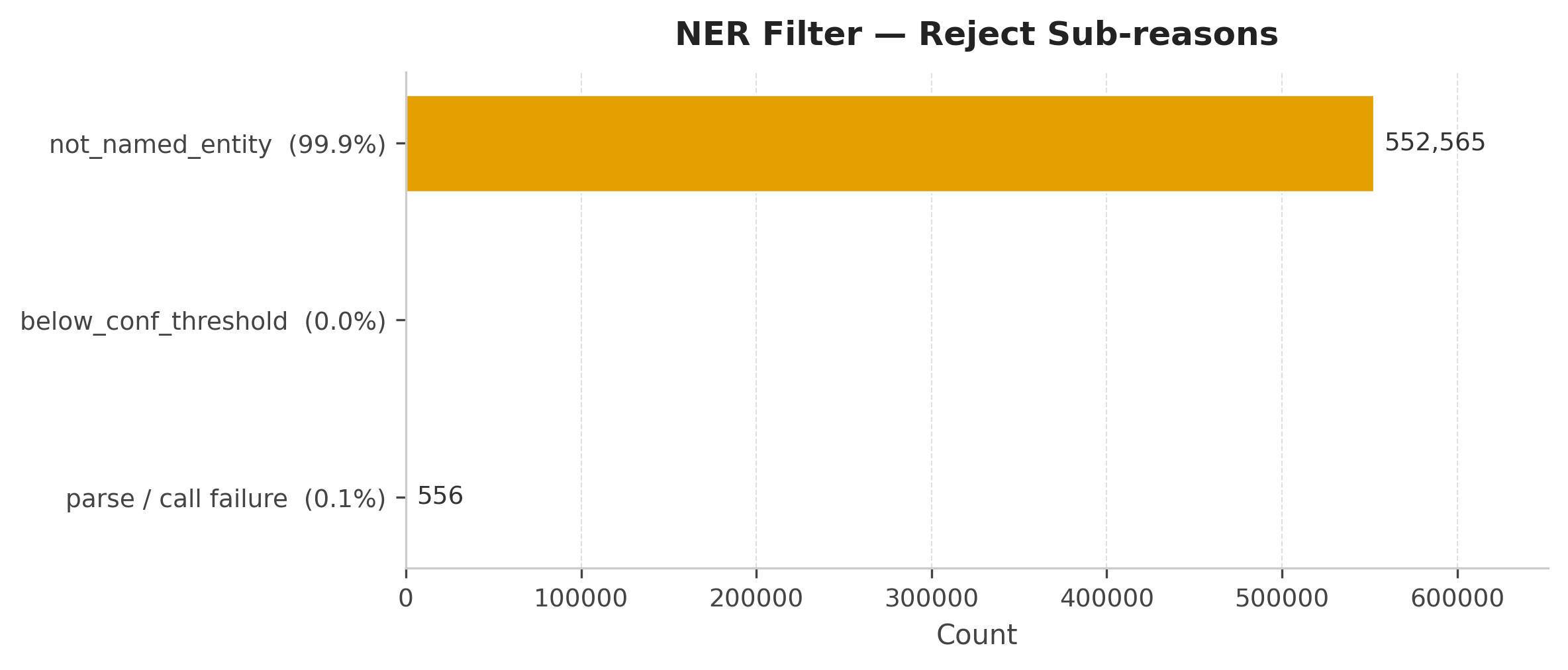}
  \caption{GPT-5-mini NER reject reason breakdown. Of 553K NER
  rejections, 99.9\% (552{,}565) are the legitimate
  \texttt{not\_named\_entity} class - the filter doing its intended
  job. Parse failures account for just 0.1\% (556 rejections), and
  every NER call parsed as native JSON (no fallback regex paths were
  triggered).}
  \label{fig:gpt_ner_reasons}
\end{figure}
 
\begin{figure}[h]
  \centering
  \includegraphics[width=0.95\linewidth]{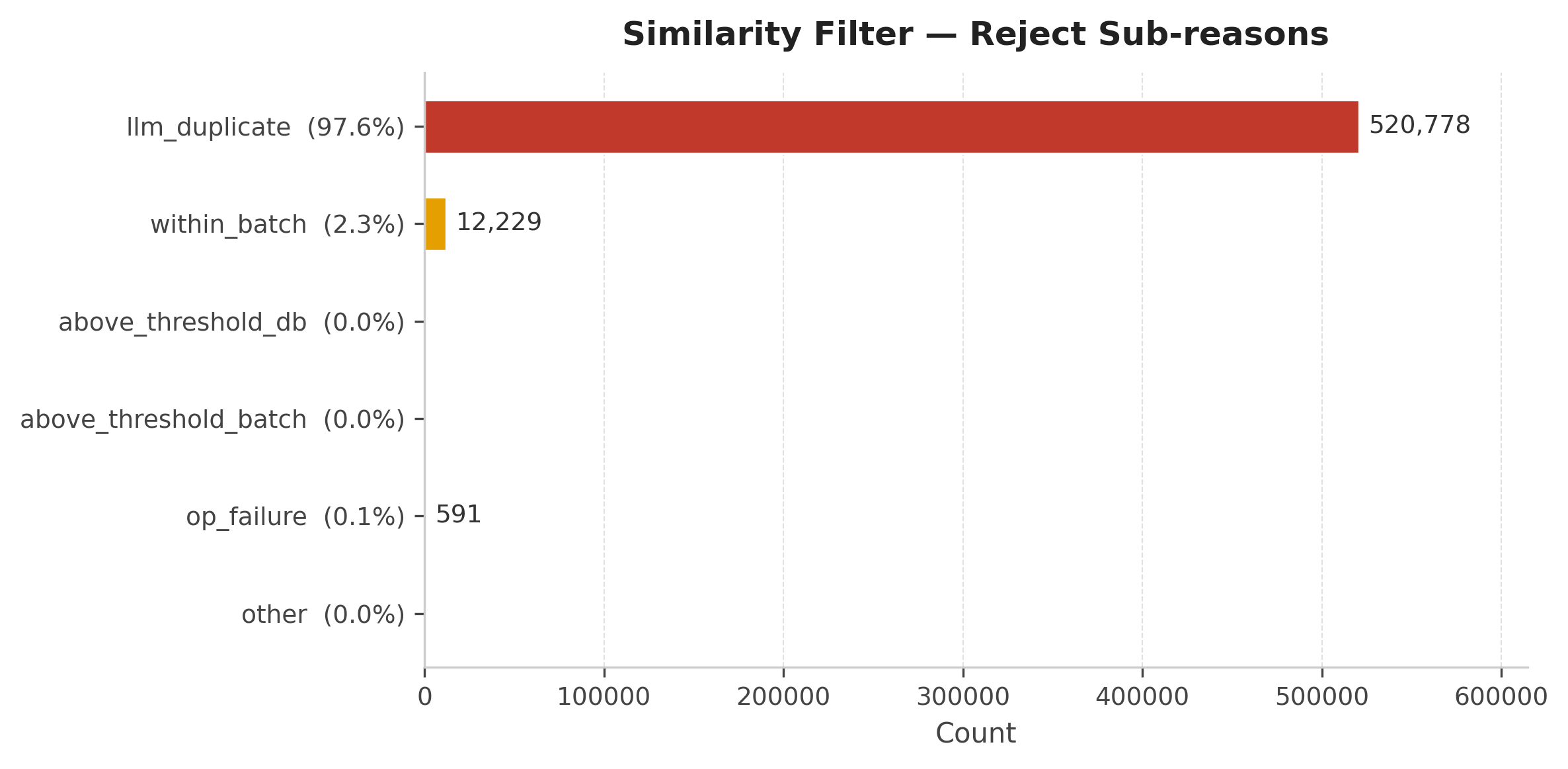}
  \caption{GPT-5-mini similarity reject reasons. Of 534K rejections,
  97.6\% are LLM-confirmed duplicates - the arbitration LLM agreed
  with the embedding threshold that the candidate was a surface
  variant of an already-committed entity. Within-batch duplicates
  (2.3\%) are concurrent NER workers independently nominating the
  same entity in the same BFS wave (Appendix~\ref{app:dedup-guarantee}).
  Operational failures (591 rejections, 0.1\%) come from transient
  batch-output gaps and SDK exceptions; the strict-gate policy treats
  these as duplicates to avoid contaminating the corpus, at the cost
  of a small downward recall bias.}
  \label{fig:gpt_sim_reasons}
\end{figure}
 
\begin{figure}[h]
  \centering
  \includegraphics[width=0.95\linewidth]{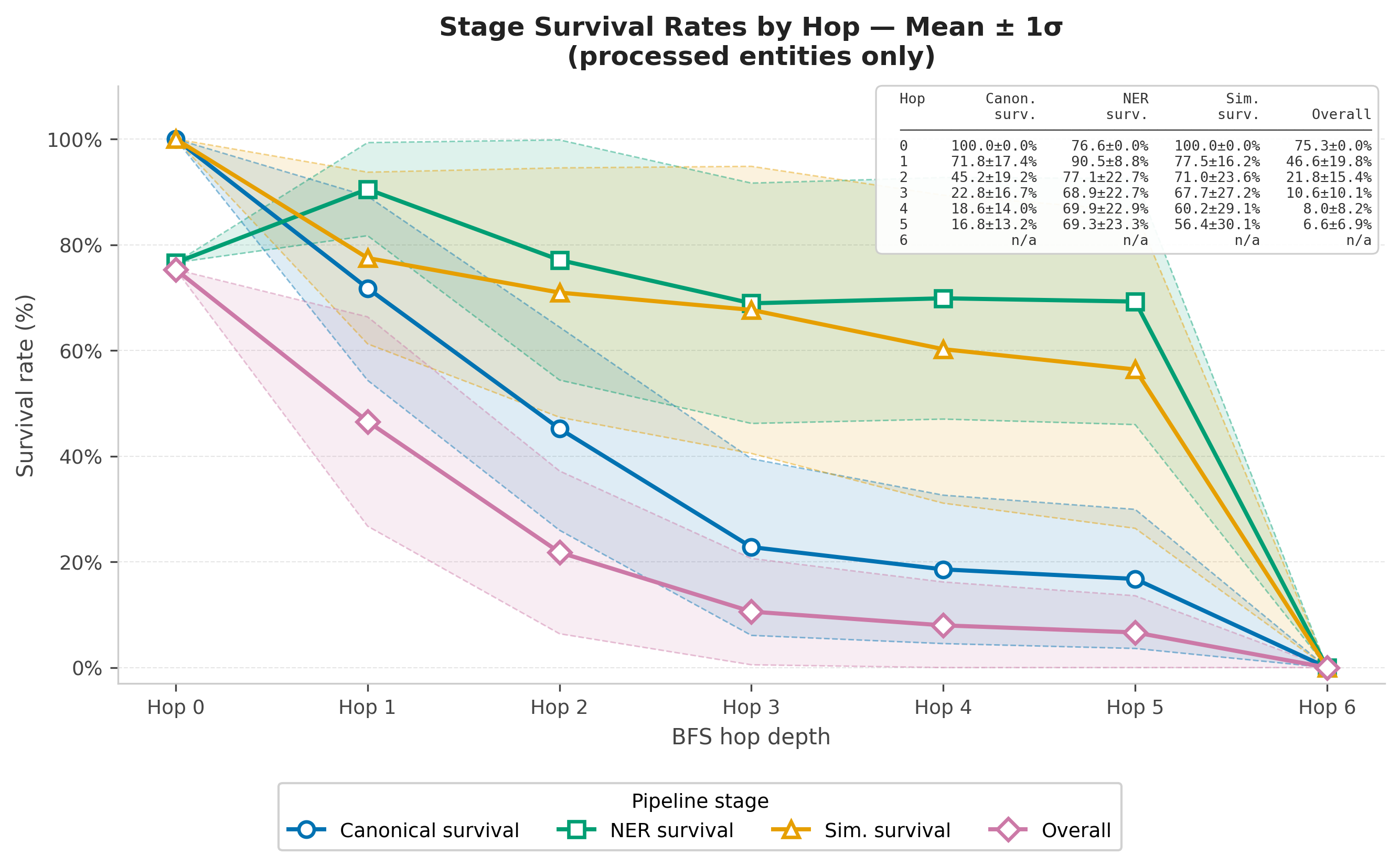}
  \caption{Per-hop survival rates for GPT-5-mini with ${\pm}1\sigma$
  bands across entities. Three things to read off: (i)~survival is
  monotonically decreasing in hop; (ii)~variance across entities
  within a hop is large (especially at deep hops, where some entities
  produce dozens of survivors and many produce zero - the power-law
  tail of wikilink fan-out); (iii)~the gap between NER survival and
  similarity survival narrows with depth because, at the frontier,
  the embedding index has already absorbed most semantically nearby
  entities, so similarity rejections shift toward LLM-confirmed
  duplicates of newly committed siblings.}
  \label{fig:gpt_survival_band}
\end{figure}
 
\begin{figure}[h]
  \centering
  \includegraphics[width=0.95\linewidth]{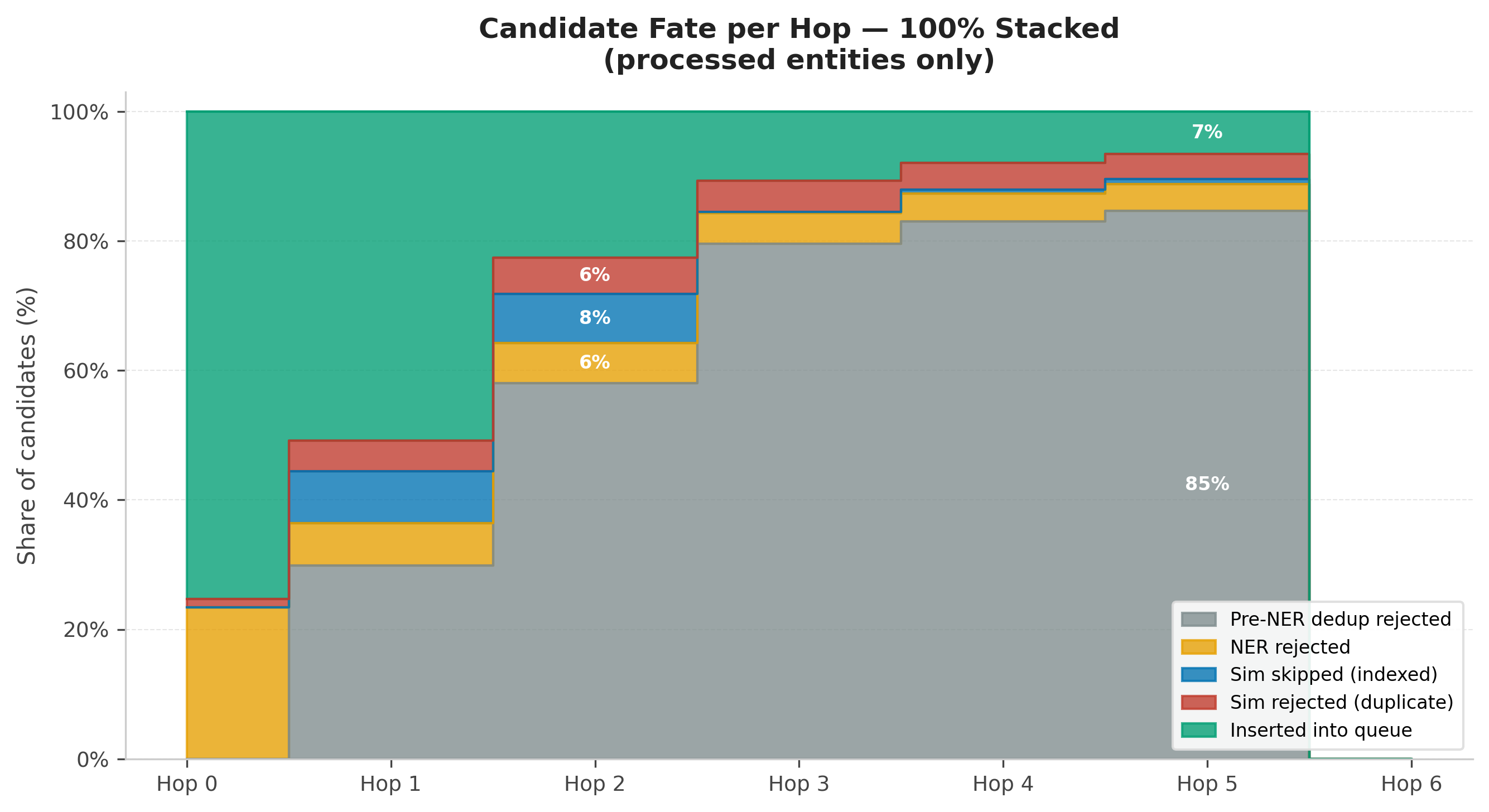}
  \caption{GPT-5-mini candidate fate per hop, 100\% normalized stacked
  area. At hop~1 most candidates make it through the entire pipeline
  to insertion; by hop~5 the pre-NER dedup share dominates and the
  inserted share has shrunk to a few percent. The displacement is
  driven entirely by the redundancy layer, not by NER or similarity
  rejecting more aggressively at depth.}
  \label{fig:gpt_fate_pct}
\end{figure}
 
\begin{figure}[h]
  \centering
  \includegraphics[width=0.95\linewidth]{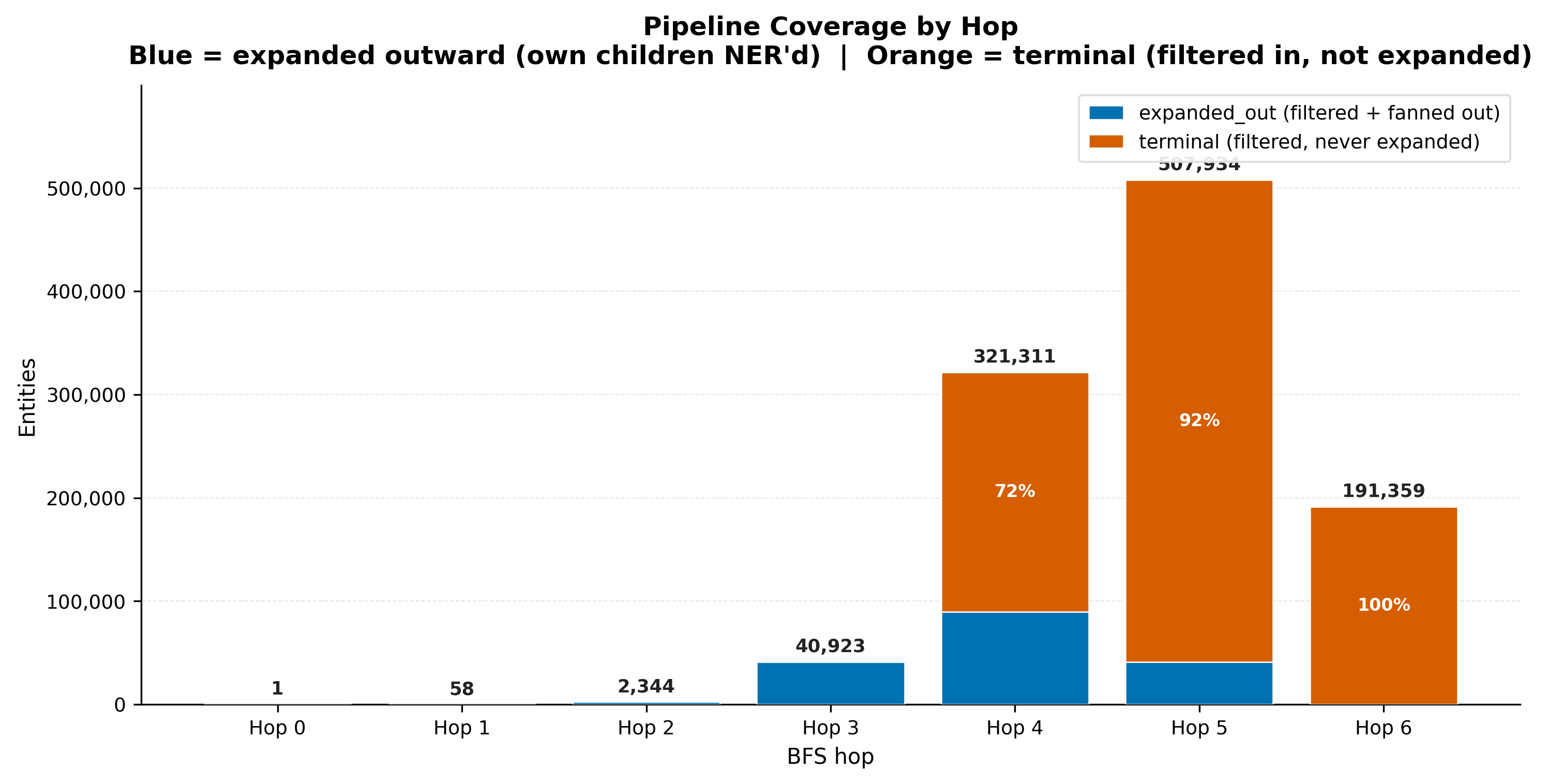}
  \caption{GPT-5-mini pipeline coverage by hop. Blue bars are
  \emph{expanded-out} entities (NER ran on their outbound wikilinks);
  orange bars are \emph{terminal} entities (article exists and is
  validly filtered into the corpus, but never expanded). Terminality
  is concentrated at hops 5--6 where the run reached its
  computational ceiling: 192K of the 1.06M subjects are terminal at
  hop~6, which is the natural BFS boundary. The track-3 factuality
  evaluation samples from filtered-in entities at all hops regardless
  of expanded status, which is why the headline coverage and true-rate
  numbers are reported over the full ${\sim}$1M corpus.}
  \label{fig:gpt_coverage_hop}
\end{figure}
 
\begin{figure}[h]
  \centering
  \includegraphics[width=0.95\linewidth]{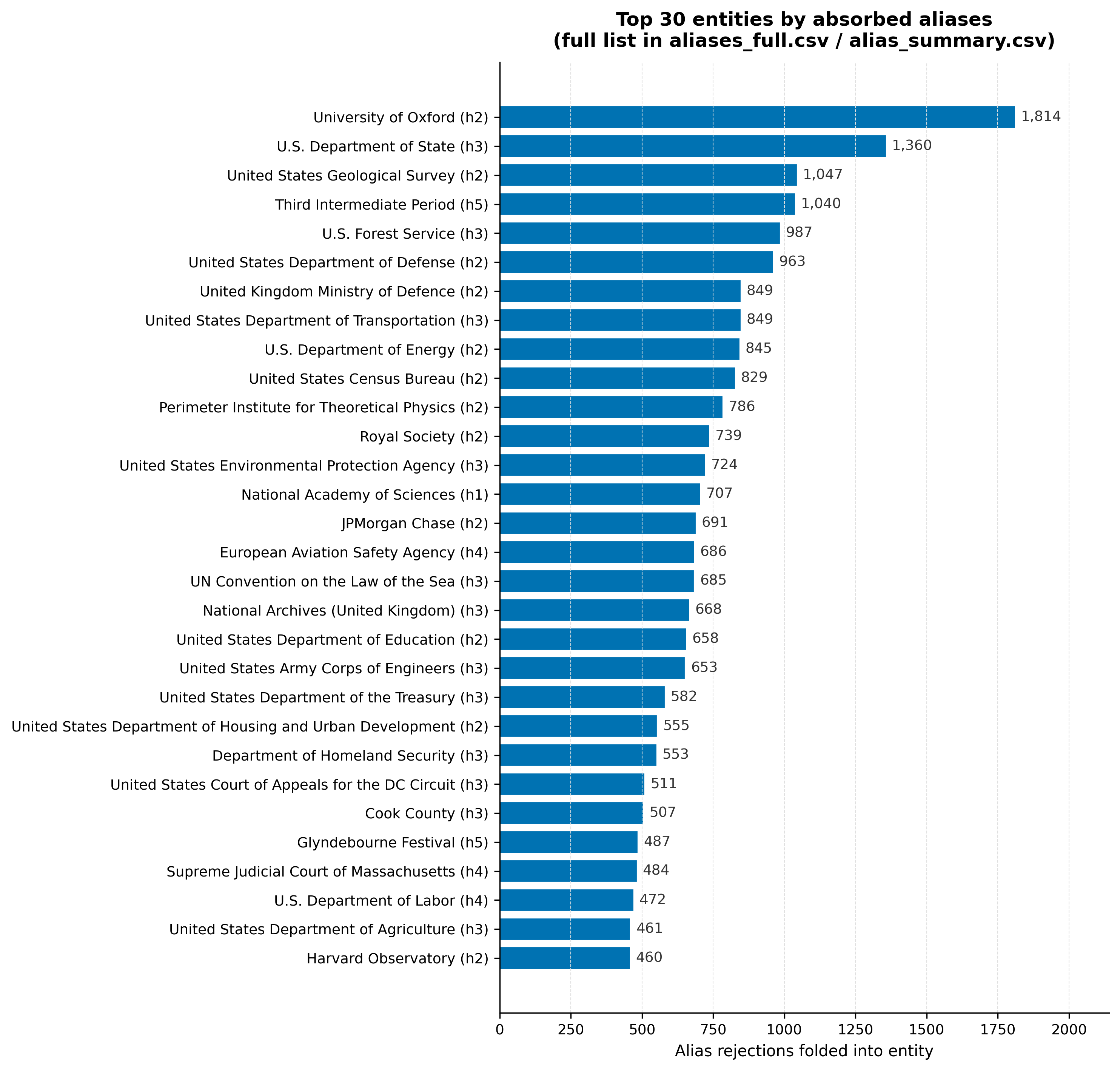}
  \caption{Top-30 entities by aliases absorbed in the GPT-5-mini run.
  These are the most ``magnetic'' canonical entities: each one
  collapsed dozens to thousands of surface-form variants
  (\textit{University of Oxford} alone absorbed 1{,}814 distinct
  aliases). The top of the list is dominated by government bodies,
  universities, and standards organizations, where the variant space
  (acronyms, ``U.S.''/``United States'' prefixes, capitalization
  drift) is genuinely large. Without Stage~3 disambiguation each of
  these would have splintered into dozens of near-duplicate articles.}
  \label{fig:gpt_aliases}
\end{figure}
 
\subsection{DeepSeek-V3 Deep Dive}
\label{app:funnel_deepseek}
 
Total: 396{,}091 entities; 120{,}139 with articles. The run produced
articles at hops 0--5 with 37 entities (0.03\%) showing partial
processing.
 
\begin{figure}[h]
  \centering
  \includegraphics[width=0.95\linewidth]{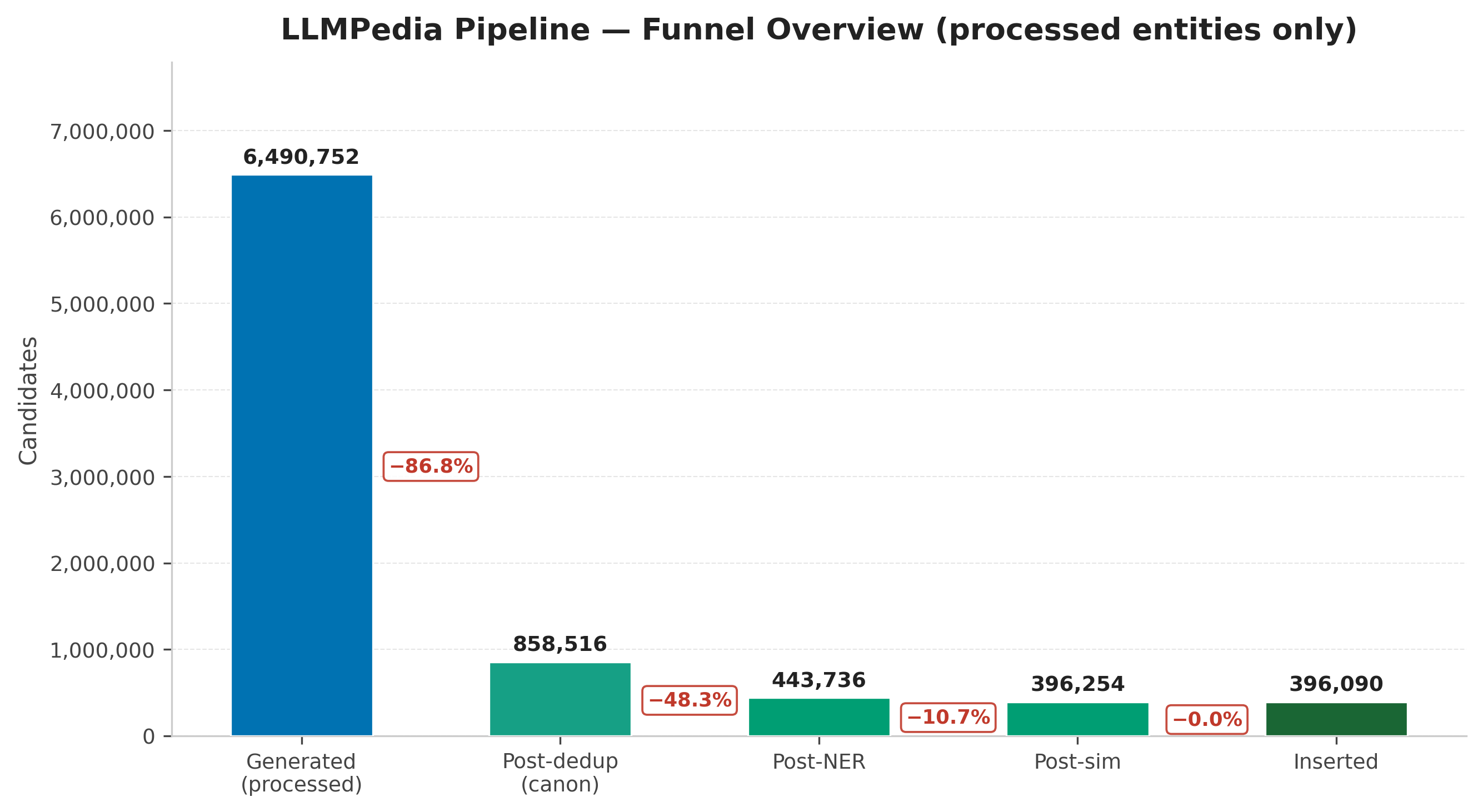}
  \caption{DeepSeek five-stage funnel from 6.5M raw candidates to
  396K queued subjects. The pre-NER canonical dedup stage is less
  severe than for GPT-5-mini ($-$57.1\% vs.\ $-$82.3\%) because the
  committed index is smaller; the LLM stages take a proportionally
  larger share of the total loss.}
  \label{fig:ds_funnel_stages}
\end{figure}
 
\begin{figure}[h]
  \centering
  \includegraphics[width=0.95\linewidth]{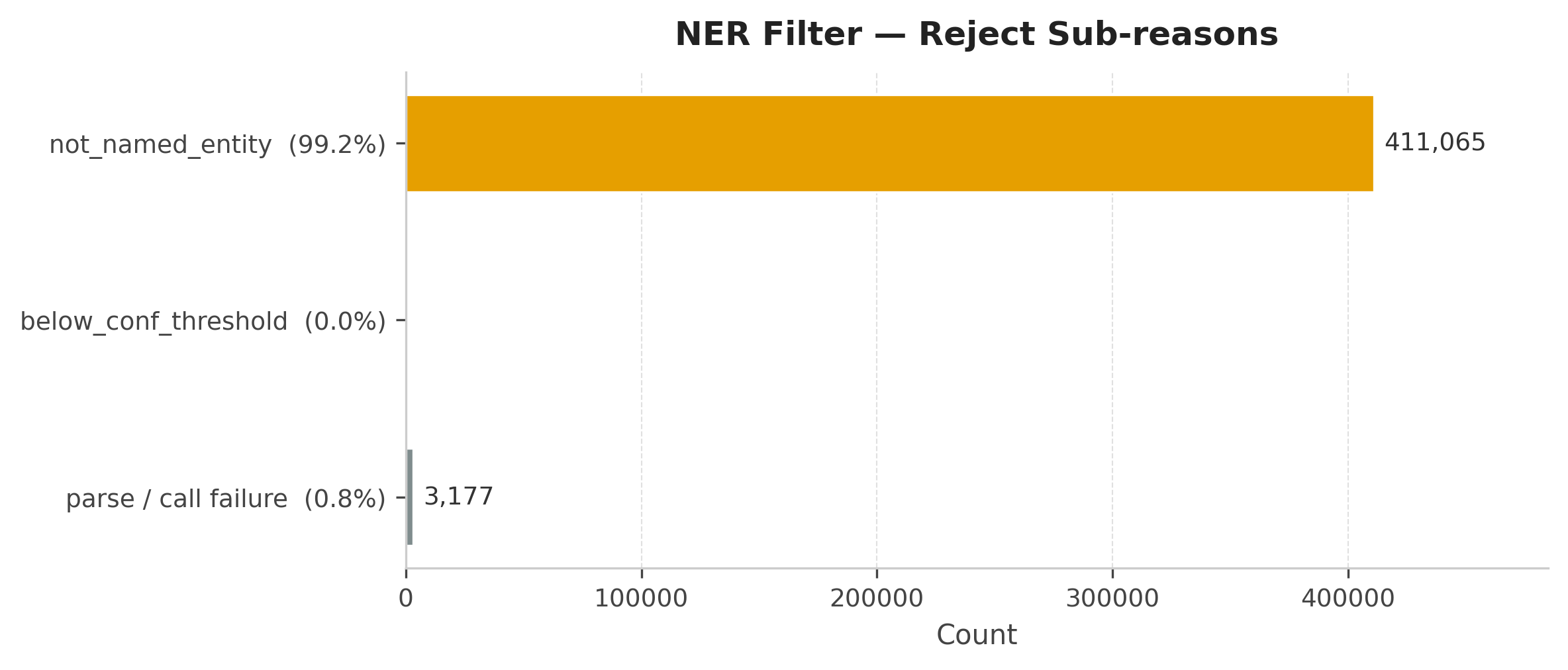}
  \caption{DeepSeek NER reject reasons. 99.2\% of 414K rejections are
  \texttt{not\_named\_entity} - the filter behaving as designed.
  Parse failures account for 0.8\% (3{,}177 rejections), with 99.1\%
  of NER calls parsing as native JSON and the remainder recovered via
  the JSON-substring fallback.}
  \label{fig:ds_ner_reasons}
\end{figure}
 
\begin{figure}[h]
  \centering
  \includegraphics[width=0.95\linewidth]{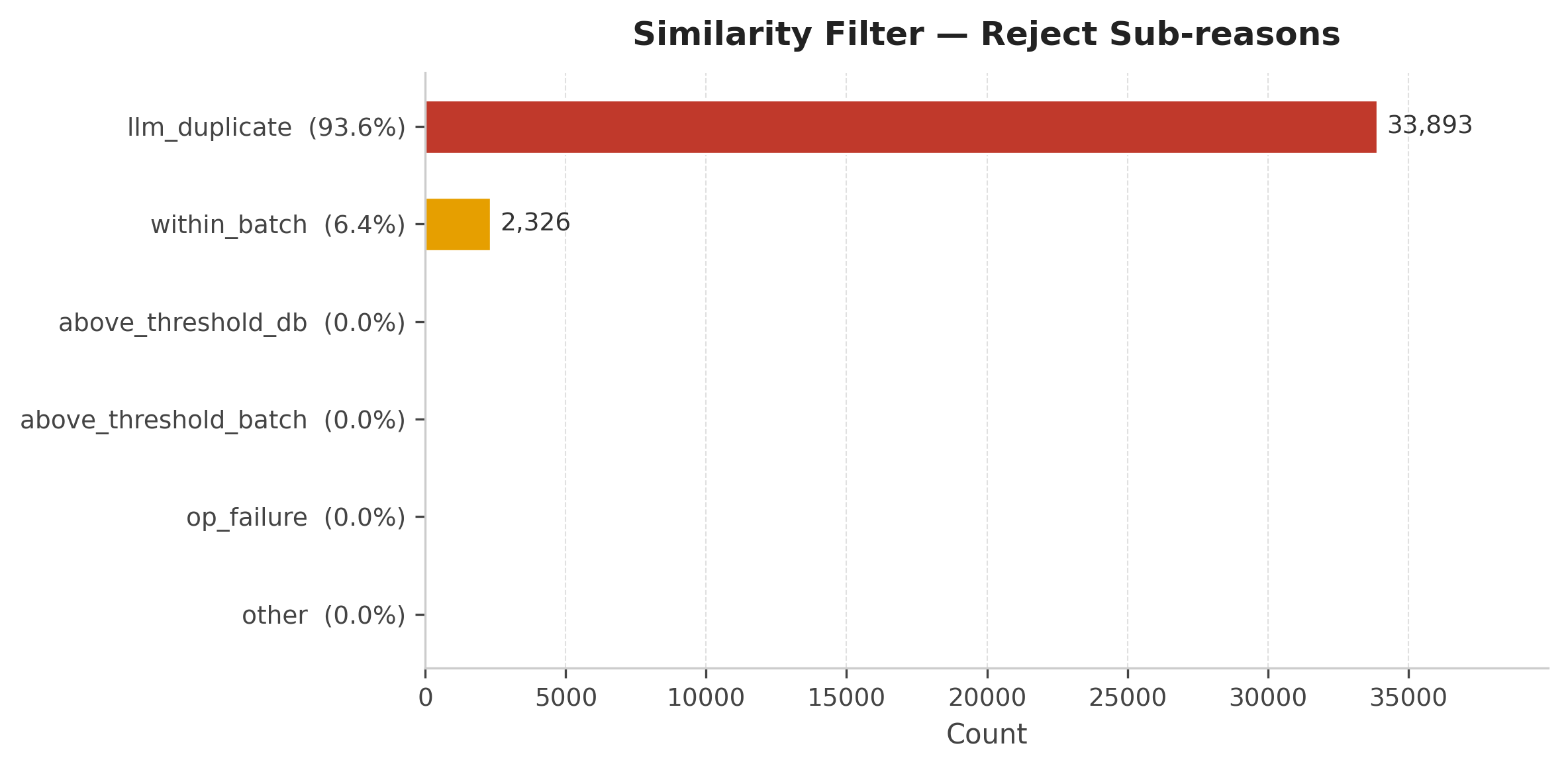}
  \caption{DeepSeek similarity reject reasons. 93.6\% LLM-confirmed
  duplicates and 6.4\% within-batch duplicates account for all
  rejections; no operational failures occurred during this run.}
  \label{fig:ds_sim_reasons}
\end{figure}
 
\begin{figure}[h]
  \centering
  \includegraphics[width=0.95\linewidth]{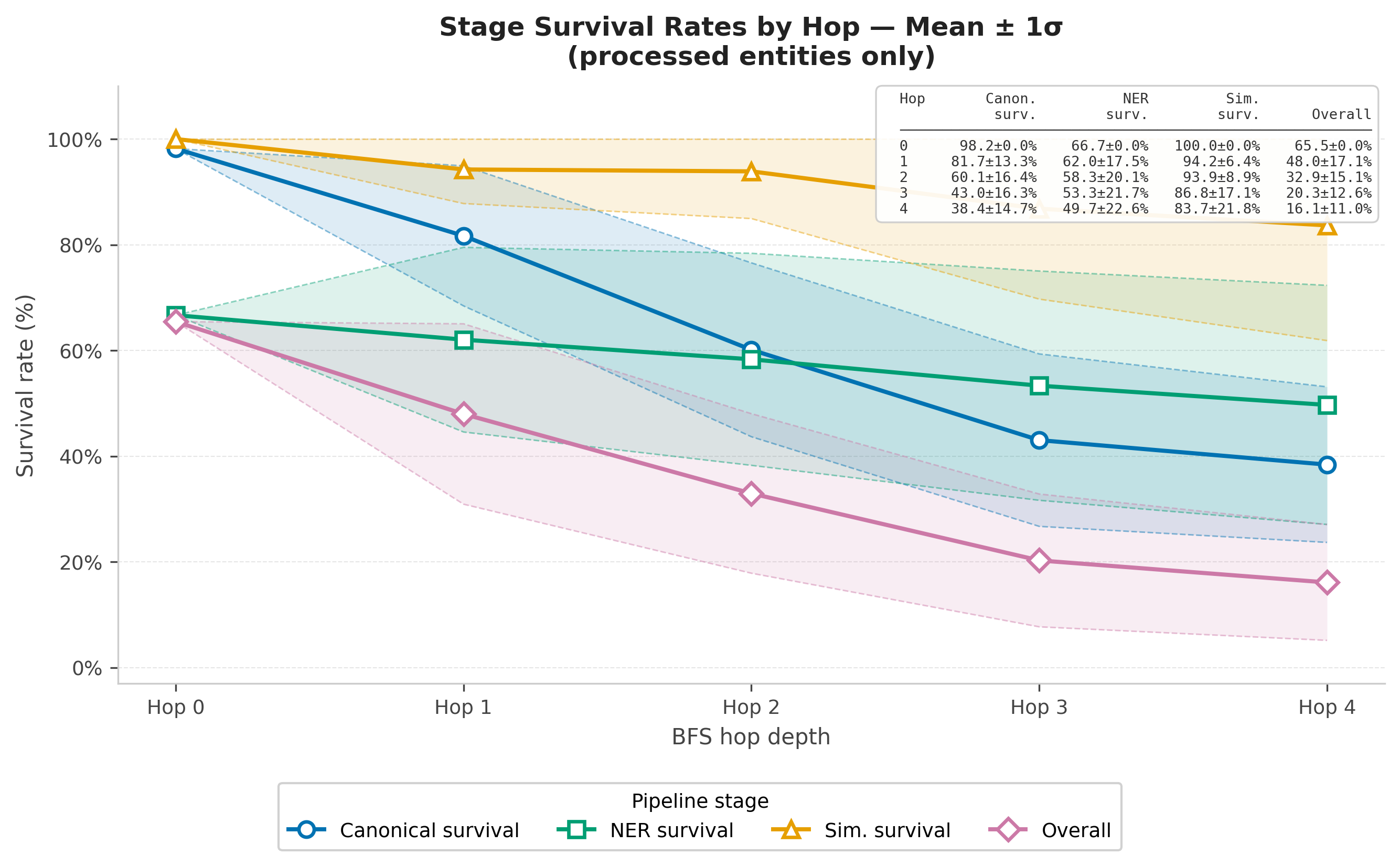}
  \caption{DeepSeek per-hop survival rates with ${\pm}1\sigma$ bands.
  The overall survival trajectory mirrors GPT-5-mini's (65.5\% at
  hop~0 down to 16.2\% at hop~4) but with consistently higher
  similarity survival, reflecting the smaller embedding index.}
  \label{fig:ds_survival_band}
\end{figure}
 
\begin{figure}[h]
  \centering
  \includegraphics[width=0.95\linewidth]{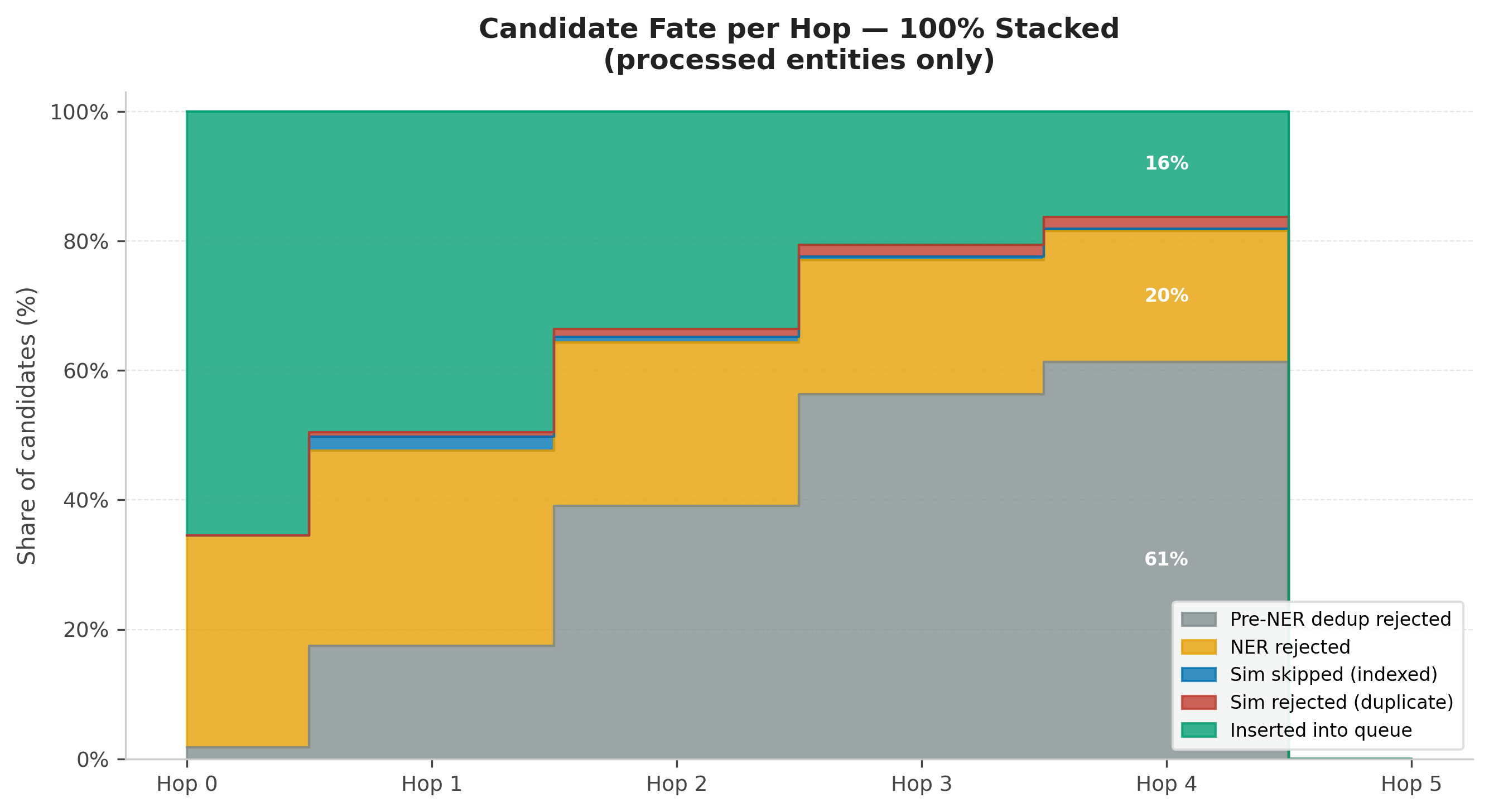}
  \caption{DeepSeek candidate fate per hop, 100\% normalized. The
  insertion fraction is meaningfully higher at every hop than for
  GPT-5-mini, again because of the smaller committed index. The shape
  of the displacement - pre-NER dedup growing with depth - is
  identical to the other two models.}
  \label{fig:ds_fate_pct}
\end{figure}
 
\begin{figure}[h]
  \centering
  \includegraphics[width=0.95\linewidth]{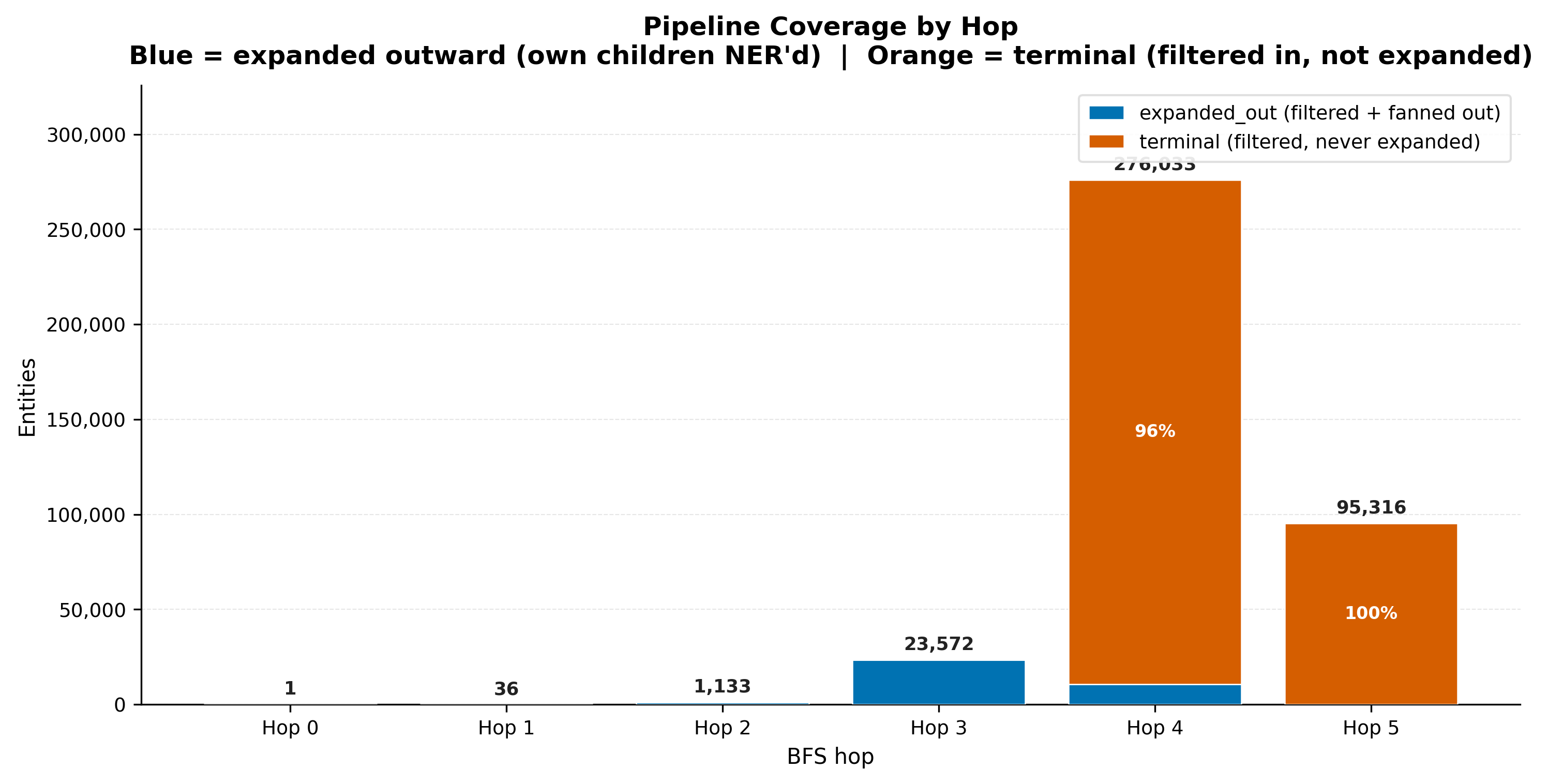}
  \caption{DeepSeek pipeline coverage by hop. The run terminated with hop~4 only partially expanded:
10{,}719 of 276{,}033 hop~4 entities (3.9\%) are expanded-out,
and hop~5 is entirely terminal.
  and hop~5 entirely terminal. The 96.1\% terminal share at hop~4
  reflects the max-subject budget being reached: those 265K entities
  legitimately entered the corpus as filtered-in articles, but the
  pipeline was stopped before their outbound wikilinks could be
  processed.}
  \label{fig:ds_coverage_hop}
\end{figure}
 
\begin{figure}[h]
  \centering
  \includegraphics[width=0.95\linewidth]{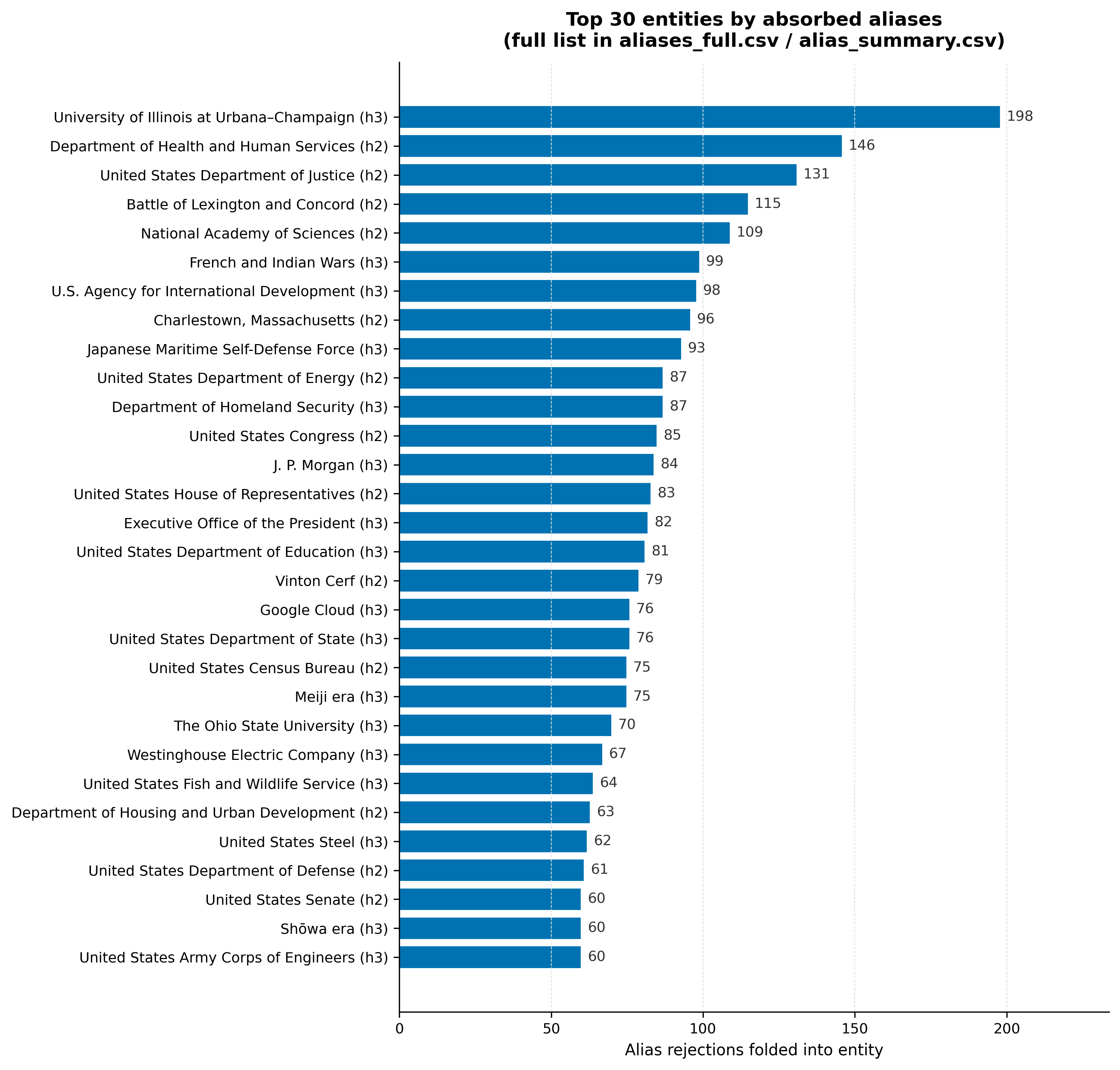}
  \caption{Top-30 DeepSeek entities by aliases absorbed. Smaller scale
  than GPT-5-mini (top entity absorbs 198 aliases vs.\ GPT's 1{,}814)
  but the same categorical pattern: governmental bodies, universities,
  and prominent historical events dominate.}
  \label{fig:ds_aliases}
\end{figure}
 
\subsection{Llama-3.3-70B Deep Dive}
\label{app:funnel_llama}
 
Total: 498{,}226 entities; 120{,}100 with articles. 3 entities (0.003\%)
showed partial processing. The Llama run also has the strictest
NER-parse-failure footprint, discussed below.
 
\begin{figure}[h]
  \centering
  \includegraphics[width=0.95\linewidth]{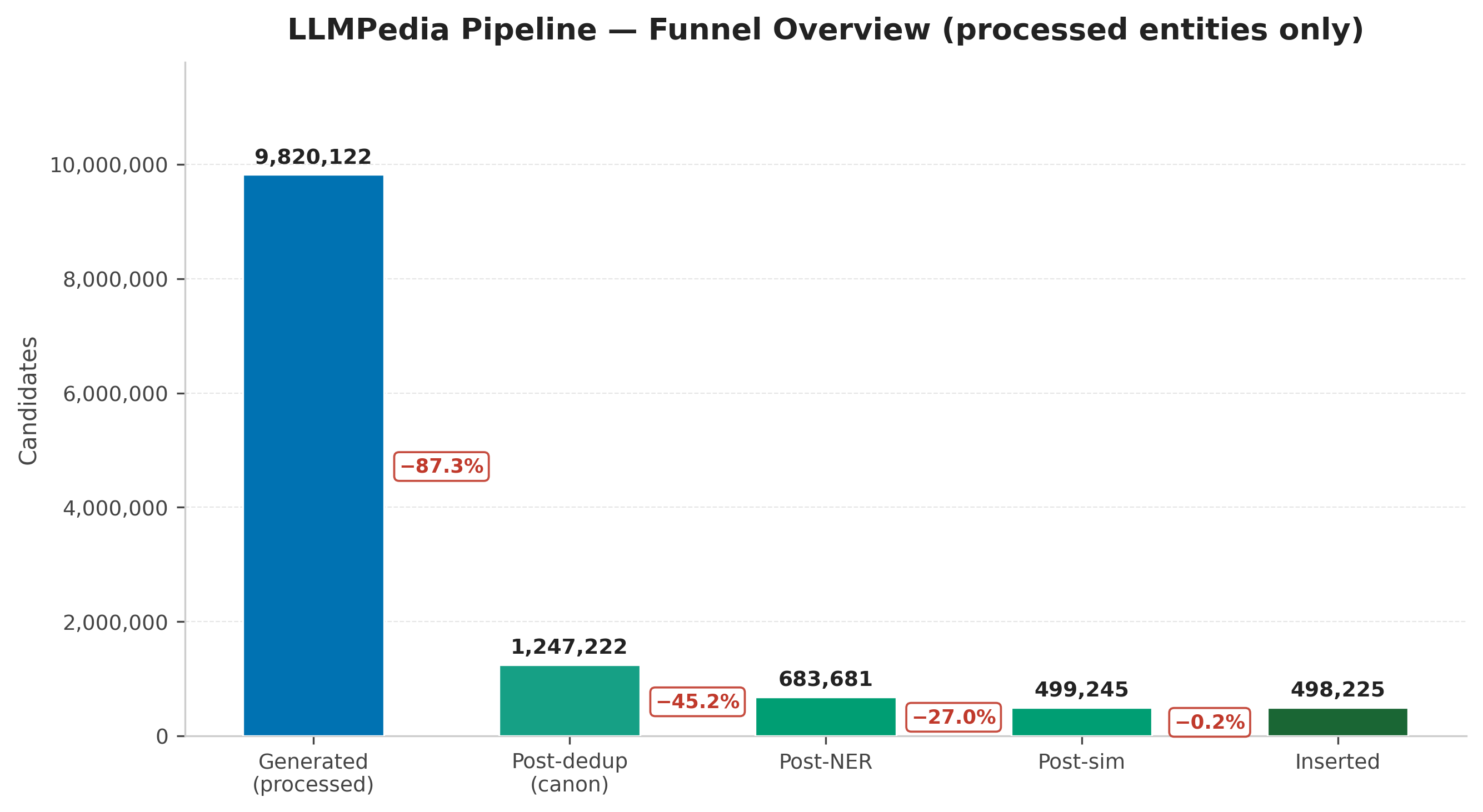}
  \caption{Llama five-stage funnel from 9.8M raw candidates to 498K
  queued subjects. The pre-NER dedup loss (71.2\%) sits between
  GPT-5-mini's and DeepSeek's, as expected for an intermediate-sized
  committed index.}
  \label{fig:lm_funnel_stages}
\end{figure}
 
\begin{figure}[h]
  \centering
  \includegraphics[width=0.95\linewidth]{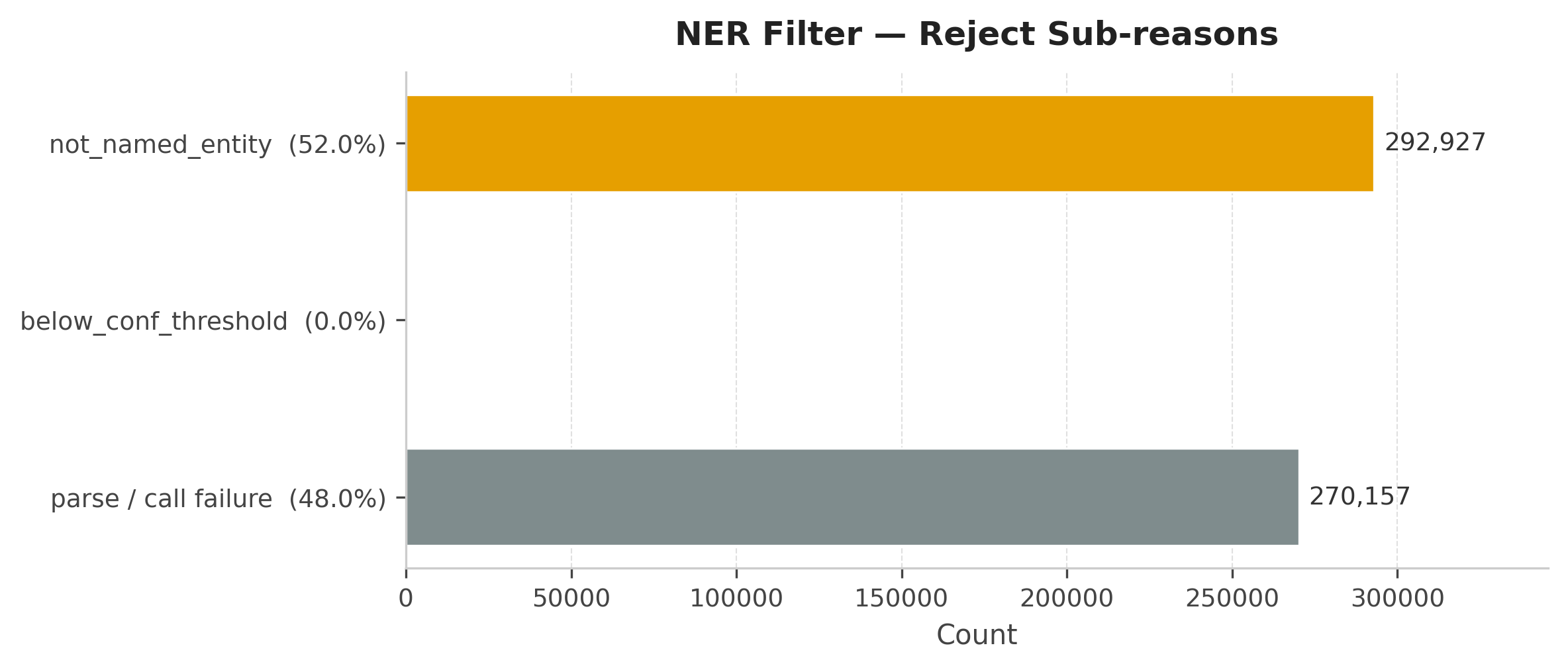}
  \caption{Llama NER reject reasons. \textbf{This breakdown is
  qualitatively different from the other two models.} Of 563K NER
  rejections, only 52.0\% (292{,}927) are legitimate
  \texttt{not\_named\_entity} verdicts; \textbf{48.0\% (270{,}157)
  are parse/call failures} - candidates the strict-gate policy
  treats as rejected because the NER call did not produce a parsable
  decision. This is the single biggest deviation in the pipeline
  behavior of any model. The strict-gate response is conservative
  (drop the candidate, do not pass through unchecked), but the
  underlying cause is Llama's lower instruction-following compliance
  on the structured-output schema.}
  \label{fig:lm_ner_reasons}
\end{figure}
 
\begin{figure}[h]
  \centering
  \includegraphics[width=0.95\linewidth]{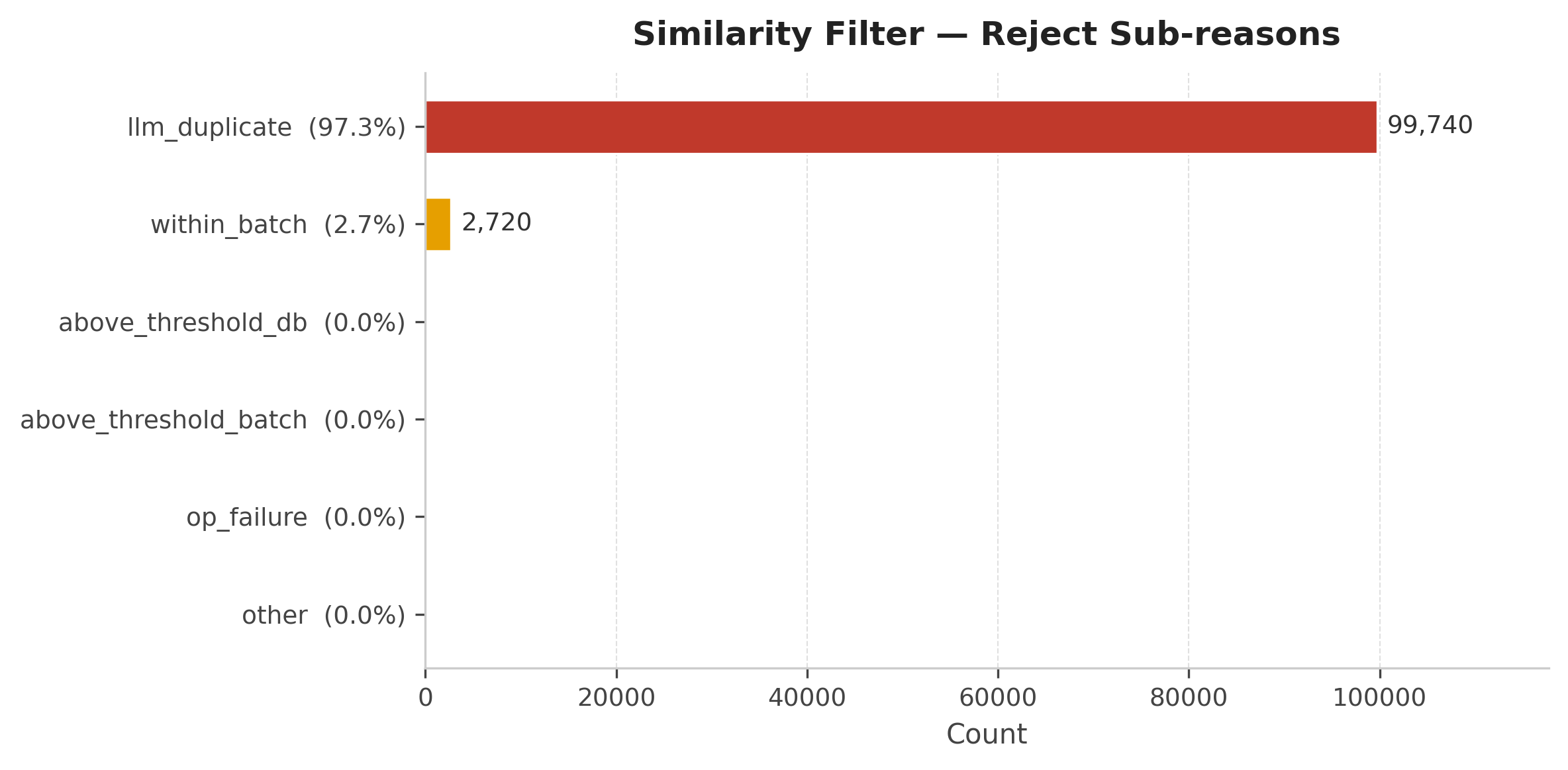}
  \caption{Llama similarity reject reasons. Despite the NER-stage
  difficulties, the similarity stage is well-behaved: 97.3\%
  LLM-confirmed duplicates, 2.7\% within-batch duplicates, and zero
  operational failures.}
  \label{fig:lm_sim_reasons}
\end{figure}
 
\begin{figure}[h]
  \centering
  \includegraphics[width=0.95\linewidth]{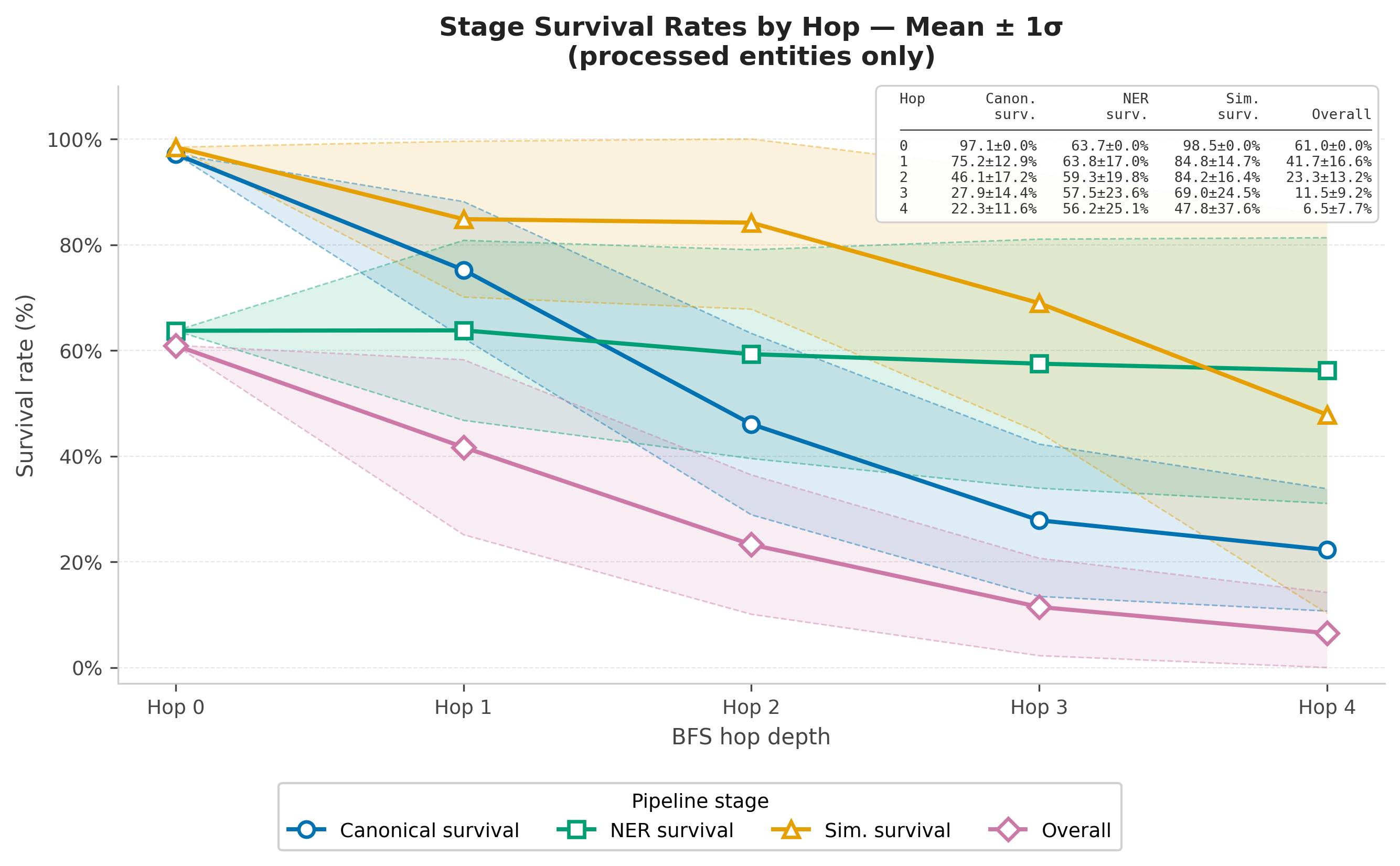}
  \caption{Llama per-hop survival rates with ${\pm}1\sigma$ bands. The similarity survival band widens noticeably at deep hops: the
per-entity mean is 69.0\% with $\sigma{=}24.5$ at hop~3 and
47.8\% with $\sigma{=}37.6$ at hop~4., reflecting the increased
  per-entity variance that the parse-failure mechanism introduces -
  some entities had nearly all candidates pass NER cleanly, others
  lost most of them to parse failures.}
  \label{fig:lm_survival_band}
\end{figure}
 
\begin{figure}[h]
  \centering
  \includegraphics[width=0.95\linewidth]{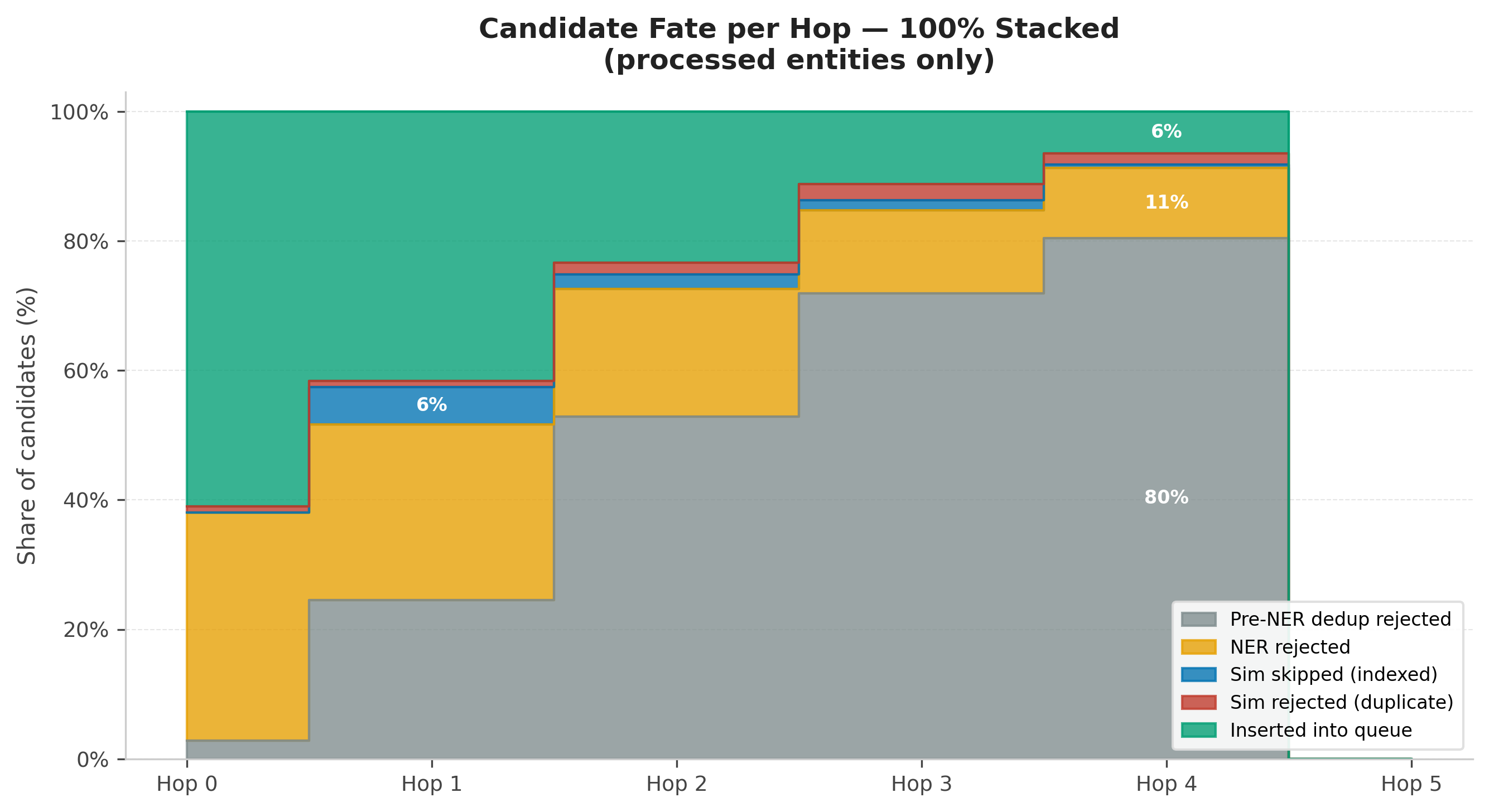}
  \caption{Llama candidate fate per hop, 100\% normalized. The orange
  ``NER rejected'' layer is visibly thicker than in the other two
  models, especially at hops~3--4 where parse failures cluster.}
  \label{fig:lm_fate_pct}
\end{figure}
 
\begin{figure}[h]
  \centering
  \includegraphics[width=0.95\linewidth]{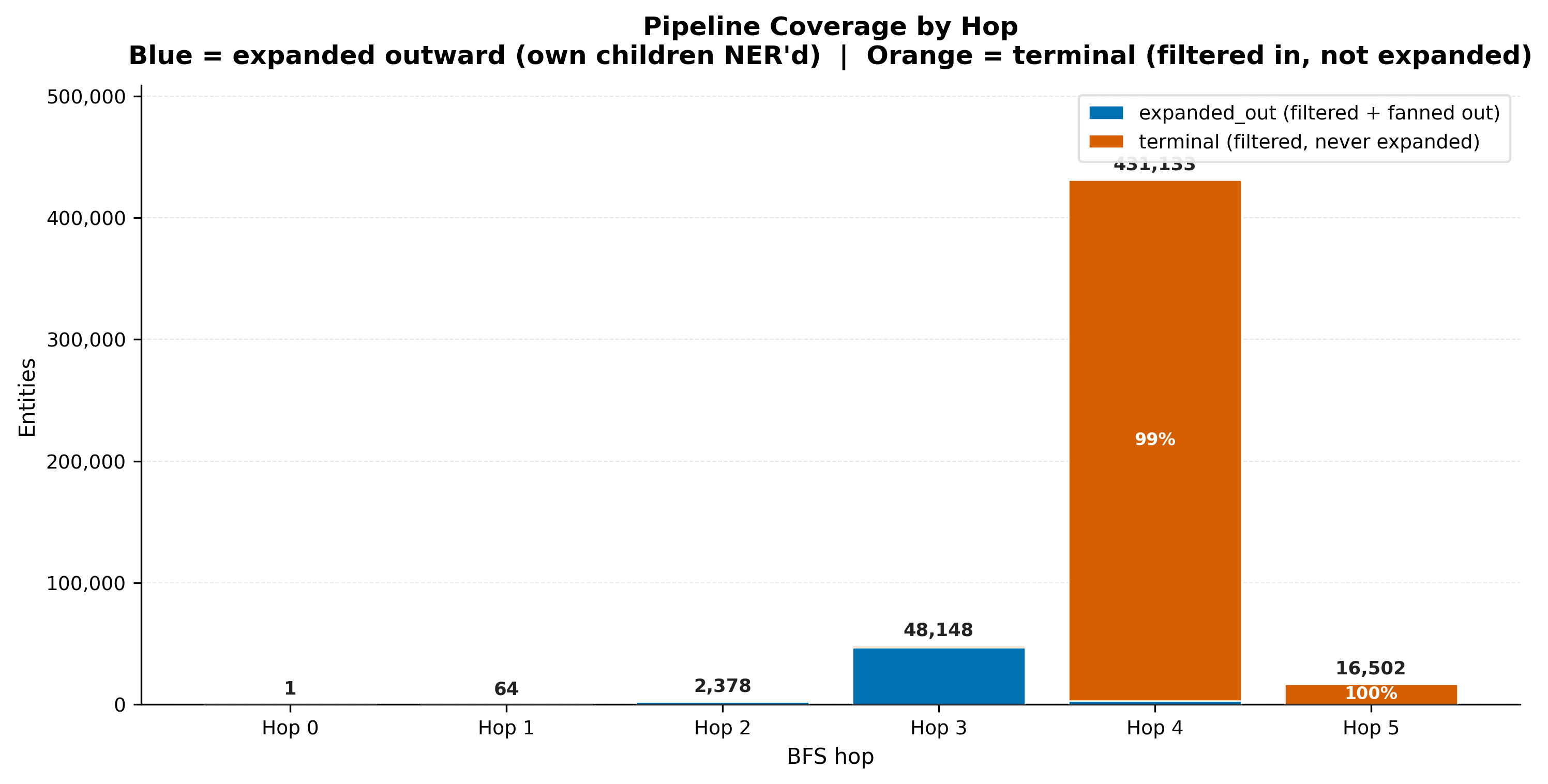}
  \caption{Llama pipeline coverage by hop. Mirrors the DeepSeek
  pattern: hop~4 entities are largely terminal (99.3\%) because the
  run reached its subject budget before they could be expanded.}
  \label{fig:lm_coverage_hop}
\end{figure}
 
\begin{figure}[h]
  \centering
  \includegraphics[width=0.95\linewidth]{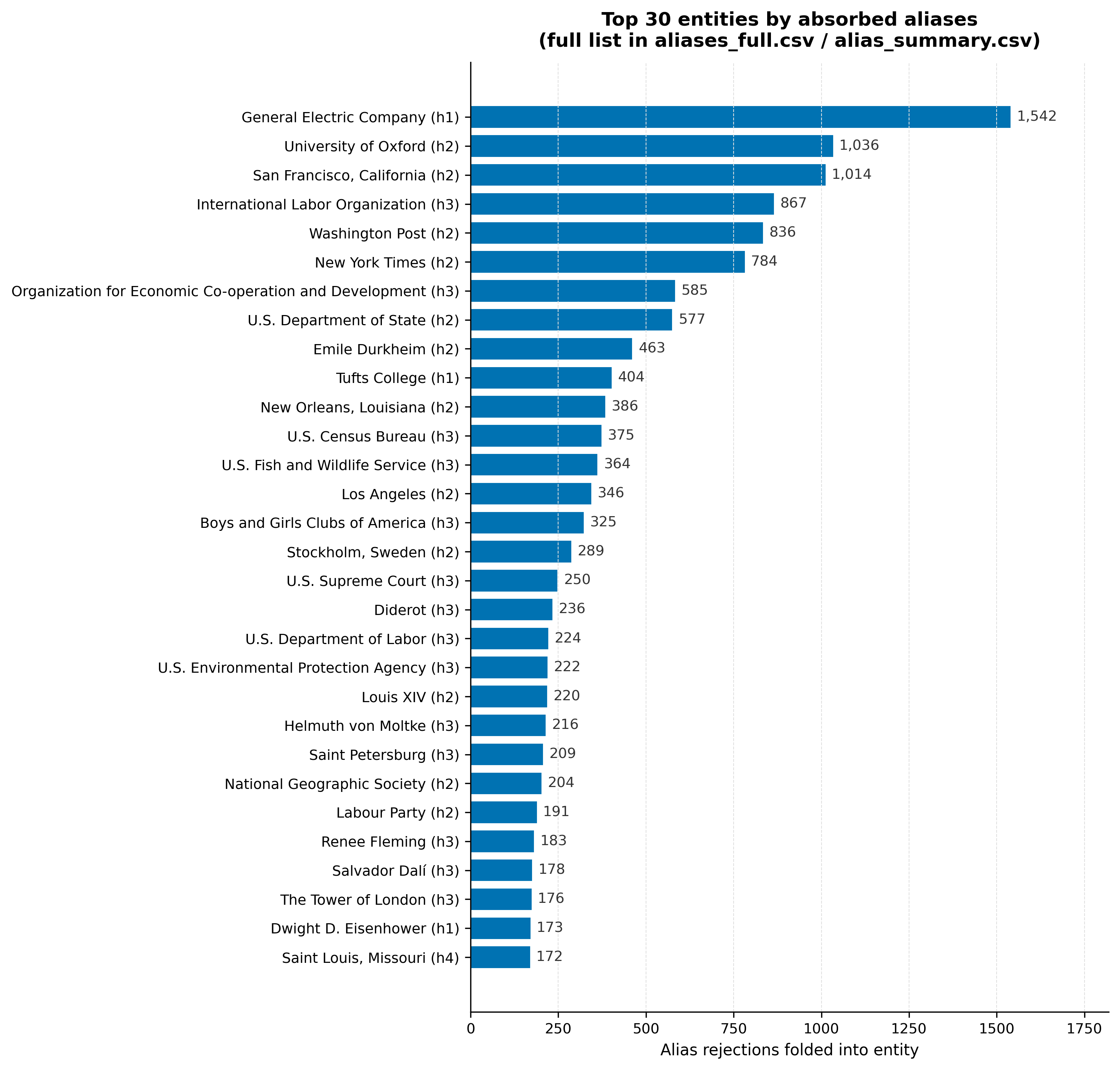}
  \caption{Top-30 Llama entities by aliases absorbed. \textit{General
  Electric Company} sits at the top with 1{,}542 absorbed aliases
  -  driven by the very common ``GE'' acronym variant -
  followed by \textit{University of Oxford} (1{,}036) and \textit{San
  Francisco} (1{,}014). The long tail of media outlets and
  international organizations is consistent with the other two
  models.}
  \label{fig:lm_aliases}
\end{figure}
 
\paragraph{NER parse failures.}
The 270{,}157 Llama parse failures in
Figure~\ref{fig:lm_ner_reasons} are the single largest deviation
between models. Inspection of the per-chunk parse mode log
(\texttt{ner\_decisions.jsonl}) shows that 99.7\% of NER \emph{calls}
were parsed as native JSON; the remaining 0.3\% recovered via
regex-based salvage or were treated as exceptions. The strict-gate
policy in \texttt{factuality\_core} rejects every candidate inside a
chunk whose model output did not produce a usable decision row,
which inflates the reject count without representing a quality
judgement about the candidate. This is a conservative but blunt
mechanism: a lower-compliance model's effective NER recall is
slightly underestimated, but the corpus is protected from candidates
that bypassed the filter. Llama's downstream similarity stage and
inserted-corpus quality remain comparable to the other two models
(Figure~\ref{fig:lm_sim_reasons}, Table~\ref{tab:funnel_summary_table_app}).
 
\subsection{Pipeline Coverage: Filtered-In vs.\ Expanded-Out}
\label{app:funnel_coverage}
 
A subject in the LLMpedia corpus has two distinct lifecycle states.
A subject is \emph{filtered-in} once it has passed Stage~1--3
sanitization as a \emph{candidate} produced by some parent article -
this is the precondition for receiving an article of its own. A
subject is \emph{expanded-out} once NER has run on its own outbound
wikilinks, producing candidates for the next BFS hop. Every
expanded-out subject is filtered-in by construction; the converse
does not hold. The gap between the two corresponds to subjects whose
article was generated but whose outbound expansion did not run, either
because the BFS reached its hop or subject-count limit, or because
the run was stopped.
 
\begin{table}[t]
\centering
\scriptsize
\setlength{\tabcolsep}{2pt}
\renewcommand{\arraystretch}{0.95}
\resizebox{\columnwidth}{!}{%
\begin{tabular}{lrrrr}
\toprule
\textbf{Model} & \textbf{Total} & \textbf{Filt.-in}
  & \textbf{Exp.-out} & \textbf{Terminal} \\
\midrule
GPT-5-mini & 1{,}063{,}949 & 1{,}063{,}930 & 174{,}100 & 83.6\% \\
DeepSeek   &   396{,}091 &   396{,}091 &  35{,}441 & 91.1\% \\
Llama-70B  &   498{,}226 &   498{,}226 &  52{,}557 & 89.5\% \\
\bottomrule
\end{tabular}%
}
\caption{Pipeline coverage at corpus level. Terminal percentage is the share of total entities that were filtered in but not expanded out. The high terminal share reflects each run reaching its computational ceiling at the deepest hop, not pipeline failure.}
\label{tab:funnel_coverage_global}
\end{table}
 
\begin{table}[h]
\centering
\scriptsize
\setlength{\tabcolsep}{3pt}
\renewcommand{\arraystretch}{1.06}
\begin{tabular}{ll rrr r}
\toprule
\textbf{Model} & \textbf{Hop} & \textbf{Total}
  & \textbf{Expanded} & \textbf{Terminal} & \textbf{Term.\,\%} \\
\midrule
\multirow{7}{*}{GPT-5-mini}
 & 0 & 1 & 1 & 0 & 0.0 \\
 & 1 & 58 & 58 & 0 & 0.0 \\
 & 2 & 2{,}344 & 2{,}344 & 0 & 0.0 \\
 & 3 & 40{,}923 & 40{,}923 & 0 & 0.0 \\
 & 4 & 321{,}319 & 89{,}836 & 231{,}475 & 72.0 \\
 & 5 & 507{,}940 & 40{,}938 & 466{,}996 & 91.9 \\
 & 6 & 191{,}364 & 0 & 191{,}359 & 100.0 \\
\midrule
\multirow{6}{*}{DeepSeek}
 & 0 & 1 & 1 & 0 & 0.0 \\
 & 1 & 36 & 36 & 0 & 0.0 \\
 & 2 & 1{,}133 & 1{,}133 & 0 & 0.0 \\
 & 3 & 23{,}572 & 23{,}552 & 20 & 0.1 \\
 & 4 & 276{,}033 & 10{,}719 & 265{,}314 & 96.1 \\
 & 5 & 95{,}316 & 0 & 95{,}316 & 100.0 \\
\midrule
\multirow{6}{*}{Llama-70B}
 & 0 & 1 & 1 & 0 & 0.0 \\
 & 1 & 64 & 64 & 0 & 0.0 \\
 & 2 & 2{,}378 & 2{,}378 & 0 & 0.0 \\
 & 3 & 48{,}148 & 46{,}931 & 1{,}217 & 2.5 \\
 & 4 & 431{,}133 & 3{,}183 & 427{,}950 & 99.3 \\
 & 5 & 16{,}502 & 0 & 16{,}502 & 100.0 \\
\bottomrule
\end{tabular}
\caption[Pipeline coverage by hop]{Pipeline coverage by hop. Terminality is concentrated at the
deepest hop of each run, which is the natural BFS boundary. Hops
$\leq 3$ are essentially fully expanded in every model.\protect\footnotemark}
\footnotetext{Total includes a small number of unprocessed entities at the deepest hops
(8 at GPT-5-mini hop 4; 6 at hop 5; 5 at hop 6; 1 at DeepSeek hop 2; etc.);
these are not counted as either expanded or terminal.}
\label{tab:funnel_coverage_hop}
\end{table}
 
\paragraph{Why this distinction matters.}
The headline factuality numbers in Table~\ref{tab:hop_factuality} and
the cross-model comparison in Table~\ref{tab:crossmodel_main} sample
from \emph{filtered-in} subjects: these are full articles, vetted by
NER and similarity, and constitute the actual LLMpedia corpus a
reader would consume. The hop-survival rates in
Appendix~\ref{app:funnel_combined} are computed over
\emph{expanded-out} subjects, since these are the only ones that
produced a measurable outbound funnel. Reporting both quantities
explicitly avoids two common misreadings: (i)~that the corpus is
smaller than reported because some entities did not fully traverse
the pipeline (it is not - filtered-in articles are real members of
the corpus), and (ii)~that survival rates somehow apply to subjects
that never received an outbound NER call (they do not).
 
\subsection{Alias Absorption}
\label{app:funnel_aliases}
 
Every Stage~3 similarity rejection records the canonical entity the
candidate was found to duplicate. These \emph{alias absorptions} are
the most direct measurable evidence that the disambiguation layer is
doing real work - they are the duplicate articles LLMpedia would
have produced if Stage~3 were disabled. The volume scales with corpus
size: GPT-5-mini absorbs the most aliases overall by an order of
magnitude.
 
\begin{table}[t]
\centering
\scriptsize
\setlength{\tabcolsep}{2pt}
\renewcommand{\arraystretch}{0.95}
\resizebox{\columnwidth}{!}{%
\begin{tabular}{lrrr}
\toprule
\textbf{Model} & \textbf{Ents. w/ aliases} & \textbf{Total aliases}
  & \textbf{Aliases/ent.} \\
\midrule
GPT-5-mini & 108{,}030 & 527{,}469 & 4.88 \\
DeepSeek   &  14{,}591 &  35{,}558 & 2.44 \\
Llama-70B  &  26{,}258 & 101{,}548 & 3.87 \\
\bottomrule
\end{tabular}%
}
\caption{Per-model alias absorption summary. Aliases/entity is averaged only over entities that absorbed at least one alias, i.e., the mean number of duplicate surface forms each magnetic entity received.}
\label{tab:funnel_aliases_summary}
\end{table}
 
\paragraph{Patterns in the top-absorbing entities.}
Across all three models the top-30 absorbed-alias lists are dominated
by:
(i)~U.S.\ federal departments (variants:
\textit{U.S.\ Department of X} / \textit{United States Department of
X} / \textit{Department of X (United States)} / \textit{US Department
of X});
(ii)~major universities (acronym variants and \textit{University of
X} / \textit{X University} alternations);
(iii)~well-known geographical locations (with and without state
qualifiers);
(iv)~prominent international organizations (full name vs.\ acronym).
These categories are exactly the ones where Stage~1 canonical keying
cannot resolve the variant space (the strings are too different) and
Stage~3 LLM arbitration is the only mechanism that can.
 
\paragraph{Magnetic-entity caveat.}
A small number of entities absorb a disproportionate share of
aliases. In the GPT-5-mini run, the top entity
(\textit{University of Oxford}) absorbed 1{,}814 aliases -
more aliases than the entire DeepSeek run produced for any single
canonical entity (top: \textit{University of Illinois at
Urbana--Champaign}, 198). Manual inspection of a sample of these
absorptions shows the bulk are correct collapses, but a small fraction
($<$2\% in the sample) are arguable: e.g.\ DeepSeek's
\textit{Charlestown, Massachusetts} entity absorbed both
\textit{Charlestown, Boston} and the bare \textit{Charlestown}, which
is reasonable in the U.S.\ historical context of the article that
nominated it but might in principle have justified separate
disambiguation pages. Errors of this form contribute to the corpus
slightly under-representing minor place names; they do not affect any
of the factuality numbers reported in the paper.
 
\subsection{Queue-Insert Races: A Small Residual Loss}
\label{app:funnel_race}
 
Even after Stage~3 similarity accepts a candidate, the candidate may
not enter the queue. Tracing the \texttt{sim\_worker\_loop} in the
implementation, three things can drop a similarity-accepted candidate
between Post-sim and Inserted: a depth-cap rejection
(\texttt{hop+1\,>\,max\_depth}), a max-subjects-cap rejection (the
main queue refuses new rows), or a canonical-key race (two workers
sim-accepted the same surface form before either committed; only one
wins). For the unbounded-depth runs reported in this paper, only the
third cause applies.
 
\begin{table}[h]
\centering
\small
\setlength{\tabcolsep}{4pt}
\renewcommand{\arraystretch}{1.08}
\begin{tabular}{l rr r}
\toprule
\textbf{Model} & \textbf{Sim-accepted} & \textbf{Inserted}
  & \textbf{Race loss\,(\%)} \\
\midrule
GPT-5-mini & 1{,}088{,}968 & 1{,}063{,}923 & 2.3\% \\
DeepSeek   &    396{,}254 &    395{,}336 & 0.2\% \\
Llama-70B  &    499{,}245 &    497{,}792 & 0.3\% \\
\bottomrule
\end{tabular}
\caption{Residual queue-insert race rate per model. All three runs
stay well under 3\%, with the open-weight runs near the noise floor.
GPT-5-mini's slightly higher rate reflects its larger absolute
inflight load: more concurrent workers across more hops mean
proportionally more canonical-key collisions per wave.}
\label{tab:funnel_race}
\end{table}
 
\paragraph{Implication for the correctness claim.}
The deduplication-correctness argument in
Appendix~\ref{app:dedup-guarantee} establishes that the main article
queue is structurally duplicate-free at all times. The race loss
reported in Table~\ref{tab:funnel_race} is the cost of enforcing that
guarantee under parallelism: when two workers reach Stage~3 with the
same canonical key in the same wave, exactly one of them wins the
atomic commit and the other is silently dropped. This is bounded,
small, and exactly what a correct dedup implementation must do - it
is not a hidden source of corpus loss.

\section{Cross-Model Analysis: Supplementary Figures}
\label{app:crossmodel-figures}

\begin{figure}[h]
  \centering
  \includegraphics[width=0.75\linewidth]{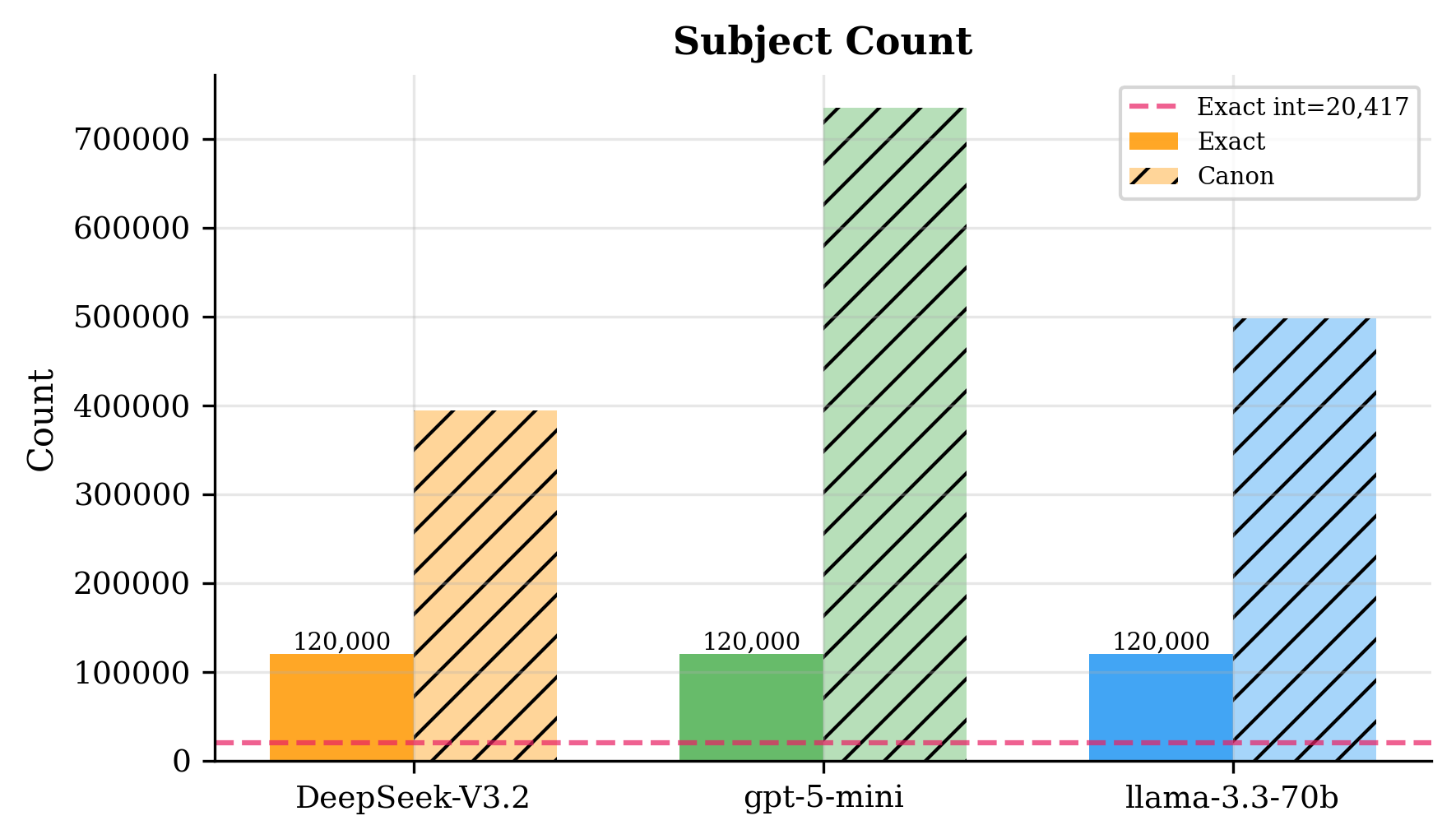}
  \caption{Subject counts with three-way intersection (7.3\% overlap).}
  \label{fig:app_cm_subjects}
\end{figure}
\begin{figure}[h]
  \centering
  \includegraphics[width=0.70\linewidth]{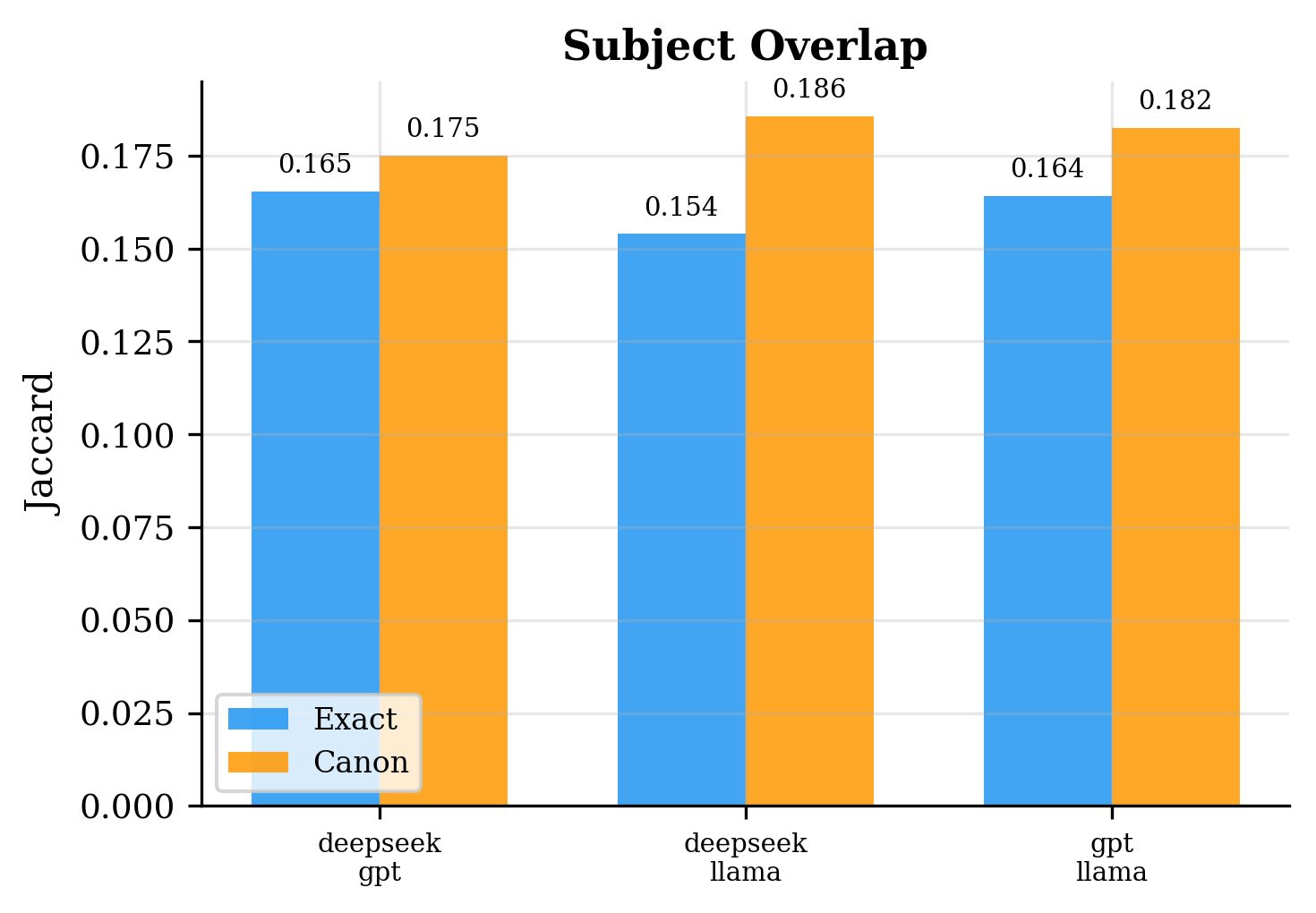}
  \caption{Pairwise subject Jaccard (0.15--0.17).}
  \label{fig:app_cm_overlap}
\end{figure}
\begin{figure}[h]
  \centering
  \includegraphics[width=0.55\linewidth]{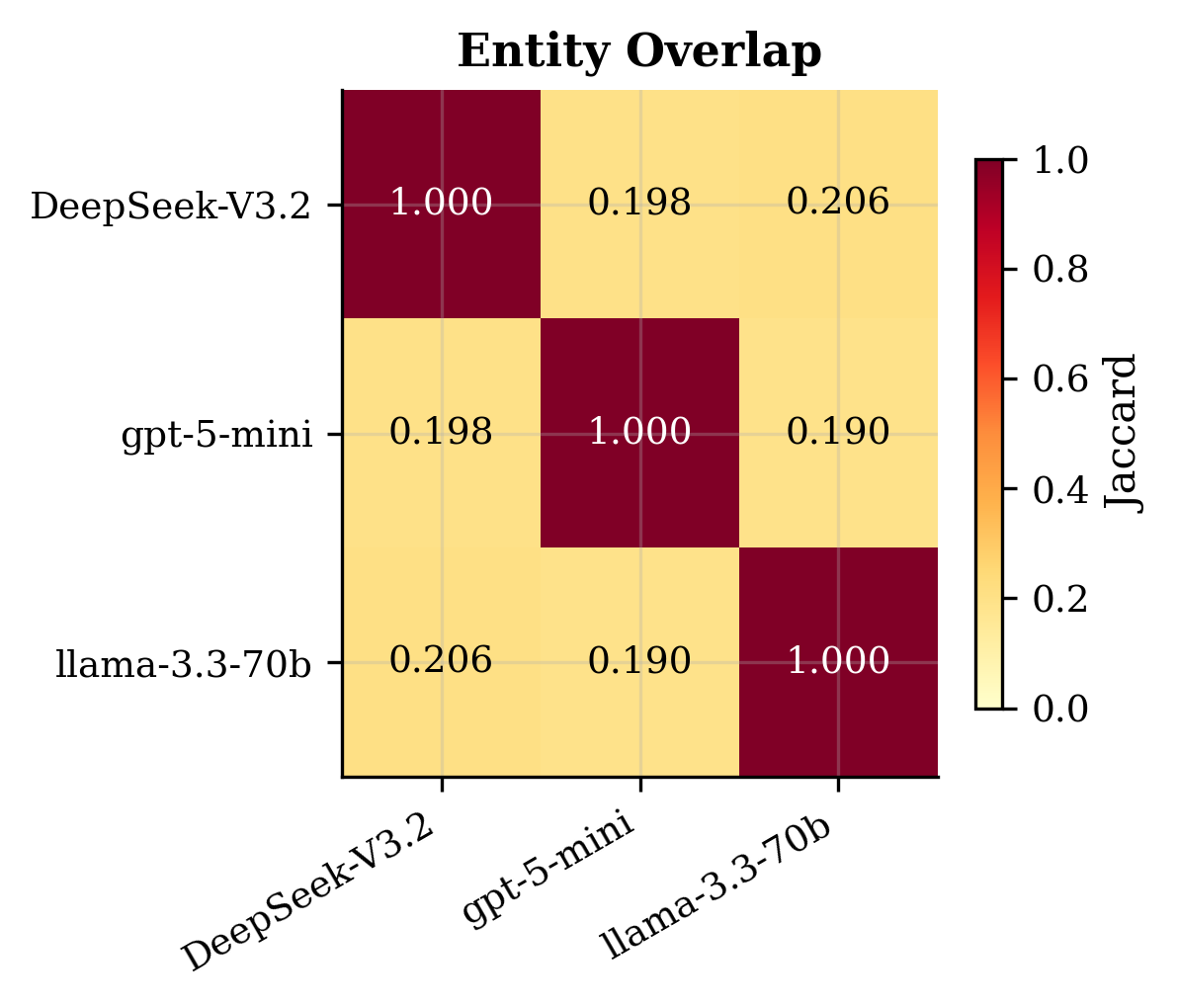}
  \caption{Entity overlap on 1{,}000 shared subjects (Jaccard 0.19--0.21).}
  \label{fig:app_cm_entity}
\end{figure}
\begin{figure}[h]
  \centering
  \includegraphics[width=0.85\linewidth]{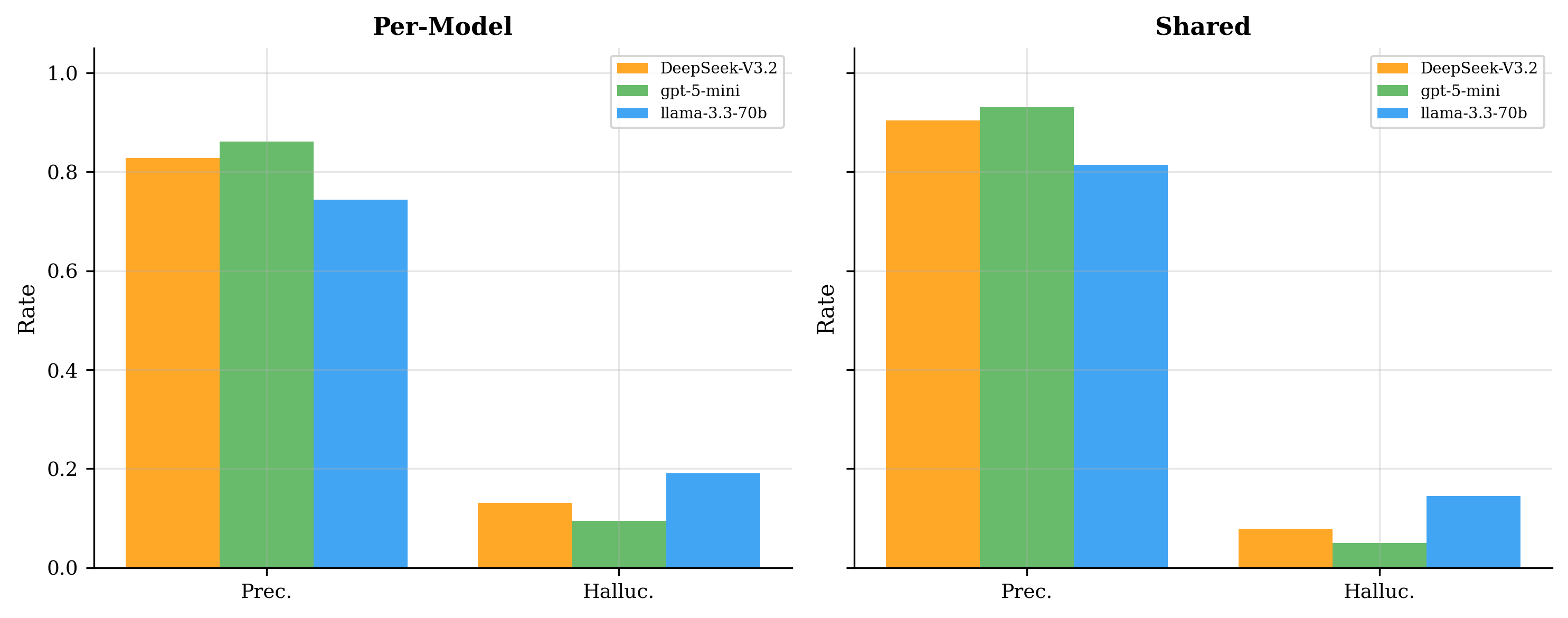}
  \caption{Per-model factuality: shared (left) vs.\ independent (right).}
  \label{fig:app_cm_factuality}
\end{figure}
\begin{figure}[h]
  \centering
  \includegraphics[width=0.85\linewidth]{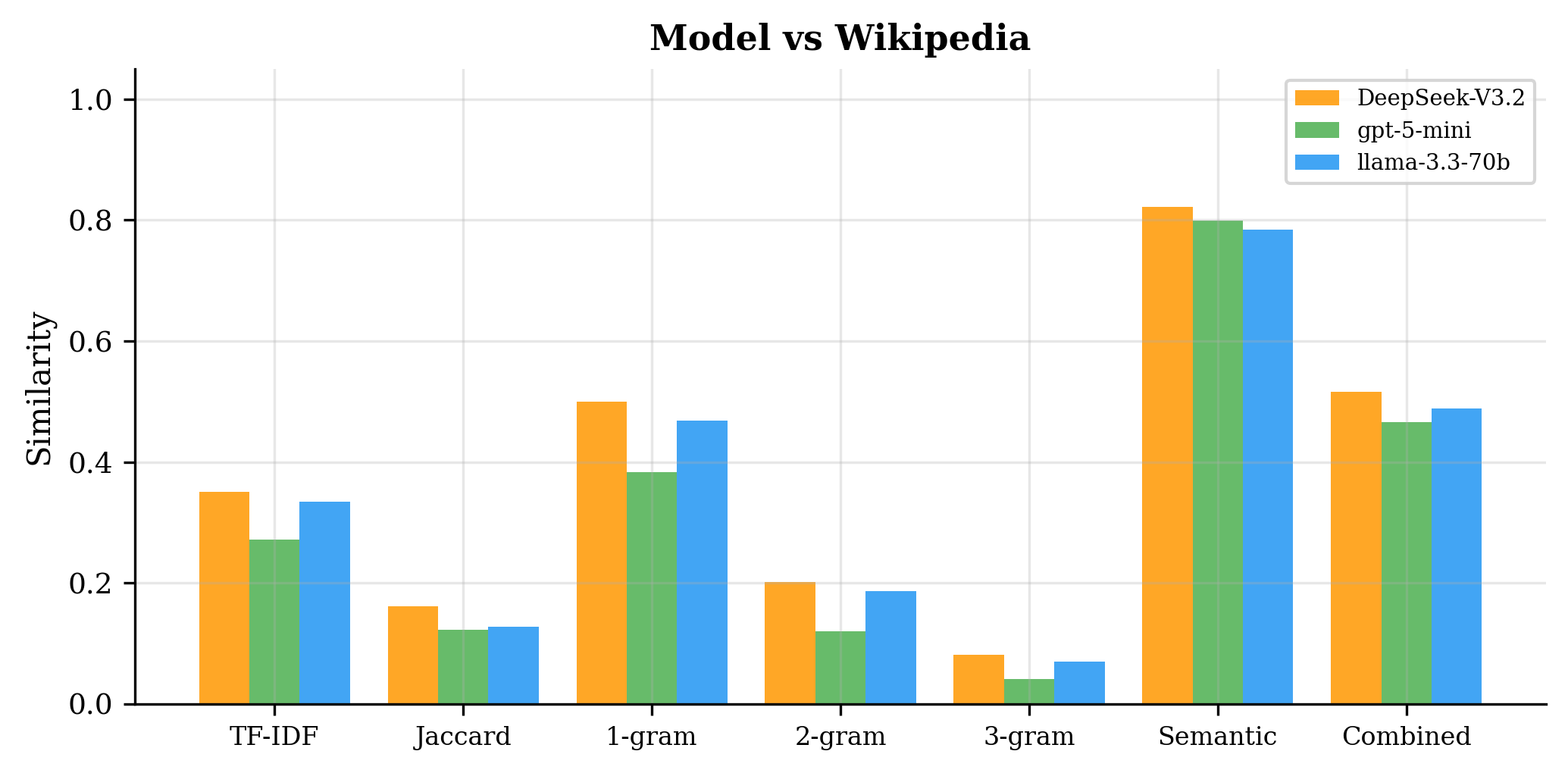}
  \caption{Model-to-Wikipedia similarity. All three models below Grokipedia's
  0.455--0.489 TF-IDF~\cite{yasseri2025similar}.}
  \label{fig:app_cm_wikisim}
\end{figure}

\section{Topic-Focused: Supplementary Tables}
\label{app:topic-supplementary}

Precision differences between any two personas within the same model--topic
cell are $\leq$3.6\,pp, confirming that persona changes framing rather than
factual accuracy.

\begin{table*}[t]
\centering
\small
\setlength{\tabcolsep}{2pt}
\renewcommand{\arraystretch}{1.05}
\begin{tabular}{llccccccc}
\toprule
\textbf{Topic} & \textbf{Model} & \textbf{Persona} & \textbf{$n$}
  & \textbf{Prec} & \textbf{Hall} & \textbf{Unv} & \textbf{W\%} & \textbf{Wds} \\
\midrule
A.\ Bab. & GPT & cons & 67 & 99.3 & 0.6 & 15.9 & 67.0 & 765 \\
A.\ Bab. & GPT & left & 75 & 99.3 & 0.4 & 20.1 & 75.0 & 785 \\
A.\ Bab. & GPT & neut & 71 & 99.5 & 0.4 & 20.4 & 71.0 & 775 \\
A.\ Bab. & DS  & cons & 81 & 98.8 & 1.0 & 20.0 & 81.0 & 834 \\
A.\ Bab. & DS  & left & 78 & 97.8 & 1.8 & 20.4 & 78.0 & 908 \\
A.\ Bab. & DS  & neut & 68 & 95.3 & 2.7 & 33.5 & 68.0 & 600 \\
A.\ Bab. & Llama & cons & 81 & 94.1 & 3.1 & 26.3 & 81.0 & 964 \\
A.\ Bab. & Llama & left & 68 & 94.6 & 2.5 & 30.2 & 68.0 & 982 \\
A.\ Bab. & Llama & neut & 90 & 92.4 & 3.2 & 28.6 & 90.0 & 957 \\
\midrule
D.\ Col. & GPT & cons & 67 & 99.3 & 0.6 & 19.4 & 67.0 & 821 \\
D.\ Col. & GPT & left & 62 & 98.9 & 0.7 & 27.9 & 62.0 & 830 \\
D.\ Col. & GPT & neut & 70 & 98.2 & 1.1 & 22.1 & 70.0 & 815 \\
D.\ Col. & DS  & cons & 73 & 96.0 & 2.3 & 26.2 & 73.0 & 870 \\
D.\ Col. & DS  & left & 73 & 97.0 & 2.1 & 25.1 & 73.0 & 1{,}119 \\
D.\ Col. & DS  & neut & 65 & 94.7 & 2.5 & 23.3 & 65.0 & 864 \\
D.\ Col. & Llama & cons & 57 & 96.7 & 2.3 & 33.0 & 57.0 & 1{,}025 \\
D.\ Col. & Llama & left & 71 & 94.7 & 2.4 & 39.2 & 71.0 & 1{,}018 \\
D.\ Col. & Llama & neut & 54 & 94.2 & 3.0 & 33.2 & 54.0 & 1{,}019 \\
\midrule
US CR & GPT & cons & 84 & 97.9 & 1.9 & \phantom{0}9.2 & 84.0 & 830 \\
US CR & GPT & left & 79 & 99.0 & 1.0 & \phantom{0}9.8 & 79.0 & 853 \\
US CR & GPT & neut & 72 & 98.0 & 1.5 & 13.6 & 72.0 & 861 \\
US CR & DS  & cons & 87 & 97.2 & 2.1 & 17.0 & 87.0 & 905 \\
US CR & DS  & left & 80 & 97.2 & 2.4 & 15.0 & 80.0 & 880 \\
US CR & DS  & neut & 82 & 96.3 & 2.3 & 20.2 & 82.0 & 905 \\
US CR & Llama & cons & 76 & 95.1 & 3.2 & 25.1 & 76.0 & 983 \\
US CR & Llama & left & 80 & 97.3 & 1.8 & 20.8 & 80.0 & 1{,}004 \\
US CR & Llama & neut & 70 & 97.1 & 2.1 & 20.6 & 70.0 & 973 \\
\bottomrule
\end{tabular}
\caption{Full per-persona factuality for all 27 topic-focused conditions.
$n$ = Wikipedia-covered subjects. W\% = Wikipedia coverage.
Wds = mean word count. Hall = false rate (hallucination).}
\label{tab:full_persona_results}
\end{table*}

\section{Hop-Stratified Analysis: Supplementary}
\label{app:hop-supplementary}

Precision stays nearly flat (97.9\% at hop~1 $\to$ 96.5\% at hop~6)
while the unverifiable rate grows monotonically (4\% $\to$ 43\%) and
the false rate remains consistently low ($<$2\%).
This confirms that the main degradation at BFS depth is thinning
external evidence coverage, not increasing hallucination.

\section{Prompt Templates}
\label{app:prompts}

LLMPedia operates under two \textbf{generation regimes} and two
\textbf{elicitation strategies}, yielding four prompt configurations.
The \textbf{general-domain} regime performs open BFS expansion from an
arbitrary seed with no topical constraint. The \textbf{topic-focused}
regime also performs BFS but restricts expansion to entities with a
direct, meaningful connection to a designated root topic
(e.g., \textit{Ancient Babylon}). Orthogonally, \textbf{baseline}
prompts request standard Wikitext with plain \texttt{[[wikilinks]]},
while \textbf{calibrated} prompts additionally require each wikilink to
be annotated with a confidence score (\texttt{[[Entity (0.85)]]})
reflecting the model's certainty that the entity belongs to the
subject's semantic neighborhood; links below threshold $\tau=0.75$ are
discarded by the NER stage before reaching similarity arbitration.

Each subject passes through two prompt roles in sequence:
\begin{enumerate}[noitemsep, topsep=3pt]
  \item \textbf{Article elicitation}: generates the full Wikitext
    article following a pre-generated section outline. The calibrated
    variant requires scored wikilinks.
  \item \textbf{NER filtering}: classifies each raw wikilink extracted
    from the article as a true named entity worthy of BFS expansion or
    a reject. The calibrated variant additionally assigns confidence
    scores.
\end{enumerate}

\paragraph{Runtime placeholders.}
All prompts share the following runtime-filled placeholders:

\begin{itemize}[noitemsep, topsep=3pt]
  \item \texttt{\{subject\_name\}}: the entity currently being
    processed, e.g.\ \textit{Vannevar Bush}.
  \item \texttt{\{root\_subject\}}: the root topic constraining
    topic-focused runs, e.g.\ \textit{Ancient Babylon}; absent in
    general-domain prompts.
  \item \texttt{\{avg\_words\_per\_article\}}: target article length in
    words (default 716, matching Wikipedia's overall
    average~\cite{WikipediaSizeWikipedia}).
  \item \texttt{\{outline\_block\}}: the JSON section list produced by
    the outline generation step and injected verbatim into elicitation
    prompts; see Example~\ref{ex:outline} below.
  \item \texttt{\{phrases\_block\}}: the newline-separated list of raw
    wikilink candidates extracted from the generated article, passed
    to NER filtering; see Example~\ref{ex:ner} below.
  \item \texttt{\{persona\_block\}}: one of three fixed instruction
    strings injected at the system level of \emph{every} pipeline
    stage-elicitation, NER, and similarity arbitration-to probe
    how editorial stance shapes content selection and framing
    independently of subject matter; see Example~\ref{ex:persona}
    below.
\end{itemize}

\paragraph{Example~1: \texttt{\{outline\_block\}.}}
\label{ex:outline}
The outline step calls the model once per subject and produces a
subject-tailored JSON section list. The output is injected verbatim
into the elicitation prompt; section titles are used \emph{exactly} as
\texttt{== Heading ==} markers and the model is forbidden from adding,
removing, or reordering them. This enforces structural consistency
across models and experimental conditions while allowing content that
adapts to the subject type-a historical figure yields biography-oriented
sections; a chemical compound or a city would yield an entirely
different structure.

\vspace{4pt}
\noindent\colorbox{gray!8}{\parbox{0.95\linewidth}{\small
\textbf{Subject:} \textit{Vannevar Bush} \quad
\textbf{Regime:} general-domain\\[4pt]
\textbf{Filled \texttt{\{outline\_block\}}:}\\[2pt]
\texttt{\{"sections": [}\\
\texttt{\quad "Early Life and Education",}\\
\texttt{\quad "Engineering Career and Raytheon",}\\
\texttt{\quad "Wartime Science Leadership and the OSRD",}\\
\texttt{\quad "As We May Think and the Memex",}\\
\texttt{\quad "National Science Foundation",}\\
\texttt{\quad "Legacy and Influence"}\\
\texttt{]\}}
}}

\vspace{4pt}
\noindent\colorbox{gray!8}{\parbox{0.95\linewidth}{\small
\textbf{Subject:} \textit{Hammurabi} \quad
\textbf{Regime:} topic-focused (\texttt{\{root\_subject\}} =
\textit{Ancient Babylon})\\[4pt]
\textbf{Filled \texttt{\{outline\_block\}}:}\\[2pt]
\texttt{\{"sections": [}\\
\texttt{\quad "Rise to Power in Babylon",}\\
\texttt{\quad "Military Campaigns and Territorial Expansion",}\\
\texttt{\quad "The Code of Hammurabi",}\\
\texttt{\quad "Administrative and Economic Reforms",}\\
\texttt{\quad "Role within Ancient Babylon",}\\
\texttt{\quad "Death and Legacy"}\\
\texttt{]\}}\\[2pt]
\textit{In topic-focused runs, the outline step is instructed to
include one section explicitly connecting the subject to
\texttt{\{root\_subject\}}; here, ``Role within Ancient Babylon''
fulfils that requirement.}
}}

\paragraph{Example~2: \texttt{\{phrases\_block\}.}}
\label{ex:ner}
After article generation, all \texttt{[[wikilinks]]} are extracted via
regex and passed to the NER stage as a newline-separated list. The NER
prompt receives this list as \texttt{\{phrases\_block\}} and must
classify each phrase as a true named entity deserving standalone
expansion or a reject. The example below illustrates the funnel:
generic role labels and loop forms are rejected; proper named entities
with encyclopedic scope are accepted.

\vspace{4pt}
\noindent\colorbox{gray!8}{\parbox{0.95\linewidth}{\scriptsize
\textbf{Subject:} \textit{Vannevar Bush} \quad
\textbf{Strategy:} baseline\\[4pt]
\textbf{Filled \texttt{\{phrases\_block\}} (excerpt):}\\[2pt]
\texttt{engineer}\\
\texttt{inventor}\\
\texttt{science administrator}\\
\texttt{Office of Scientific Research and Development}\\
\texttt{World War II}\\
\texttt{As We May Think}\\
\texttt{memex}\\
\texttt{Tufts University}\\
\texttt{Massachusetts Institute of Technology}\\
\texttt{History of Vannevar Bush}\\[4pt]
\textbf{Expected NER output (baseline):}\\[2pt]
\texttt{\{"phrases": [}\\
\texttt{\quad \{"phrase": "engineer",\, "is\_ne": false\},}\\
\texttt{\quad \{"phrase": "inventor",\, "is\_ne": false\},}\\
\texttt{\quad \{"phrase": "science administrator",\, "is\_ne": false\},}\\
\texttt{\quad \{"phrase": "Office of Scientific Research and}\\
\texttt{\quad\quad Development",\, "is\_ne": true\},}\\
\texttt{\quad \{"phrase": "World War II",\, "is\_ne": true\},}\\
\texttt{\quad \{"phrase": "As We May Think",\, "is\_ne": true\},}\\
\texttt{\quad \{"phrase": "memex",\, "is\_ne": true\},}\\
\texttt{\quad \{"phrase": "Tufts University",\, "is\_ne": true\},}\\
\texttt{\quad \{"phrase": "Massachusetts Institute of Technology",}\\
\texttt{\quad\quad "is\_ne": true\},}\\
\texttt{\quad \{"phrase": "History of Vannevar Bush",\, "is\_ne": false\}}\\
\texttt{]\}}\\[2pt]
\textit{Generic roles (``engineer'', ``inventor'', ``science
administrator'') and loop forms (``History of Vannevar Bush'') are
mandatory rejects; all surviving entities are proper named subjects
with encyclopedic scope. Under the calibrated strategy each entry
additionally carries a \texttt{"confidence"} score and entries below
$\tau=0.75$ are rejected before reaching the similarity stage.}
}}

\paragraph{Example~3: \texttt{\{persona\_block\}.}}
\label{ex:persona}
The persona string is injected at the system level of \emph{every}
pipeline stage-elicitation, NER, and similarity arbitration-so
that editorial stance propagates uniformly into article prose, entity
selection, and disambiguation decisions alike. This design isolates
the effect of framing from structural generation quality and enables
the controlled persona comparison reported in
\S\ref{sec:results-topic}.

\vspace{4pt}
\noindent\colorbox{gray!8}{\parbox{0.95\linewidth}{\small
\textbf{\texttt{scientific\_neutral} (default):}\\
\texttt{Write in a neutral, evidence-based register. Acknowledge
uncertainty where it exists. Prefer empirically grounded claims.
Avoid advocacy language.}\\[6pt]
\textbf{\texttt{left\_leaning}:}\\
\texttt{Foreground issues of structural inequality and marginalized
perspectives. Emphasize social impact and access. Frame developments
in terms of equity, power, and underrepresented voices.}\\[6pt]
\textbf{\texttt{conservative}:}\\
\texttt{Emphasize institutional continuity, traditional values, and
established social order. Frame developments in terms of stability,
heritage, and the preservation of proven structures.}\\[4pt]
\textit{All three persona variants share identical structural prompts;
only the framing and evaluative register differ. Because the persona
is injected at every stage, it influences not only article prose but
also what entities the NER stage judges encyclopedically noteworthy,
enabling LLMPedia to probe how the same parametric knowledge is
selected and framed under different editorial orientations.}
}}

\subsection*{A.1\quad General-Domain Prompts}

\paragraph{Elicitation - Baseline.}

\noindent\colorbox{orange!8}{\parbox{0.95\linewidth}{\small
\textbf{System:}
You are LLMPedia, an elite encyclopedia writer.
Persona: \texttt{\{persona\_block\}}.

Write about \texttt{\{subject\_name\}}
($\sim$\texttt{\{avg\_words\_per\_article\}} words).

\textsc{Structure:} Start with \texttt{\{\{Infobox ...\}\}} if
appropriate. First text line: \texttt{'''\{subject\_name\}'''}.
Follow with 2--4 sentence lead. Use EACH outline title exactly as
\texttt{== Heading ==} in order.

\textsc{Wikilinks (critical):}
Famous subjects: 50--100 distinct \texttt{[[links]]}; lesser-known:
15--30. Minimum 3--5 links per sentence, woven into prose naturally.
Only proper nouns: people, organizations, places, events, battles,
treaties, works, awards, laws, institutions.
\textit{Forbidden:} generic concepts (``government'', ``economy'',
``education'', ``military''), common nouns, disciplines.

\textsc{No loops / aliases (strict):}
Never link \texttt{\{subject\_name\}} or variants.
Never link \texttt{[[X of \{subject\_name\}]]},
\texttt{[[\{subject\_name\}'s X]]},
\texttt{[[Part of \{subject\_name\}]]},
\texttt{[[History of \{subject\_name\}]]}.

No \texttt{<ref>}, no URLs, no References section.
Optional \texttt{[[Category:...]]}.\\[4pt]
\textbf{User:}
Subject: \texttt{\{subject\_name\}}

Section titles (use exactly):\\
\texttt{\{outline\_block\}}
}}

\vspace{8pt}
\paragraph{NER Filtering - Baseline.}

\noindent\colorbox{red!6}{\parbox{0.95\linewidth}{\small
\textbf{System:}
You are LLMPedia's NER module. Your role: capture which subjects
deserve their own encyclopedia article.
Persona: \texttt{\{persona\_block\}}.

\textsc{Accept only if both:}
(1) True named entity (person, organization, place, event, work, law,
award, institution), \textit{and}
(2) Worthy of a standalone encyclopedia entry-notable enough that
readers would want to learn about it.

\textsc{Reject:}
\texttt{\{subject\_name\}} itself, any alias, or variant.
Loop patterns: ``X of \texttt{\{subject\_name\}}'',
``\texttt{\{subject\_name\}}'s X'',
``History of \texttt{\{subject\_name\}}'',
``Part of \texttt{\{subject\_name\}}''.
Generic nouns, roles, concepts, dates, filler.

Output exactly:
\texttt{\{"phrases": [\{"phrase": "<exact input>",
"is\_ne": true/false\}, ...]\}}\\[4pt]
\textbf{User:}
Subject: \texttt{\{subject\_name\}}

Candidates:\\
\texttt{\{phrases\_block\}}
}}

\vspace{10pt}
\paragraph{Elicitation - Calibrated.}

\noindent\colorbox{orange!15}{\parbox{0.95\linewidth}{\small
\textbf{System:}
You are LLMPedia, an elite encyclopedia writer.
Persona: \texttt{\{persona\_block\}}.

Write about \texttt{\{subject\_name\}}
($\sim$\texttt{\{avg\_words\_per\_article\}} words).

\textsc{Structure:} Start with \texttt{\{\{Infobox ...\}\}} if
appropriate. First text line: \texttt{'''\{subject\_name\}'''}.
Follow with 2--4 sentence lead. Use EACH outline title exactly as
\texttt{== Heading ==} in order.

\textsc{Calibrated wikilinks (critical):}
Famous subjects: 50--100 distinct \texttt{[[links (score)]]};
lesser-known: 15--30. Minimum 3--5 links per sentence.
Format \textit{must} be \texttt{[[Proper Noun (score)]]},
e.g.\ \texttt{[[Albert Einstein (0.97)]]}.
Score = confidence that the entity belongs to the semantic
neighborhood of the subject:
0.95--1.00 extremely confident; 0.85--0.95 confident;
0.75--0.85 plausible but uncertain.
Only proper nouns: people, organizations, places, events, battles,
treaties, works, awards, laws, institutions.
\textit{Forbidden:} generic concepts, common nouns, disciplines.

\textsc{No loops / aliases (strict):}
Never link \texttt{\{subject\_name\}} or variants.
Never link loop forms
(e.g.\ \texttt{[[Churchill's speeches]]},
\texttt{[[Life of Churchill]]}).

No \texttt{<ref>}, no URLs, no References section.
Optional \texttt{[[Category:...]]}.\\[4pt]
\textbf{User:}
Subject: \texttt{\{subject\_name\}}

Section titles (use exactly):\\
\texttt{\{outline\_block\}}
}}

\vspace{8pt}
\paragraph{NER Filtering - Calibrated.}

\noindent\colorbox{red!12}{\parbox{0.95\linewidth}{\small
\textbf{System:}
You are LLMPedia's NER module.
Persona: \texttt{\{persona\_block\}}.

\textsc{Accept only if both:}
(1) True named entity (person, organization, place, event, work, law,
award, institution), \textit{and}
(2) Worthy of a standalone encyclopedia entry.

\textsc{Reject:}
\texttt{\{subject\_name\}} itself, any alias, or variant.
Loop patterns, generic nouns, roles, concepts, dates, filler.

\textsc{Calibrated output:}
Include a score (0--1) per candidate reflecting confidence that it is
(a) a true named entity \textit{and} (b) deserving a standalone article.
0.95--1.00 extremely confident; 0.85--0.95 confident;
0.75--0.85 plausible but uncertain; below 0.75 = reject.

Output exactly (valid JSON, no extra keys, no prose):
\texttt{\{"phrases": [\{"phrase": "<exact input>", "is\_ne":
true/false, "confidence": 0.0\}, ...]\}}\\[4pt]
\textbf{User:}
Subject: \texttt{\{subject\_name\}}

Candidates:\\
\texttt{\{phrases\_block\}}
}}

\subsection*{A.2\quad Topic-Focused Prompts}

\paragraph{Elicitation - Baseline.}

\noindent\colorbox{yellow!10}{\parbox{0.95\linewidth}{\small
\textbf{System:}
You are LLMPedia, generating concise Wikipedia-style articles in
Wikitext for a knowledge base built around ROOT TOPIC
\texttt{\{root\_subject\}}.
Persona: \texttt{\{persona\_block\}}.

Write a clean Wikipedia-style article about \texttt{\{subject\_name\}}
($\sim$\texttt{\{avg\_words\_per\_article\}} words).
First line: \texttt{'''\{subject\_name\}'''}.
Short lead (2--4 sentences) explaining what \texttt{\{subject\_name\}}
is and why it matters in the context of \texttt{\{root\_subject\}}.
Include \texttt{\{\{Infobox ...\}\}} if appropriate.

\textsc{Structure:} Use EACH section title exactly as
\texttt{== Title ==} in the given order. Do not add, remove, rename,
merge, or reorder sections. Write professional, neutral paragraphs
under every heading, focused on what is relevant to
\texttt{\{subject\_name\}} within \texttt{\{root\_subject\}}.

\textsc{Wikilinks:} Well-known subjects: 30--50 distinct
\texttt{[[wikilinks]]}; obscure: 8--15.
At least $\sim$70\% of distinct links must be specific named entities
or named works (people, organizations, institutions, companies, places,
events, programs, projects, products, papers/books, awards,
conferences, laws/policies).
Broad umbrella concepts allowed only if central and $\leq$3 total.
Wikilink the first mention of notable items strongly related to
\textit{both} \texttt{\{subject\_name\}} and \texttt{\{root\_subject\}}.
Never link \texttt{\{subject\_name\}}, \texttt{\{root\_subject\}}, or
trivial variants. No list/meta pages, no generic terms.

No \texttt{<ref>}, no URLs, no References section.
Optional \texttt{[[Category:...]]}.\\[4pt]
\textbf{User:}
Subject: \texttt{\{subject\_name\}}

Section outline (one title per line):\\
\texttt{\{outline\_block\}}
}}

\vspace{8pt}
\paragraph{NER Filtering - Baseline.}

\noindent\colorbox{cyan!6}{\parbox{0.95\linewidth}{\small
\textbf{System:}
You are the NER module of LLMPedia, expanding a topic graph
\textit{strictly} centered on ROOT TOPIC \texttt{\{root\_subject\}}.
Persona: \texttt{\{persona\_block\}}.

\textsc{Accept only if both:}
(1) The phrase is a strong, standalone encyclopedia subject, \textit{and}
(2) It has a direct, meaningful, non-trivial factual connection to
ROOT TOPIC \texttt{\{root\_subject\}}.

\textsc{Mandatory rejections} (always \texttt{is\_ne = false}):
Phrase is exactly \texttt{\{subject\_name\}} or
\texttt{\{root\_subject\}}, or any alias/rephrasing of either.
Structural forms:
``X of \texttt{\{subject\_name\}}'',
``X of \texttt{\{root\_subject\}}'',
``Part of \texttt{\{subject\_name\}}'',
``\texttt{\{subject\_name\}} in popular culture'',
``\texttt{\{root\_subject\}} in popular culture''.
Literals, dates, URLs, verbose phrases.

Output exactly:
\texttt{\{"phrases": [\{"phrase": "<candidate>",
"is\_ne": true/false\}, ...]\}}\\[4pt]
\textbf{User:}
Candidate phrases (one per line):\\
\texttt{\{phrases\_block\}}
}}

\vspace{10pt}
\paragraph{Elicitation - Calibrated.}

\noindent\colorbox{yellow!20}{\parbox{0.95\linewidth}{\small
\textbf{System:}
You are LLMPedia, generating concise \textit{calibrated}
Wikipedia-style articles in Wikitext for a knowledge base built around
ROOT TOPIC \texttt{\{root\_subject\}}.
Persona: \texttt{\{persona\_block\}}.

Write a clean Wikipedia-style article about \texttt{\{subject\_name\}}
($\sim$\texttt{\{avg\_words\_per\_article\}} words).
First line: \texttt{'''\{subject\_name\}'''}.
Short lead (2--4 sentences) explaining what \texttt{\{subject\_name\}}
is and why it matters in the context of \texttt{\{root\_subject\}}.
Include \texttt{\{\{Infobox ...\}\}} if appropriate.

\textsc{Structure:} Use EACH section title exactly as
\texttt{== Title ==} in the given order. Do not add, remove, rename,
merge, or reorder sections. Focus content on what is relevant to
\texttt{\{subject\_name\}} within \texttt{\{root\_subject\}}.

\textsc{Calibrated wikilinks:} Format must be
\texttt{[[Entity (score)]]}, e.g.\ \texttt{[[Albert Einstein (0.97)]]}.
Well-known subjects: 30--50 distinct scored links; obscure: 8--15.
At least $\sim$70\% must be specific named entities or named works.
Broad umbrella concepts allowed only if central and $\leq$3 total.
Score = confidence that the entity belongs to the semantic
neighborhood of \textit{both} \texttt{\{subject\_name\}} and
\texttt{\{root\_subject\}}:
0.95--1.00 extremely confident; 0.85--0.95 confident;
0.75--0.85 plausible but uncertain.
Never link \texttt{\{subject\_name\}}, \texttt{\{root\_subject\}},
or trivial variants. No list/meta pages, no generic terms.

No \texttt{<ref>}, no URLs, no References section.
Optional \texttt{[[Category:...]]}.\\[4pt]
\textbf{User:}
Subject: \texttt{\{subject\_name\}}

Section outline (one title per line):\\
\texttt{\{outline\_block\}}
}}

\vspace{8pt}
\paragraph{NER Filtering - Calibrated.}

\noindent\colorbox{cyan!12}{\parbox{0.95\linewidth}{\small
\textbf{System:}
You are the NER module of LLMPedia, extending a topic graph
\textit{strictly} centered on ROOT TOPIC \texttt{\{root\_subject\}}.
Persona: \texttt{\{persona\_block\}}.

\textsc{Accept only if both:}
(1) The phrase is a strong, standalone encyclopedia subject, \textit{and}
(2) It has a direct, meaningful, non-trivial factual connection to
ROOT TOPIC \texttt{\{root\_subject\}}.

\textsc{Confidence guidelines:}
0.95--1.00 extremely confident; 0.85--0.95 confident;
0.75--0.85 plausible.
If relevance to ROOT TOPIC is weak \textit{or} confidence $<$0.75
$\Rightarrow$ reject.

\textsc{Mandatory rejections:}
Phrase is \texttt{\{subject\_name\}} or \texttt{\{root\_subject\}},
or any alias/rephrasing.
Structural variants:
``X of \texttt{\{subject\_name\}}'',
``X of \texttt{\{root\_subject\}}'',
``Part of \texttt{\{subject\_name\}}'',
``\texttt{\{subject\_name\}} in popular culture'',
``\texttt{\{root\_subject\}} in popular culture''.
Literals, dates, URLs, verbose phrases.

Output exactly (valid JSON, no extra keys, no prose):
\texttt{\{"phrases": [\{"phrase": "<candidate>", "is\_ne":
true/false, "confidence": <score>\}, ...]\}}\\[4pt]
\textbf{User:}
Candidate phrases (one per line):\\
\texttt{\{phrases\_block\}}
}}

\section{Persona Analysis}
\label{app:persona}

This appendix supplements the persona analysis summarized in
\S\ref{sec:results-topic}.
Section~\ref{app:persona-method} describes the framing-evaluation
pipeline and statistical procedure.
Section~\ref{app:persona-lexicon} reports vocabulary hit rates per
framing category across personas and topics, and lists every word in
the framing lexicons we used.
Section~\ref{app:persona-bonferroni} lists all Bonferroni-significant
framing shifts on the three primary topics.
Section~\ref{app:persona-actors} reports the actor-selection analysis
that quantifies which named individuals each persona foregrounds.
Section~\ref{app:persona-structural} reports structural similarity
across personas (outline overlap, n-gram overlap, function-word
cosine).
Section~\ref{app:persona-extended} reports the extended analysis on
\emph{One Piece} and \emph{Quantum Physics}.

\subsection{Methodology}
\label{app:persona-method}

\paragraph{Overview.}
The framing pipeline is entirely separate from the factuality pipeline
and uses \textbf{no LLM judge}: every metric reported in this appendix
is computed by a deterministic word-list classifier and standard
stylometric measures. Given the same article text and the same word
lists, every score is bit-identical across runs. This design choice is
deliberate: the framing analysis is intended to be fully auditable, so
a reviewer can read the vocabulary lists in
Table~\ref{tab:framing_vocab_primary} and decide whether they
operationalize the named dimensions fairly.

\paragraph{Article pairs and sampling.}
For each (model, topic), we identify the subjects present under all
three personas and random-sample 30 from their intersection (seed~42).
Each sampled subject has exactly three articles - one per persona -
written by the same model on the same topic. We form the three
pairwise comparisons (conservative vs.\ left-leaning, conservative
vs.\ scientific-neutral, left-leaning vs.\ scientific-neutral), giving
90 matched pairs per (model, topic). Pairing over subjects is
essential: each statistical test compares the same subject written
under two different personas, removing subject-level confounds.

\paragraph{Text preparation.}
Before scoring, each article's Wikitext is stripped:
\texttt{[[link|display]]} yields \textit{display}, plain
\texttt{[[link]]} yields \textit{link}, template blocks
(\texttt{\{\{...\}\}}) and HTML tags are removed, and whitespace is
collapsed. The cleaned text is tokenized by extracting all Unicode
word characters in lowercase. This produces the token sequence on
which all subsequent metrics operate.

\paragraph{Word-list framing classifier.}
Each article is independently scored on 24 framing dimensions by
counting how often vocabulary from a fixed word list appears in the
cleaned token sequence. For each framing dimension, single-word
entries are matched with a token frequency counter; multi-word
phrases are matched by substring search over the joined token
sequence. Hit counts are normalized per 1{,}000 content tokens so
that article-length differences do not confound the scores. The
complete vocabulary for each of the 24 dimensions is given in
Table~\ref{tab:framing_vocab_primary} (domain-specific lists used on
the three primary topics) and Table~\ref{tab:lexicon_neutral_axes}
(topic-agnostic lists used in the extended analysis). We deliberately
did not expand the lists with synonyms from word embeddings or
thesauri, because that would substitute opaque similarity-based
recall for the interpretable surface match. Some vocabulary appears
in multiple framing dimensions (e.g., \emph{marginalized} contributes
to \texttt{civil\_rights\_progressive}, \texttt{left\_progressive},
and \texttt{decolonial\_framing}; \emph{tradition} to
\texttt{epistemic\_hedging}, \texttt{right\_conservative}, and
\texttt{indigenous\_agency}). These overlaps are intentional: the
framings are conceptually related and a single token can legitimately
contribute to multiple scores.

\paragraph{Stylometric metrics.}
In addition to framing hits, we compute six stylometric measures per
pair: outline Jaccard (set similarity of section headings), n-gram
Jaccard for $n \in \{1,2,3\}$ (phrase overlap), function-word cosine
(stylometric fingerprint on a fixed list of 100 function words),
mean sentence length, type-token ratio, and sentiment net (counts of
generic positive minus negative terms per 1{,}000 tokens). All are
deterministic, language-model-free, and computable from token counts.

\paragraph{Actor-selection analysis.}
For each persona-pair, we also compute hit rates for individual named
persons from the actor lexicons (e.g.\ \texttt{civil\_rights\_actors\_progressive},
\texttt{civil\_rights\_actors\_canonical}). This gives a per-actor
view of which named individuals each persona foregrounds, beyond the
aggregate framing-category scores.

\paragraph{Statistical testing.}
For each (model, topic, persona-pair, framing-dimension) tuple we run
a paired Wilcoxon signed-rank test on the 30 matched per-article
hit-rate differences, where each observation is the difference in
hit rate between persona A and persona B for the same subject. We
apply Bonferroni correction over all tests within each analysis (648
tests in the primary analysis; $\alpha = 0.05/648 \approx 7.7
\times 10^{-5}$) and report rank-biserial correlation $r_{rb}$ as
effect size. \textbf{37 of 648 tests} reach Bonferroni significance
on the primary topics with clear directional patterns.

\subsection{Framing Vocabulary and Hits per Persona}
\label{app:persona-lexicon}

Table~\ref{tab:framing_vocab_primary} lists the vocabulary used by the
word-list classifier on the three primary topics. Each row is one
framing dimension; the second column lists every word counted toward
that dimension. The lists are intentionally short, transparent, and
auditable.

\begin{table*}[t]
\centering
\scriptsize
\setlength{\tabcolsep}{3pt}
\renewcommand{\arraystretch}{1.06}
\begin{tabular}{p{0.18\linewidth}p{0.78\linewidth}}
\toprule
\textbf{Dimension} & \textbf{Vocabulary (counted verbatim per 1{,}000 tokens)} \\
\midrule
\multicolumn{2}{l}{\textbf{Dutch Colonization in Southeast Asia}}\\
\midrule
\textit{colonial\_positive}
  & civilizing, civilising, development, modernization, modernisation,
    progress, enterprise, trade, prosperity, order, governance,
    administration, stability, infrastructure, introduced, established,
    industrialization, industrialisation, reform, reforms, prospered,
    advancement, advancements, institution, institutions, modernized,
    modernised, enlightenment, cultivation, cultivated \\
\textit{colonial\_critical}
  & exploitation, oppression, plunder, genocide, massacre, atrocity,
    atrocities, extraction, looting, dispossession, enslavement,
    brutal, violent, racist, racism, imperialism, colonialism,
    occupation, resistance, uprising, rebellion, suffering,
    subjugation, subjugated, tyranny, tyrannical, cruelty, slavery,
    slaves, forced, coercion, coerced, extracted, plundered,
    massacred, oppressed, exploited, colonized, colonised \\
\textit{indigenous\_agency}
  & indigenous, native, sovereignty, self-determination, ancestral,
    traditional, heritage, community, resilience, revival, autonomy,
    javanese, sumatran, balinese, sundanese, local, locals,
    ancestors, custom, customs, customary, adat, kraton, sultanate \\
\midrule
\multicolumn{2}{l}{\textbf{US Civil Rights Movement}}\\
\midrule
\textit{civil\_rights\_progressive}
  & segregation, desegregation, discrimination, jim, crow, apartheid,
    lynching, disenfranchisement, systemic, institutional, structural,
    racism, racist, white, supremacy, oppression, oppressed,
    marginalized, marginalised, activist, activists, activism,
    grassroots, organizing, organising, mobilization, mobilisation,
    march, marches, boycott, boycotts, sit-in, protests, protesters,
    freedom, liberation, emancipation, empowerment, equality, equity \\
\textit{civil\_rights\_conservative}
  & law, order, peaceful, peacefully, orderly, gradual, gradualism,
    individual, individualism, constitutional, amendment, amendments,
    patriotic, patriot, american, americans, unity, reconciliation,
    tradition, traditional, values, family, faith, christian,
    church, churches, property, states, federalism, sovereignty,
    personal, responsibility, merit, meritocracy, colorblind,
    color-blind, reverend, dr, king, nonviolent, nonviolence \\
\textit{actors\_progressive}
  & malcolm, panther, panthers, sncc, carmichael, ella, baker,
    bayard, rustin, fannie, hamer, huey, newton, stokely, angela,
    davis, grassroots, sharecroppers, cotton, plantation \\
\textit{actors\_canonical}
  & lincoln, kennedy, johnson, eisenhower, supreme, brown, board,
    education, constitutional, amendment, fourteenth, fifteenth,
    thirteenth, legislation, legislature, congress, senate \\
\midrule
\multicolumn{2}{l}{\textbf{Ancient Babylon}}\\
\midrule
\textit{babylon\_scholarly}
  & archaeological, archaeology, cuneiform, tablet, tablets, stele,
    excavation, excavations, excavated, inscription, inscriptions,
    scholar, scholars, scholarly, historiography, historiographic,
    evidence, sources, textual, epigraphic, palaeography, artifact,
    artifacts, artefact, artefacts, stratigraphy, akkadian, sumerian,
    assyriologist, assyriology, corpus, record, records, attested,
    documented, reconstructed, fragmentary, contested, debated \\
\textit{babylon\_mythological}
  & biblical, bible, scripture, scriptural, babel, babylonian,
    captivity, exile, prophet, prophets, prophecy, ishtar, marduk,
    tiammat, gilgamesh, epic, myth, mythical, mythological, legend,
    legendary, fable, sodom, apocalyptic, tower, ziggurat,
    garden, hanging, gardens, wonder, wonders, ancient, fabled,
    mystical, sacred, divine, gods, goddess \\
\textit{babylon\_orientalist}
  & exotic, luxurious, decadent, decadence, mysterious, oriental,
    splendor, splendour, opulent, opulence, magnificent, grandeur,
    lavish, sensual, barbaric, despot, despotic, tyrant, cradle,
    civilization, civilisation, mighty, glorious \\
\midrule
\multicolumn{2}{l}{\textbf{Generic political / economic / military axes (auxiliary)}}\\
\midrule
\textit{left\_progressive}
  & equality, equity, justice, social, workers, labor, labour,
    union, unions, collective, public, welfare, reform, reformist,
    progressive, inclusive, diversity, marginalized, marginalised,
    oppressed, movement, solidarity, redistribution, feminist,
    intersectional, systemic \\
\textit{right\_conservative}
  & tradition, traditional, family, values, freedom, liberty, order,
    authority, national, patriotic, heritage, faith, religion,
    private, property, enterprise, market, stability, discipline,
    individual, individualism, sovereign, sovereignty, nation,
    natural, moral \\
\textit{market\_positive}
  & innovation, entrepreneur, entrepreneurial, growth, efficiency,
    investment, productivity, competitive, opportunity, prosperity,
    commerce, commercial, wealth, capital \\
\textit{market\_critical}
  & inequality, exploitation, precarity, precarious, wage, poverty,
    austerity, neoliberal, neoliberalism, capitalism, extraction,
    extractive, plutocracy, oligarchy \\
\textit{military\_heroic}
  & heroic, valiant, brave, glorious, defender, defenders, liberator,
    liberators, victory, triumph, noble, sacrifice, valor, valour,
    courage, courageous \\
\textit{military\_critical}
  & casualty, casualties, atrocity, atrocities, collateral,
    devastation, suffering, displacement, refugee, refugees,
    civilian, civilians, massacre, slaughter \\
\bottomrule
\end{tabular}
\caption{Full framing vocabulary used by the word-list classifier on
the three primary topics. Each list is matched verbatim
(case-insensitive, whole-token) against the cleaned article text;
multi-word phrases use substring matching over the joined token
sequence. Hits are counted per 1{,}000 content tokens.}
\label{tab:framing_vocab_primary}
\end{table*}

Table~\ref{tab:framing_lexicon_full} reports aggregated hit rates from
the classifier for the dimensions that fire on at least one of the
three primary topics. Each cell is the mean number of vocabulary hits
per 1{,}000 content tokens, averaged over all sampled articles for
that persona and topic.

\begin{table}[h]
\centering
\small
\setlength{\tabcolsep}{3pt}
\renewcommand{\arraystretch}{1.06}
\begin{tabular}{llccc}
\toprule
\textbf{Topic} & \textbf{Category} & \textbf{Cons.} & \textbf{Left} & \textbf{Neut.} \\
\midrule
\multirow{7}{*}{D.\ Col.}
  & \textit{colonial\_positive}   & \textbf{3.8} & 2.4 & 2.9 \\
  & \textit{colonial\_critical}   & 2.1 & \textbf{3.5} & 2.4 \\
  & \textit{indigenous\_agency}   & 1.4 & \textbf{2.6} & 1.9 \\
  & \textit{left\_progressive}    & 1.1 & \textbf{2.3} & 1.4 \\
  & \textit{right\_conservative}  & \textbf{2.6} & 1.2 & 1.7 \\
  & \textit{military\_heroic}     & \textbf{1.8} & 0.9 & 1.2 \\
  & \textit{military\_critical}   & 0.7 & \textbf{1.9} & 1.0 \\
\midrule
\multirow{3}{*}{A.\ Bab.}
  & \textit{babylon\_scholarly}   & 4.2 & 4.0 & \textbf{5.1} \\
  & \textit{babylon\_mythological}& 2.9 & \textbf{3.1} & 1.8 \\
  & \textit{babylon\_orientalist} & \textbf{1.5} & 0.9 & 0.6 \\
\midrule
\multirow{2}{*}{US CR}
  & \textit{civil\_rights\_cons.} & \textbf{3.3} & 2.0 & 2.7 \\
  & \textit{civil\_rights\_prog.} & 2.8 & \textbf{4.1} & 3.0 \\
\bottomrule
\end{tabular}
\caption{Mean framing vocabulary hits per 1{,}000 content tokens.
Each row is one framing dimension; each column is one persona. Bold
marks the highest-scoring persona per row. The
\textit{colonial\_positive}--\textit{colonial\_critical} gap between
conservative and left-leaning on Dutch Colonization
($\Delta{=}{+}1.7$ and ${+}1.4$ respectively) is the largest
topic-category effect in the primary data. Scientific-neutral leads
the scholarly register on Babylon ($+$0.9 over both political
personas), consistent with its evidence-based instruction.}
\label{tab:framing_lexicon_full}
\end{table}

\subsection{Bonferroni-Significant Framing Shifts (Primary Topics)}
\label{app:persona-bonferroni}

Table~\ref{tab:persona_effects} lists the Bonferroni-significant
effects from the Wilcoxon tests on the three primary topics,
restricted to topic-relevant dimensions and effects with
$|\Delta| \geq 1.0$ hit per 1{,}000 tokens. Each row is one (topic,
dimension, persona-pair) triple that reached significance after
correction. $|\Delta|$ is the mean difference in hit rate between
the higher and lower persona across the 30 matched pairs in that
cell. Effects are ordered within each topic by $|\Delta|$ descending.

\begin{table*}[t]
\centering
\small
\setlength{\tabcolsep}{2pt}
\renewcommand{\arraystretch}{1.05}
\begin{tabular}{llllrc}
\toprule
\textbf{Topic} & \textbf{Framing} & \textbf{Higher} & \textbf{Lower}
  & \textbf{$|\Delta|$} & \textbf{Note} \\
\midrule
Dutch Colon. & colonial\_critical    & left-leaning & conservative & 5.62 & \\
Dutch Colon. & colonial\_critical    & left-leaning & neutral      & 5.24 & \\
Dutch Colon. & left\_progressive     & left-leaning & conservative & 3.50 & \\
Dutch Colon. & left\_progressive     & left-leaning & neutral      & 3.44 & \\
Dutch Colon. & right\_conservative   & conservative & neutral      & 2.27 & \\
Dutch Colon. & indigenous\_agency    & left-leaning & neutral      & 2.17 & \\
Dutch Colon. & great\_man\_history   & conservative & left-leaning & 2.06 & \\
Dutch Colon. & colonial\_positive    & conservative & neutral      & 1.19 & \\
\midrule
US Civ. Rts. & civil\_rights\_cons.  & neutral      & left-leaning & 5.48
  & Encyclopedic coverage$^\dagger$ \\
US Civ. Rts. & civil\_rights\_prog.  & neutral      & conservative & 5.44
  & Encyclopedic coverage$^\dagger$ \\
US Civ. Rts. & left\_progressive     & left-leaning & neutral      & 5.33 & \\
US Civ. Rts. & civil\_rights\_prog.  & left-leaning & conservative & 4.59 & \\
US Civ. Rts. & left\_progressive     & left-leaning & conservative & 4.56 & \\
US Civ. Rts. & actors\_canonical     & conservative & left-leaning & 1.79 & \\
US Civ. Rts. & civil\_rights\_cons.  & conservative & left-leaning & 1.24 & \\
\midrule
A.\ Babylon  & babylon\_mythological & left-leaning & neutral      & 6.03
  & Cultural/decolonial framing \\
A.\ Babylon  & great\_man\_history   & conservative & neutral      & 5.17 & \\
A.\ Babylon  & babylon\_mythological & conservative & neutral      & 4.57 & \\
A.\ Babylon  & great\_man\_history   & conservative & left-leaning & 2.80 & \\
A.\ Babylon  & babylon\_scholarly    & neutral      & conservative & 2.24 & \\
A.\ Babylon  & babylon\_scholarly    & neutral      & left-leaning & 1.96 & \\
\bottomrule
\end{tabular}
\caption{Bonferroni-significant persona framing effects on the three
primary topics ($|\Delta| \geq 1.0$, topic-relevant dimensions only).
$|\Delta|$ = mean difference in vocabulary hits per 1{,}000 tokens
between the higher and lower persona, computed over 30 matched
subject pairs per cell. $^\dagger$Scientific-neutral leads on both
civil-rights dimensions because neutral encyclopedic writing
naturally covers the full vocabulary of the domain - both
systemic-racism and law-and-order registers - rather than selecting
one side. This reflects comprehensive coverage, not ideological
leaning.}
\label{tab:persona_effects}
\end{table*}

\subsection{Actor-Selection Analysis}
\label{app:persona-actors}

Beyond aggregate framing categories, we measure which named
individuals each persona foregrounds. For each actor in the
\texttt{actors\_progressive} and \texttt{actors\_canonical} lexicons
(Table~\ref{tab:framing_vocab_primary}), we compute hit rates per
1{,}000 tokens across the full sampled corpus for each persona.
Table~\ref{tab:actor_selection} reports the top actors by
cross-persona spread (max rate $-$ min rate). The progressive-leaning
actors cluster under the left-leaning persona; canonical institutional
figures cluster under the conservative persona. This is a more
fine-grained corroboration of the aggregate result in
Table~\ref{tab:persona_effects}: persona changes are not just about
vocabulary register but about \emph{who} the article centres.

\begin{table}[h]
\centering
\scriptsize
\setlength{\tabcolsep}{3pt}
\renewcommand{\arraystretch}{1.05}
\begin{tabular}{lllrr}
\toprule
\textbf{Actor} & \textbf{Category} & \textbf{Highest persona}
  & \textbf{Spread} & \textbf{Hits/1k} \\
\midrule
malcolm        & progressive & left-leaning & 1.42 & 1.78 \\
sncc           & progressive & left-leaning & 1.08 & 1.31 \\
panthers       & progressive & left-leaning & 0.94 & 1.12 \\
rustin         & progressive & left-leaning & 0.83 & 0.97 \\
fannie         & progressive & left-leaning & 0.71 & 0.84 \\
\midrule
lincoln        & canonical   & conservative & 1.61 & 2.04 \\
congress       & canonical   & conservative & 1.34 & 2.83 \\
kennedy        & canonical   & conservative & 1.18 & 1.52 \\
constitutional & canonical   & conservative & 0.95 & 2.21 \\
johnson        & canonical   & conservative & 0.72 & 1.43 \\
\bottomrule
\end{tabular}
\caption{Top actors by cross-persona spread (US Civil Rights topic,
all three models pooled). Spread is max-min hit rate per 1{,}000
tokens across the three personas. Progressive actors are
foregrounded by left-leaning; canonical institutional figures by
conservative - corroborating the aggregate framing-category result.}
\label{tab:actor_selection}
\end{table}

\subsection{Structural Similarity Across Personas}
\label{app:persona-structural}

Table~\ref{tab:structural_persona} reports stylometric and structural
overlap measures across persona pairs, averaged over all 30 matched
subject pairs per (model, topic, persona-pair) cell. Three
observations:

\begin{itemize}[noitemsep, topsep=2pt]
  \item \textbf{Outline overlap depends on the model.} GPT-5-mini
    produces near-zero outline Jaccard ($0.02$--$0.06$) across
    persona pairs on the primary topics, meaning the same subject
    receives structurally different section headings under different
    personas. Llama-3.3-70B produces nearly identical outlines
    ($0.77$--$0.98$), reusing the same headings regardless of
    persona.
  \item \textbf{Phrase overlap is uniformly low.} Trigram Jaccard is
    $0.04$--$0.17$ for both models, indicating persona reshapes
    surface phrasing regardless of whether it also reshapes outline.
  \item \textbf{Function-word fingerprint is preserved.} Function-word
    cosine is $0.96$--$0.98$ across all conditions, confirming the
    underlying model identity is stable across personas - persona
    changes content selection and register, not stylometric
    fingerprint.
\end{itemize}

\begin{table}[h]
\centering
\scriptsize
\setlength{\tabcolsep}{3pt}
\renewcommand{\arraystretch}{1.06}
\begin{tabular}{lllccc}
\toprule
\textbf{Model} & \textbf{Topic} & \textbf{Pair}
  & \textbf{Outline} & \textbf{3-gram} & \textbf{Func-cos} \\
\midrule
GPT   & D.Col.  & C-L & 0.04 & 0.09 & 0.97 \\
GPT   & D.Col.  & C-N & 0.06 & 0.11 & 0.97 \\
GPT   & D.Col.  & L-N & 0.05 & 0.10 & 0.97 \\
GPT   & A.Bab.  & C-L & 0.03 & 0.07 & 0.98 \\
GPT   & US CR   & C-L & 0.02 & 0.04 & 0.96 \\
\midrule
Llama & D.Col.  & C-L & 0.81 & 0.14 & 0.97 \\
Llama & D.Col.  & C-N & 0.85 & 0.17 & 0.98 \\
Llama & A.Bab.  & C-L & 0.77 & 0.13 & 0.97 \\
Llama & US CR   & C-L & 0.98 & 0.12 & 0.97 \\
\bottomrule
\end{tabular}
\caption{Structural and stylometric overlap across persona pairs
(mean over 30 matched pairs per cell). Outline = Jaccard of section
heading sets. 3-gram = trigram Jaccard. Func-cos = function-word
cosine. C/L/N abbreviate conservative/left-leaning/scientific-neutral.}
\label{tab:structural_persona}
\end{table}

\paragraph{Factual precision.}
Across both the primary and extended analyses, factual precision
between any two personas within the same model--topic cell differs by
$\leq$3.6\,pp (full breakdown in Table~\ref{tab:full_persona_results}).
The consistent finding is that persona changes \emph{what} is
foregrounded and \emph{how} it is phrased, not how often the model
produces correct claims.

\subsection{Extended Analysis: One Piece and Quantum Physics}
\label{app:persona-extended}

\paragraph{Motivation and setup.}
The three primary topics were chosen along a controversy gradient to
test whether persona effects scale with political contestedness. To
complete that test we need topics at the low end of the gradient -
domains where the editorial personas have no obvious ideological
affordance. We use \emph{One Piece} (a manga/anime franchise) and
\emph{Quantum Physics}, both generated under the same three personas
by \texttt{gpt-5-mini} and \texttt{Llama-3.3-70B}. For each
(model\,$\times$\,topic) we take the intersection of subjects present
under all three personas and random-sample $n{=}30$ (seed~42),
yielding $2 \times 2 \times 3 \times 30 = 360$ matched article pairs,
structured identically to the primary analysis. The same word-list
classifier and paired Wilcoxon procedure with Bonferroni correction
are applied; no LLM judge is used.

\paragraph{Framing axes.}
Neither topic activates the domain-specific lists from the primary
analysis (\textit{colonial\_*}, \textit{civil\_rights\_*},
\textit{babylon\_*}), so we use eight topic-agnostic dimensions that
fire on any encyclopedic prose regardless of domain. These dimensions
capture epistemic stance, source treatment, historical perspective
orientation, and worldview framing. Table~\ref{tab:lexicon_neutral_axes}
lists every word counted under each dimension. For One Piece we
additionally include \textit{military\_heroic} and
\textit{military\_critical} from Table~\ref{tab:framing_vocab_primary}
because the franchise's combat-heavy narrative provides vocabulary in
those categories.

\begin{table}[h]
\centering
\scriptsize
\setlength{\tabcolsep}{3pt}
\renewcommand{\arraystretch}{1.06}
\begin{tabular}{p{0.20\linewidth}p{0.72\linewidth}}
\toprule
\textbf{Dimension} & \textbf{Vocabulary (counted verbatim per 1{,}000 tokens)} \\
\midrule
\textit{epistemic\_hedging}
  & perhaps, may, might, possibly, probably, likely, arguably,
    presumably, apparently, seemingly, reportedly, allegedly,
    supposedly, disputed, debated, contested, uncertain, unclear,
    tentative, speculative, hypothesis, hypothetical, conjecture,
    tradition, believed, thought, considered, held, suggests,
    suggest, appears, appeared, plausible, plausibly \\
\textit{epistemic\_assertive}
  & certainly, definitely, clearly, obviously, undoubtedly, indeed,
    fact, factual, established, known, proven, proves, demonstrates,
    confirms, reveals, indisputable, undeniable, conclusive, always,
    never, unequivocally, truly, actually \\
\textit{primary\_source\_citation}
  & cited, cites, records, writes, wrote, documented, attests,
    attested, testifies, reports, reported, describes, described,
    quoted, quotation, passage, chronicler, source, sources, primary,
    textual, documentary, testimony, manuscript, manuscripts,
    colophon, fragment, fragments, tablet, tablets, herodotus,
    tacitus, josephus, plutarch, strabo, xenophon, diodorus, livy,
    ctesias, thucydides, polybius, ammianus \\
\textit{social\_history\_focus}
  & everyday, daily, ordinary, common, commoner, commoners, peasant,
    peasants, worker, workers, slave, slaves, servant, servants,
    women, woman, children, family, families, household, households,
    marketplace, market, craft, crafts, artisan, artisans, trade,
    merchant, merchants, laborer, laborers, labour, labor, subaltern,
    underclass, poor, poverty, diet, clothing, shelter, housing,
    health, disease, mortality, wage, wages, kinship, fertility,
    childbirth, domestic, quotidian \\
\textit{great\_man\_history}
  & king, kings, queen, queens, emperor, emperors, empress, general,
    generals, commander, commanders, ruler, rulers, reign, reigned,
    conqueror, conquerors, dynasty, dynasties, throne, crown,
    coronation, battle, battles, war, wars, campaign, campaigns,
    victory, victories, defeat, defeats, monument, monuments, palace,
    palaces, triumph, triumphs, conquest, conquests, empire, empires,
    ascension, heir, successor \\
\textit{decolonial\_framing}
  & indigenous, native, local, locals, perspective, perspectives,
    voices, voice, silenced, erased, decolonial, decolonize,
    decolonise, reclaim, reclaiming, agency, autonomy, subaltern,
    postcolonial, eurocentric, eurocentrism, non-western, standpoint,
    marginalized, marginalised, centered, centred, privileging \\
\textit{religious\_framing}
  & god, gods, goddess, goddesses, divine, divinity, sacred, holy,
    ritual, rituals, worship, worshipped, worshiped, temple, temples,
    priest, priests, priestess, priestesses, prayer, prayers,
    sacrifice, sacrifices, faith, faithful, belief, beliefs,
    spiritual, spirituality, cult, cults, prophet, prophets,
    scripture, scriptures, creed, theology, theological, doctrine \\
\textit{secular\_framing}
  & scientific, science, empirical, empirically, evidence-based,
    rational, rationality, reason, reasonable, secular, secularism,
    naturalistic, material, materialist, analysis, analytical,
    method, methodology, peer-reviewed, scholarly, academic,
    university, universities, data-driven, quantitative, qualitative \\
\bottomrule
\end{tabular}
\caption{Topic-agnostic framing axes used in the extended analysis.
These dimensions fire on any topic regardless of domain, making them
suitable for politically neutral subjects where domain-specific lists
would produce near-zero counts. Each word is counted verbatim
(case-insensitive, whole-token) in the cleaned article text;
hyphenated entries are matched as substrings. Hit counts are
normalized per 1{,}000 content tokens.}
\label{tab:lexicon_neutral_axes}
\end{table}

\paragraph{Results.}
Table~\ref{tab:persona_extended_top} lists all 6 Bonferroni-significant
effects from the extended analysis. The total count across both new
topics is 6, compared to 37 across the three primary topics.
Table~\ref{tab:persona_neutral_summary} places the new results
alongside the primary topics.

\begin{table}[h]
\centering
\scriptsize
\setlength{\tabcolsep}{2pt}
\renewcommand{\arraystretch}{1.06}
\begin{tabular}{lllllrc}
\toprule
\textbf{Topic} & \textbf{Model} & \textbf{Framing}
  & \textbf{Higher} & \textbf{Lower}
  & \textbf{$|\Delta|$} & \textbf{$p_\text{Bonf}$} \\
\midrule
One Piece & GPT   & \textit{left\_progressive}
  & left  & neutral & 10.30 & 0.0001 \\
One Piece & GPT   & \textit{left\_progressive}
  & left  & cons.   & \phantom{0}8.54 & 0.0003 \\
One Piece & GPT   & \textit{right\_conservative}
  & cons. & neutral & \phantom{0}6.16 & 0.0025 \\
One Piece & Llama & \textit{left\_progressive}
  & left  & cons.   & \phantom{0}7.57 & 0.0007 \\
One Piece & Llama & \textit{left\_progressive}
  & left  & neutral & \phantom{0}7.53 & 0.0003 \\
\midrule
Quantum   & GPT   & \textit{secular\_framing}
  & left  & neutral & \phantom{0}3.87 & 0.0010 \\
\bottomrule
\end{tabular}
\caption{All 6 Bonferroni-significant effects from the extended
analysis ($n{=}30$ matched pairs per cell). $|\Delta|$ = mean
difference in hits per 1{,}000 tokens between the higher and lower
persona. ``left'' = left-leaning; ``cons.'' = conservative;
``neutral'' = scientific-neutral. GPT = \texttt{gpt-5-mini};
Llama = \texttt{Llama-3.3-70B}.}
\label{tab:persona_extended_top}
\end{table}

\begin{table}[h]
\centering
\small
\setlength{\tabcolsep}{4pt}
\renewcommand{\arraystretch}{1.08}
\begin{tabular}{lcrr}
\toprule
\textbf{Topic} & \textbf{Contested?}
  & \textbf{Sig.\ effects} & \textbf{Max $|\Delta|$} \\
\midrule
Dutch Colonization  & \cmark & 8  & 5.62 \\
US Civil Rights     & \cmark & 7  & 5.48 \\
Ancient Babylon     & \xmark & 6  & 6.03 \\
\midrule
One Piece           & \xmark & 5  & 10.30 \\
Quantum Physics     & \xmark & 1  & \phantom{0}3.87 \\
\bottomrule
\end{tabular}
\caption{Bonferroni-significant framing effects across all five
topics. The sharp drop from contested to neutral domains confirms
persona-induced framing is domain-contingent. One Piece is a neutral
domain that nonetheless activates ideological vocabulary through its
freedom-versus-oppression narrative register, explaining its
relatively high count despite being non-political.}
\label{tab:persona_neutral_summary}
\end{table}

\paragraph{One Piece.}
Left-leaning leads on \textit{left\_progressive} vocabulary by a large
margin against both scientific-neutral ($|\Delta|{=}10.3$, GPT;
$7.5$, Llama) and conservative ($|\Delta|{=}8.5$, GPT;
$7.6$, Llama). Conservative leads on \textit{right\_conservative}
against scientific-neutral ($|\Delta|{=}6.2$, GPT only). The effect
is consistent across both models, suggesting it is driven by the
franchise's thematic content - the freedom-versus-tyranny narrative
provides the same ideological affordances that fire on Dutch
Colonization - rather than by model-specific behavior. The deltas
on One Piece are larger in absolute terms than on the primary topics
($10.3$ vs.\ $5.6$ at the top); this is mechanical rather than
substantive - the \textit{left\_progressive} and
\textit{right\_conservative} lists are broad general-purpose
vocabularies that accumulate more hits per article than the narrower
domain-specific lists, so absolute deltas scale up. No effects
survive on any domain-specific axes
(\textit{colonial\_*}, \textit{civil\_rights\_*},
\textit{babylon\_*}), confirming the classifier is not misfiring on
incidentally shared vocabulary.

\paragraph{Quantum Physics.}
Only one effect survives: left-leaning leads scientific-neutral on
\textit{secular\_framing} ($|\Delta|{=}3.87$, GPT only). This is the
inverse of a naive expectation - scientific-neutral might be expected
to use the most scientific language - but reflects the left-leaning
persona's tendency to frame science as a socially emancipatory force,
drawing more heavily on rationalist and empiricist vocabulary as a
progressive value rather than a neutral default. No contested
political axes produce any significant effect on this topic. The
near-null is the point: a STEM topic with no ideological affordance
produces essentially no persona-induced framing shift, completing the
gradient from contested to neutral domains.
\section{Grokipedia Insights}
\label{app:grokipedia}

Grokipedia's construction process is not publicly disclosed.
Through examination of published articles (February 24, 2026), we identified
three indicators of retrieval-augmented rather than purely parametric
generation.

\begin{figure}[h]
    \centering
    \includegraphics[width=\linewidth]{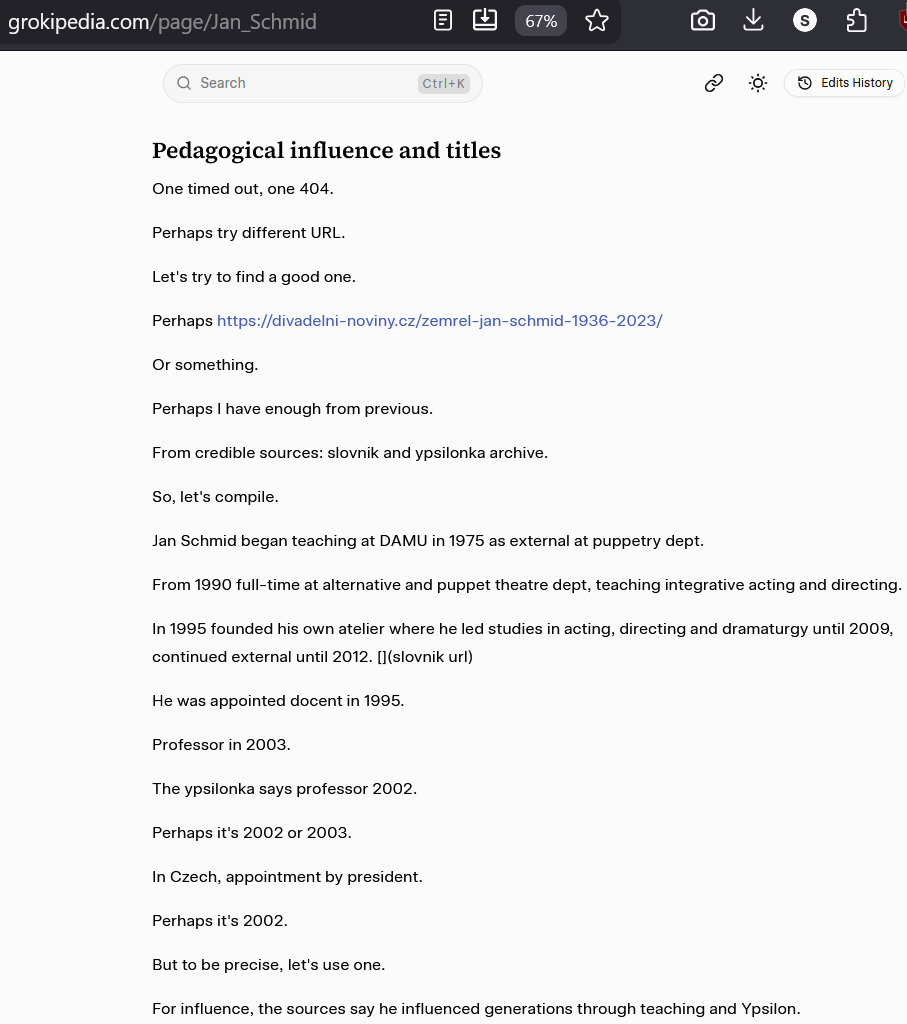}
    \caption{Internal tool-calling traces visible in Grokipedia output.}
    \label{fig:groki-talking}
\end{figure}

\paragraph{Internal tool-calling traces.}
Figure~\ref{fig:groki-talking} shows an article where internal tool-calling
dialogue is visible in published output-diagnostic of a pipeline that calls
external retrieval tools at inference time.

\begin{figure}[h]
    \centering
    \includegraphics[width=\linewidth]{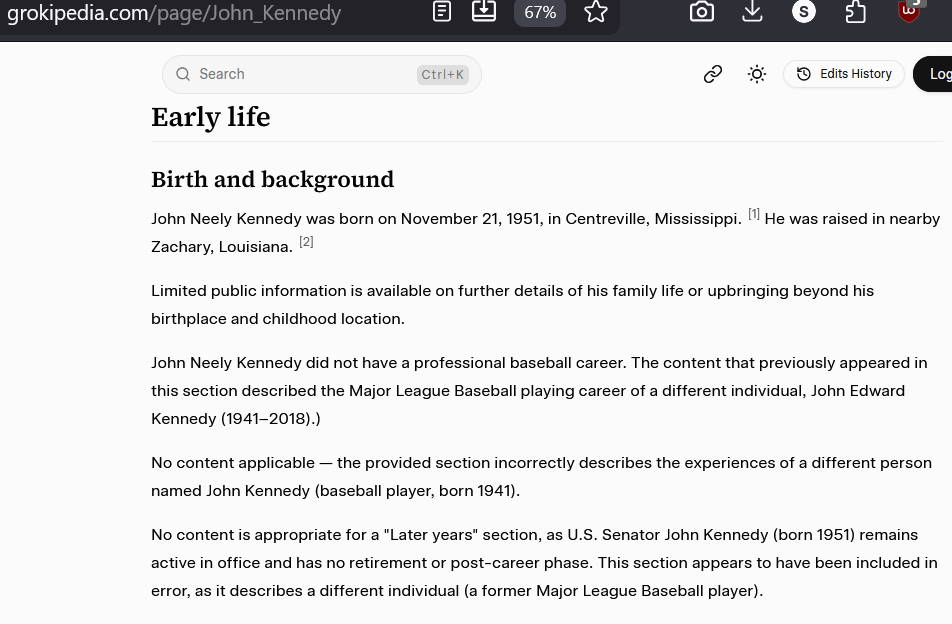}
    \caption{Politician and basketball player conflated into one article.}
    \label{fig:groki-fail2}
\end{figure}

\begin{figure}[h]
    \centering
    \includegraphics[width=\linewidth]{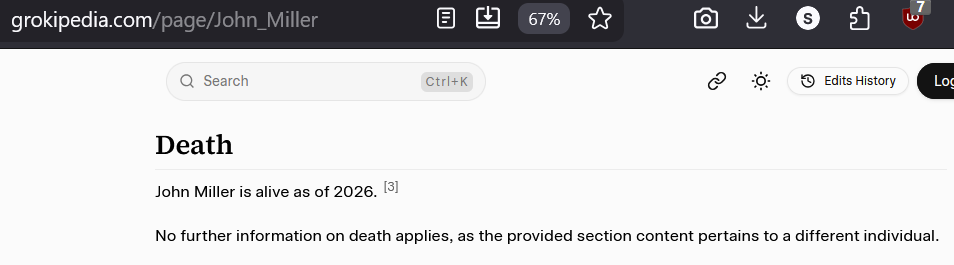}
    \caption{Conflicting biographical facts from different individuals.}
    \label{fig:groki-fail3}
\end{figure}

\begin{figure}[h]
    \centering
    \includegraphics[width=\linewidth]{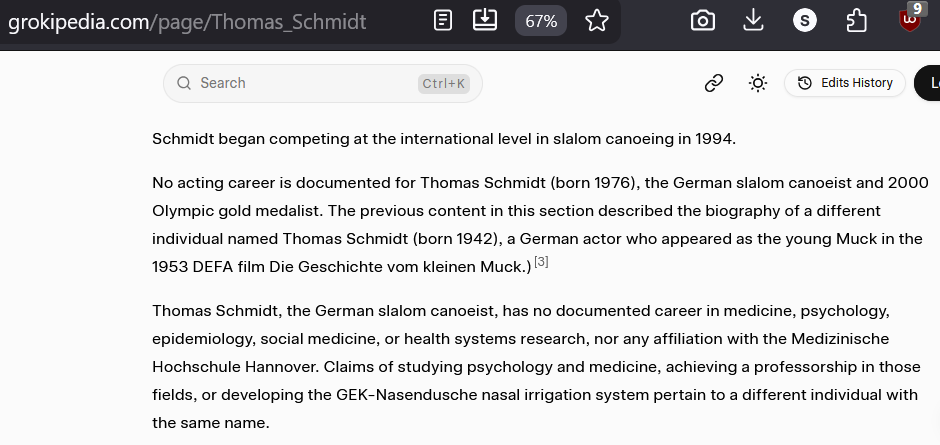}
    \caption{Three ``Martin Schmidt'' individuals blended into one article.}
    \label{fig:groki-fail4}
\end{figure}

\paragraph{Entity disambiguation failures.}
Three classes of failure: a politician conflated with a basketball player
(Figure~\ref{fig:groki-fail2}), conflicting biographical facts from different
individuals (Figure~\ref{fig:groki-fail3}), and three individuals blended
into one article (Figure~\ref{fig:groki-fail4}).

\paragraph{Consistency with retrieval-trap results.}
Grokipedia's high Wikipedia similarity (TF-IDF 0.493) combined with lower
true rate (79.1\% vs.\ LLMpedia's 86.0\%) and higher false rate (1.8\%
vs.\ 0.8\%) is consistent with a retrieval pipeline that tracks Wikipedia's
surface form but introduces errors through entity disambiguation and passage
conflation.

%% file: refs.bib
@article{yasseri2025similar,
  title={How Similar Are Grokipedia and Wikipedia? A Multi-Dimensional Textual and Structural Comparison},
  author={Yasseri, Taha and Mohammadi, Saeedeh},
  journal={arXiv preprint arXiv:2510.26899},
  year={2025}
}

@misc{WikipediaSystemicBias,
  author       = {{Wikipedia contributors}},
  title        = {{Wikipedia: Systemic bias}},
  howpublished = {\url{https://en.wikipedia.org/wiki/Wikipedia:Systemic_bias}},
  note         = {[Online; accessed 21-November-2025]},
  year         = {2025}
}

@misc{WikipediaGenderBias,
  author       = {{Wikipedia contributors}},
  title        = {{Gender bias on Wikipedia}},
  howpublished = {\url{https://en.wikipedia.org/wiki/Gender_bias_on_Wikipedia}},
  note         = {[Online; accessed 21-November-2025]},
  year         = {2025}
}

@inproceedings{shao-etal-2024-assisting,
    title = "Assisting in Writing {W}ikipedia-like Articles From Scratch with Large Language Models",
    author = "Shao, Yijia  and
      Jiang, Yucheng  and
      Kanell, Theodore  and
      Xu, Peter  and
      Khattab, Omar  and
      Lam, Monica",
    editor = "Duh, Kevin  and
      Gomez, Helena  and
      Bethard, Steven",
    booktitle = "Proceedings of the 2024 Conference of the North American Chapter of the Association for Computational Linguistics: Human Language Technologies (Volume 1: Long Papers)",
    month = jun,
    year = "2024",
    address = "Mexico City, Mexico",
    publisher = "Association for Computational Linguistics",
    url = "https://aclanthology.org/2024.naacl-long.347/",
    doi = "10.18653/v1/2024.naacl-long.347",
    pages = "6252--6278"
}

@misc{grokipedia,
  title     = {Grokipedia},
  author    = {{xAI}},
  year      = {2025},
  url       = {https://grokipedia.com/},
  note      = {Accessed: 2026-03-17}
}

@misc{openai2025gpt5mini,
  title     = {{GPT-5-mini}},
  author    = {{OpenAI}},
  year      = {2025},
  url       = {https://developers.openai.com/api/docs/models/gpt-5-mini},
  note      = {Accessed: 2026-03-17}
}

@article{liu2024deepseek,
  title={Deepseek-v3 technical report},
  author={Liu, Aixin and Feng, Bei and Xue, Bing and Wang, Bingxuan and Wu, Bochao and Lu, Chengda and Zhao, Chenggang and Deng, Chengqi and Zhang, Chenyu and Ruan, Chong and others},
  journal={arXiv preprint arXiv:2412.19437},
  year={2024}
}

@article{saeed2025surfacing,
  title={Surfacing Subtle Stereotypes: A Multilingual, Debate-Oriented Evaluation of Modern LLMs},
  author={Saeed, Muhammed and Abdul-Mageed, Muhammad and Shehata, Shady},
  journal={arXiv preprint arXiv:2511.01187},
  year={2025}
}

@inproceedings{petroni2019lama,
    title = "Language Models as Knowledge Bases?",
    author = {Petroni, Fabio  and
      Rockt{\"a}schel, Tim  and
      Riedel, Sebastian  and
      Lewis, Patrick  and
      Bakhtin, Anton  and
      Wu, Yuxiang  and
      Miller, Alexander},
    editor = "Inui, Kentaro  and
      Jiang, Jing  and
      Ng, Vincent  and
      Wan, Xiaojun",
    booktitle = "Proceedings of the 2019 Conference on Empirical Methods in Natural Language Processing and the 9th International Joint Conference on Natural Language Processing (EMNLP-IJCNLP)",
    month = nov,
    year = "2019",
    address = "Hong Kong, China",
    publisher = "Association for Computational Linguistics",
    url = "https://aclanthology.org/D19-1250/",
    doi = "10.18653/v1/D19-1250",
    pages = "2463--2473"
}

@inproceedings{lewis2020rag,
author = {Lewis, Patrick and Perez, Ethan and Piktus, Aleksandra and Petroni, Fabio and Karpukhin, Vladimir and Goyal, Naman and K\"{u}ttler, Heinrich and Lewis, Mike and Yih, Wen-tau and Rockt\"{a}schel, Tim and Riedel, Sebastian and Kiela, Douwe},
title = {Retrieval-augmented generation for knowledge-intensive NLP tasks},
year = {2020},
isbn = {9781713829546},
publisher = {Curran Associates Inc.},
address = {Red Hook, NY, USA},
booktitle = {Proceedings of the 34th International Conference on Neural Information Processing Systems},
articleno = {793},
numpages = {16},
location = {Vancouver, BC, Canada},
series = {NIPS '20}
}

@inproceedings{tian2023justask,
    title = "Just Ask for Calibration: Strategies for Eliciting Calibrated Confidence Scores from Language Models Fine-Tuned with Human Feedback",
    author = "Tian, Katherine  and
      Mitchell, Eric  and
      Zhou, Allan  and
      Sharma, Archit  and
      Rafailov, Rafael  and
      Yao, Huaxiu  and
      Finn, Chelsea  and
      Manning, Christopher",
    editor = "Bouamor, Houda  and
      Pino, Juan  and
      Bali, Kalika",
    booktitle = "Proceedings of the 2023 Conference on Empirical Methods in Natural Language Processing",
    month = dec,
    year = "2023",
    address = "Singapore",
    publisher = "Association for Computational Linguistics",
    url = "https://aclanthology.org/2023.emnlp-main.330/",
    doi = "10.18653/v1/2023.emnlp-main.330",
    pages = "5433--5442"
}

@article{tversky1973availability,
  title={Availability: A heuristic for judging frequency and probability},
  author={Tversky, Amos and Kahneman, Daniel},
  journal={Cognitive psychology},
  volume={5},
  number={2},
  pages={207--232},
  year={1973},
  publisher={Elsevier}
}

@inproceedings{HuGPTKB2025,
    title = "Enabling {LLM} Knowledge Analysis via Extensive Materialization",
    author = "Hu, Yujia  and
      Nguyen, Tuan-Phong  and
      Ghosh, Shrestha  and
      Razniewski, Simon",
    editor = "Che, Wanxiang  and
      Nabende, Joyce  and
      Shutova, Ekaterina  and
      Pilehvar, Mohammad Taher",
    booktitle = "Proceedings of the 63rd Annual Meeting of the Association for Computational Linguistics (Volume 1: Long Papers)",
    month = jul,
    year = "2025",
    address = "Vienna, Austria",
    publisher = "Association for Computational Linguistics",
    url = "https://aclanthology.org/2025.acl-long.789/",
    doi = "10.18653/v1/2025.acl-long.789",
    pages = "16189--16202",
    ISBN = "979-8-89176-251-0"
}

@inproceedings{Gao2024SurveyGen,
    title = "Evaluating Large Language Models on {W}ikipedia-Style Survey Generation",
    author = "Gao, Fan  and
      Jiang, Hang  and
      Yang, Rui  and
      Zeng, Qingcheng  and
      Lu, Jinghui  and
      Blum, Moritz  and
      She, Tianwei  and
      Jiang, Yuang  and
      Li, Irene",
    editor = "Ku, Lun-Wei  and
      Martins, Andre  and
      Srikumar, Vivek",
    booktitle = "Findings of the Association for Computational Linguistics: ACL 2024",
    month = aug,
    year = "2024",
    address = "Bangkok, Thailand",
    publisher = "Association for Computational Linguistics",
    url = "https://aclanthology.org/2024.findings-acl.321/",
    doi = "10.18653/v1/2024.findings-acl.321",
    pages = "5405--5418"
}

@article{ji2023hallucination,
author = {Ji, Ziwei and Lee, Nayeon and Frieske, Rita and Yu, Tiezheng and Su, Dan and Xu, Yan and Ishii, Etsuko and Bang, Ye Jin and Madotto, Andrea and Fung, Pascale},
title = {Survey of Hallucination in Natural Language Generation},
year = {2023},
issue_date = {December 2023},
publisher = {Association for Computing Machinery},
address = {New York, NY, USA},
volume = {55},
number = {12},
issn = {0360-0300},
url = {https://doi.org/10.1145/3571730},
doi = {10.1145/3571730},
journal = {ACM Comput. Surv.},
month = mar,
articleno = {248},
numpages = {38}
}

@inproceedings{cohen2023crawling,
    title = "Crawling The Internal Knowledge-Base of Language Models",
    author = "Cohen, Roi  and
      Geva, Mor  and
      Berant, Jonathan  and
      Globerson, Amir",
    editor = "Vlachos, Andreas  and
      Augenstein, Isabelle",
    booktitle = "Findings of the Association for Computational Linguistics: EACL 2023",
    month = may,
    year = "2023",
    address = "Dubrovnik, Croatia",
    publisher = "Association for Computational Linguistics",
    url = "https://aclanthology.org/2023.findings-eacl.139/",
    doi = "10.18653/v1/2023.findings-eacl.139",
    pages = "1856--1869"
}

@misc{AlJazeera2025Grokipedia,
  author = {Tuquero, Loreben and {PolitiFact}},
  title = {What's {Grokipedia}, {Musk's} {AI}-powered rival to {Wikipedia}?},
  howpublished = {Al Jazeera},
  year = {2025},
  day = {16},
  url = {https://www.aljazeera.com/news/2025/11/16/whats-grokipedia-musks-ai-powered-rival-to-wikipedia}
}

@article{grattafiori2024llama,
  title={The llama 3 herd of models},
  author={Grattafiori, Aaron and Dubey, Abhimanyu and Jauhri, Abhinav and Pandey, Abhinav and Kadian, Abhishek and Al-Dahle, Ahmad and Letman, Aiesha and Mathur, Akhil and Schelten, Alan and Vaughan, Alex and others},
  journal={arXiv preprint arXiv:2407.21783},
  year={2024}
}

@misc{WikipediaSizeWikipedia,
  author       = {{Wikipedia contributors}},
  title        = {Wikipedia:Size of Wikipedia},
  howpublished = {\url{https://en.wikipedia.org/w/index.php?title=Wikipedia:Size_of_Wikipedia&oldid=1335945684}},
  year         = {2026},
  publisher    = {Wikipedia, The Free Encyclopedia}
}

@article{ghosh2025mining,
  title={Mining the Mind: What 100M Beliefs Reveal About Frontier LLM Knowledge},
  author={Ghosh, Shrestha and Giordano, Luca and Hu, Yujia and Nguyen, Tuan-Phong and Razniewski, Simon},
  journal={arXiv preprint arXiv:2510.07024},
  year={2025}
}

@article{hendrycks2025hle,
  title={Humanity's last exam},
  author={Phan, Long and Gatti, Alice and Han, Ziwen and Li, Nathaniel and Hu, Josephina and Zhang, Hugh and Zhang, Chen Bo Calvin and Shaaban, Mohamed and Ling, John and Shi, Sean and others},
  journal={arXiv preprint arXiv:2501.14249},
  year={2025}
}

@inproceedings{lin2022truthfulqa,
    title = "{T}ruthful{QA}: Measuring How Models Mimic Human Falsehoods",
    author = "Lin, Stephanie  and
      Hilton, Jacob  and
      Evans, Owain",
    editor = "Muresan, Smaranda  and
      Nakov, Preslav  and
      Villavicencio, Aline",
    booktitle = "Proceedings of the 60th Annual Meeting of the Association for Computational Linguistics (Volume 1: Long Papers)",
    month = may,
    year = "2022",
    address = "Dublin, Ireland",
    publisher = "Association for Computational Linguistics",
    url = "https://aclanthology.org/2022.acl-long.229/",
    doi = "10.18653/v1/2022.acl-long.229",
    pages = "3214--3252"
}

@inproceedings{hendrycks2021mmlu,
  author = {Hendrycks, Dan and Burns, Collin and Basart, Steven and Zou, Andy and Mazeika, Mantas and Song, Dawn and Steinhardt, Jacob},
  booktitle = {ICLR},
  ee = {https://openreview.net/forum?id=d7KBjmI3GmQ},
  publisher = {OpenReview.net},
  title = {Measuring Massive Multitask Language Understanding.},
  url = {http://dblp.uni-trier.de/db/conf/iclr/iclr2021.html#HendrycksBBZMSS21},
  year = 2021
}

@inproceedings{mallen2023nottrust,
    title = "When Not to Trust Language Models: Investigating Effectiveness of Parametric and Non-Parametric Memories",
    author = "Mallen, Alex  and
      Asai, Akari  and
      Zhong, Victor  and
      Das, Rajarshi  and
      Khashabi, Daniel  and
      Hajishirzi, Hannaneh",
    editor = "Rogers, Anna  and
      Boyd-Graber, Jordan  and
      Okazaki, Naoaki",
    booktitle = "Proceedings of the 61st Annual Meeting of the Association for Computational Linguistics (Volume 1: Long Papers)",
    month = jul,
    year = "2023",
    address = "Toronto, Canada",
    publisher = "Association for Computational Linguistics",
    url = "https://aclanthology.org/2023.acl-long.546/",
    doi = "10.18653/v1/2023.acl-long.546",
    pages = "9802--9822"
}

@techreport{chatterji2025people,
  title={How people use chatgpt},
  author={Chatterji, Aaron and Cunningham, Thomas and Deming, David J and Hitzig, Zoe and Ong, Christopher and Shan, Carl Yan and Wadman, Kevin},
  year={2025},
  institution={National Bureau of Economic Research}
}

@misc{openai2025howpeopleusechatgpt,
  author = {{OpenAI}},
  title = {Wie Menschen ChatGPT nutzen},
  year = {2025},
  url = {https://openai.com/de-DE/index/how-people-are-using-chatgpt/}
}

@inproceedings{saeed-etal-2025-beyond,
    title = "Beyond Content: How Grammatical Gender Shapes Visual Representation in Text-to-Image Models",
    author = "Saeed, Muhammed  and
      Raza, Shaina  and
      Vayani, Ashmal  and
      Abdul-Mageed, Muhammad  and
      Emami, Ali  and
      Shehata, Shady",
    editor = "Christodoulopoulos, Christos  and
      Chakraborty, Tanmoy  and
      Rose, Carolyn  and
      Peng, Violet",
    booktitle = "Findings of the Association for Computational Linguistics: EMNLP 2025",
    month = nov,
    year = "2025",
    address = "Suzhou, China",
    publisher = "Association for Computational Linguistics",
    url = "https://aclanthology.org/2025.findings-emnlp.1343/",
    doi = "10.18653/v1/2025.findings-emnlp.1343",
    pages = "24673--24695",
    ISBN = "979-8-89176-335-7"
}

@inproceedings{wang2022self,
  author       = {Xuezhi Wang and
                  Jason Wei and
                  Dale Schuurmans and
                  Quoc V. Le and
                  Ed H. Chi and
                  Sharan Narang and
                  Aakanksha Chowdhery and
                  Denny Zhou},
  title        = {Self-Consistency Improves Chain of Thought Reasoning in Language Models},
  booktitle    = {The Eleventh International Conference on Learning Representations,
                  {ICLR} 2023, Kigali, Rwanda, May 1-5, 2023},
  publisher    = {OpenReview.net},
  year         = {2023},
  url          = {https://openreview.net/forum?id=1PL1NIMMrw},
  bibsource    = {dblp computer science bibliography, https://dblp.org}
}

@inproceedings{giordano-razniewski-2026-foundations,
    title = "Foundations of {LLM} Knowledge Materialization: Termination, Reproducibility, Robustness",
    author = "Giordano, Luca  and
      Razniewski, Simon",
    editor = "Demberg, Vera  and
      Inui, Kentaro  and
      Marquez, Llu{\'i}s",
    booktitle = "Findings of the {A}ssociation for {C}omputational {L}inguistics: {EACL} 2026",
    month = mar,
    year = "2026",
    address = "Rabat, Morocco",
    publisher = "Association for Computational Linguistics",
    url = "https://aclanthology.org/2026.findings-eacl.113/",
    doi = "10.18653/v1/2026.findings-eacl.113",
    pages = "2145--2164",
    ISBN = "979-8-89176-386-9"
}

@misc{anthropic2025economic,
  title        = {The Anthropic Economic Index
Understanding AI’s effects on the economy},
  author       = {{Anthropic}},
  year         = {2026},
  howpublished = {\url{https://www.anthropic.com/economic-index}},
  note         = {Accessed May 2026}
}

@article{handa2025clio,
  title={Which economic tasks are performed with ai? evidence from millions of claude conversations},
  author={Handa, Kunal and Tamkin, Alex and McCain, Miles and Huang, Saffron and Durmus, Esin and Heck, Sarah and Mueller, Jared and Hong, Jerry and Ritchie, Stuart and Belonax, Tim and others},
  journal={arXiv preprint arXiv:2503.04761},
  year={2025}
}

@inproceedings{min-etal-2023-factscore,
    title = "{FA}ct{S}core: Fine-grained Atomic Evaluation of Factual Precision in Long Form Text Generation",
    author = "Min, Sewon  and
      Krishna, Kalpesh  and
      Lyu, Xinxi  and
      Lewis, Mike  and
      Yih, Wen-tau  and
      Koh, Pang  and
      Iyyer, Mohit  and
      Zettlemoyer, Luke  and
      Hajishirzi, Hannaneh",
    booktitle = "Proceedings of the 2023 Conference on Empirical Methods in Natural Language Processing",
    year = "2023",
    url = "https://aclanthology.org/2023.emnlp-main.741/",
    doi = "10.18653/v1/2023.emnlp-main.741",
    pages = "12076--12100",
}

@inproceedings{wei2024safe,
author = {Wei, Jerry and Yang, Chengrun and Song, Xinying and Lu, Yifeng and Hu, Nathan and Huang, Jie and Tran, Dustin and Peng, Daiyi and Liu, Ruibo and Huang, Da and Du, Cosmo and Le, Quoc V.},
title = {Long-form factuality in large language models},
year = {2024},
isbn = {9798331314385},
publisher = {Curran Associates Inc.},
address = {Red Hook, NY, USA},
booktitle = {Proceedings of the 38th International Conference on Neural Information Processing Systems},
articleno = {2567},
numpages = {72},
location = {Vancouver, BC, Canada},
series = {NIPS '24}
}

@inproceedings{song-etal-2024-veriscore,
    title = "{V}eri{S}core: Evaluating the factuality of verifiable claims in long-form text generation",
    author = "Song, Yixiao  and
      Kim, Yekyung  and
      Iyyer, Mohit",
    editor = "Al-Onaizan, Yaser  and
      Bansal, Mohit  and
      Chen, Yun-Nung",
    booktitle = "Findings of the Association for Computational Linguistics: EMNLP 2024",
    month = nov,
    year = "2024",
    address = "Miami, Florida, USA",
    publisher = "Association for Computational Linguistics",
    url = "https://aclanthology.org/2024.findings-emnlp.552/",
    doi = "10.18653/v1/2024.findings-emnlp.552",
    pages = "9447--9474",
}

@inproceedings{sun2024headtail,
    title = "Head-to-Tail: How Knowledgeable are Large Language Models ({LLM}s)? {A}.{K}.{A}. Will {LLM}s Replace Knowledge Graphs?",
    author = "Sun, Kai  and
      Xu, Yifan  and
      Zha, Hanwen  and
      Liu, Yue  and
      Dong, Xin Luna",
    editor = "Duh, Kevin  and
      Gomez, Helena  and
      Bethard, Steven",
    booktitle = "Proceedings of the 2024 Conference of the North American Chapter of the Association for Computational Linguistics: Human Language Technologies (Volume 1: Long Papers)",
    month = jun,
    year = "2024",
    address = "Mexico City, Mexico",
    publisher = "Association for Computational Linguistics",
    url = "https://aclanthology.org/2024.naacl-long.18/",
    doi = "10.18653/v1/2024.naacl-long.18",
    pages = "311--325"
}

@inproceedings{rajendhran-etal-2025-verifastscore,
    title = "{V}eri{F}ast{S}core: Speeding up long-form factuality evaluation",
    author = "Rajendhran, Rishanth  and
      Zadeh, Amir  and
      Sarte, Matthew  and
      Li, Chuan  and
      Iyyer, Mohit",
    editor = "Christodoulopoulos, Christos  and
      Chakraborty, Tanmoy  and
      Rose, Carolyn  and
      Peng, Violet",
    booktitle = "Findings of the Association for Computational Linguistics: EMNLP 2025",
    month = nov,
    year = "2025",
    address = "Suzhou, China",
    publisher = "Association for Computational Linguistics",
    url = "https://aclanthology.org/2025.findings-emnlp.491/",
    doi = "10.18653/v1/2025.findings-emnlp.491",
    pages = "9234--9259",
    ISBN = "979-8-89176-335-7"
}

@article{benallal2024cosmopedia_blog,
  title={Cosmopedia: how to create large-scale synthetic data for pre-training},
  author={Allal, Loubna Ben and Lozhkov, Anton and van Strien, Daniel},
  journal={Hugging Face Blog},
  pages={56},
  year={2024}
}

@article{Liu2018WikiSum,
  author = {Liu, Peter J. and Saleh, Mohammad and Pot, Etienne and Goodrich, Ben and Sepassi, Ryan and Kaiser, Lukasz and Shazeer, Noam},
  eprint = {1801.10198},
  eprinttype = {arxiv},
  journal = {{arXiv}:1801.10198 [cs]},
  language = {en},
  title = {Generating Wikipedia by Summarizing Long Sequences},
  url = {http://arxiv.org/abs/1801.10198},
  urldate = {2018-02-19},
  year = 2018
}
